\documentclass[11pt]{article}

\usepackage[final]{acl}

\usepackage{times}
\usepackage{latexsym}

\usepackage[T1]{fontenc}

\usepackage[utf8]{inputenc}

\usepackage{microtype}
\usepackage{inconsolata}
\usepackage{hyperref}       
\usepackage{url}            
\usepackage{booktabs}       
\usepackage{amsfonts}       
\usepackage{nicefrac}                
\usepackage{CJKutf8}
\usepackage{multirow}
\usepackage{amssymb}
\usepackage{amsmath}
\usepackage{amsthm}
\usepackage{bbding}
\usepackage{graphicx}
\usepackage{makecell}
\usepackage{tcolorbox}
\usepackage{colortbl}  
\usepackage{xcolor}
\usepackage{array}
\usepackage{pgfplots}
\usepackage{utfsym}
\usepackage{listings}
\usepackage{adjustbox}
\usepackage{subfigure}
\usepackage{setspace}
\usepackage{natbib}

\usepackage{titlesec}
\titlespacing{\section}{0pt}{*1}{*1}

\definecolor{codegray}{rgb}{0.95,0.95,0.95}
\definecolor{wordcolor}{rgb}{0.2,0.6,1}
\title{TaxPraBen: A Scalable Benchmark for Structured  Evaluation of LLMs in Chinese Real-World Tax Practice}

\author{ 
	\textbf{Gang Hu$^{a}$, Yating Chen$^{a}$, Haiyan Ding$^{a}$, Wang Gao$^{c}$, Jiajia Huang$^{d}$}  \\
	\textbf{Min Peng$^{b}$, Qianqian Xie$^{b}$, Kun Yue$^{a,*}$} \\ 
	$^{a}$Yunnan University, Yunnan, China. 
	$^{b}$Wuhan University, Wuhan, China. \\
	$^{c}$Jianghan University, Wuhan, China. 
	$^{d}$Nanjing Audit University, Nanjing, China. \\ 
	hugang@ynu.edu.cn, yating\_chen@stu.ynu.edu.cn, dinghaiyan@ynu.edu.cn, \\ 
	gaow@jhun.edu.cn, huangjj@nau.edu.cn, pengm@whu.edu.cn \\
	xqq.sincere@gmail.com, kyue@ynu.edu.cn
}

\begin{document}
\maketitle

\begin{CJK}{UTF8}{gkai}

\begin{abstract}
While Large Language Models (LLMs) excel in various general domains, they exhibit notable gaps in the highly specialized, knowledge-intensive, and legally regulated Chinese tax domain. Consequently, while tax-related benchmarks are gaining attention, many focus on isolated NLP tasks, neglecting real-world practical capabilities. To address this issue, we introduce TaxPraBen, the first dedicated benchmark for Chinese taxation practice. It combines 10 traditional application tasks, along with 3 pioneering real-world scenarios: tax risk prevention, tax inspection analysis, and tax strategy planning, sourced from 14 datasets totaling 7.3K instances. TaxPraBen features a scalable structured evaluation paradigm designed through  process of "structured parsing—field alignment extraction—numerical and textual matching", enabling end-to-end tax practice assessment while being extensible to other domains. We evaluate 19 LLMs based on Bloom's taxonomy. The results indicate significant performance disparities: all closed-source large-parameter LLMs excel, and Chinese LLMs like Qwen2.5 generally exceed multilingual LLMs, while the YaYi2 LLM, fine-tuned with some tax data, shows only limited improvement. TaxPraBen\footnote{\url{https://github.com/Yating-Chen/TaxPraBen}} serves as a vital resource for advancing evaluations of LLMs in practical applications.
\end{abstract}

\section{Introduction}

Taxation is vital for governance and economic regulation, demanding precise and timely expertise due to its complex, evolving policies~\cite{rixen2022taxation}. Recent advancements in Large Language Models (LLMs) in natural language processing (NLP) offer new ways to interpret tax laws and enhance tax management, indicating a significant technological trend~\cite{nay2024large}.

However, general LLMs such as OpenAI's ChatGPT~\cite{brown2020language}, Google's Gemini~\cite{team2024gemma}, and open-source DeepSeek~\cite{guo2025deepseek}, while capable in broad domains, still underperform in specialized fields like finance~\cite{hu2024no} and auditing~\cite{jiajia2024auditwen} that demand deep semantic understanding and complex numerical reasoning. This has spurred the development of various evaluation benchmarks for domain-specific LLMs. General benchmarks like GLUE~\cite{wang2018glue}, and MMLU~\cite{hendrycks2009measuring} focus on tasks like reading comprehension, making them inadequate for specialized assessments. This has led to the rise of domain-specific benchmarks, such as FinBen~\cite{xie2024finben} for finance, MedBench~\cite{cai2024medbench} for Medicine, and LAiW~\cite{dai2025laiw} for law. However, there is a relative scarcity of models and benchmarks specific to the tax scenarios, primarily due to a lack of large high-quality annotated data~\cite{choi2025taxation}. Existing overseas tax data studies~\cite{steinigen2023semantic,louisbrulenaudet2023} often differ significantly from China's tax management realities, complicating adaptation efforts. Moreover, some LLM studies~\cite{luo2023yayi} incorporate tax data that is not open source and lacks real-world applicability. Consequently, the proficiency of these LLMs in tax knowledge remains unclear, underscoring the need for custom tax-specific benchmarks.


Even the recently proposed TaxBen~\cite{chen2025taxben}, like other domain-specific benchmarks, focuses mainly on isolated NLP tasks such as text classification, generation, and reasoning; comparisons are provided in Appendix~\ref{sec:benchmark}. General benchmarks have similar limitations. For example, MATH~\cite{hendrycks2021measuring} emphasizes mathematical correctness over explanation quality, while GAOKAO~\cite{zhang2023evaluating} and AGIEval~\cite{zhong2024agieval} focus on exam-style reasoning but overlook mixed-format outputs in professional settings. However, many real-world tax and financial auditing tasks require both semantic reasoning and quantitative calculation~\cite{krumdick2024bizbench}. Existing domain-specific benchmarks, including TaxBen, overemphasize language understanding and neglect structured outputs, thus overstating the practical capabilities of current LLMs. As shown in Appendix~\ref{sec:contrast}, some LLM rankings on TaxBen are inflated: models perform well on isolated numerical reasoning or semantic matching tasks, but struggle in real scenarios requiring both. Overall, existing benchmarks, including TaxBen, cannot reliably assess holistic practical capabilities by averaging NLP tasks. This suggests that real-world tax scenarios require LLMs to integrate semantic understanding and numerical reasoning for tax law interpretation, calculation, and compliance assessment. Therefore, traditional NLP benchmarks fail to capture the real demands of Chinese tax practice, posing challenges for LLM evaluation in this domain.

With this in mind, we introduce TaxPraBen, the first benchmark for evaluating LLMs in Chinese tax practice. Developed with tax experts, TaxPraBen contains 14 datasets with 7.3K samples, covering carefully annotated tax data and guiding prompts. All data are manually collected and annotated with model assistance rather than taken from public datasets. We also build a complete workflow integrating data annotation, evaluation metrics, and structured assessment, making TaxPraBen scalable. Following Bloom's taxonomy of cognitive skills~\cite{fei2024lawbench,chen2025taxben}, we divide the tasks into three groups: (1) \textit{Knowledge Memorization}, (2) \textit{Knowledge Understanding}, and (3) \textit{Knowledge Application}.

Evaluating 19 representative general LLMs on TaxPraBen yields the following results: (1) \textbf{Overall Performance Variation}: ERNIE-3.5, Grok3, and ChatGPT perform best in both zero-shot and one-shot settings, highlighting the value of large-scale parameters and knowledge enhancement for tax tasks. (2) \textbf{Task Performance Bias}: Tax knowledge application is more challenging than understanding and memorization, reflecting the domain’s reliance on economic activities and contextual scenarios. (3) \textbf{Language Background Advantage}: Chinese models perform better on tasks involving Chinese tax terminology and policies, showing the benefit of language-specific optimization. (4) \textbf{Data Coverage Gaps}: Tax-data fine-tuning still brings limited gains, likely due to insufficient data coverage or task mismatch. (5) \textbf{Reasoning Task Challenges}: All models struggle with reasoning tasks, revealing weaknesses in numerical computation and tax-related logic understanding. (6) \textbf{Example Introduction Pitfall}: One-shot examples may cause over-reliance on exemplars, hurting generalization across diverse reasoning tasks.

Our contributions are as follows:
1) \textbf{Introducing TaxPraBen}, the first benchmark specifically designed for Chinese tax practice, incorporating real-world scenarios such as risk prevention, inspection analysis, and strategy planning.
2) \textbf{Filling the gap in scarce tax datasets}, curated by domain experts and enhanced with ChatGPT-assisted annotations, providing high-quality data and well-designed prompts.
3) \textbf{Establishing a taxonomy for tax tasks}, organizing the dataset according to Bloom's cognitive taxonomy to assess capabilities in memorization, understanding, and application.
4) \textbf{Conducting in-depth evaluations of 19 popular LLMs}, revealing and discussing their strengths and weaknesses in tax-related tasks.
5) \textbf{Providing multi-evaluations across tax and NLP tasks}, helping analyze LLMs' shortcomings and potential strengths.
6) \textbf{Proposing a structured evaluation method} that extends beyond tax, combining semantic reasoning and numerical accuracy for practical assessment in law, healthcare, finance, and more.

\begin{figure*}
    \fbox{\includegraphics[width=0.98\textwidth]{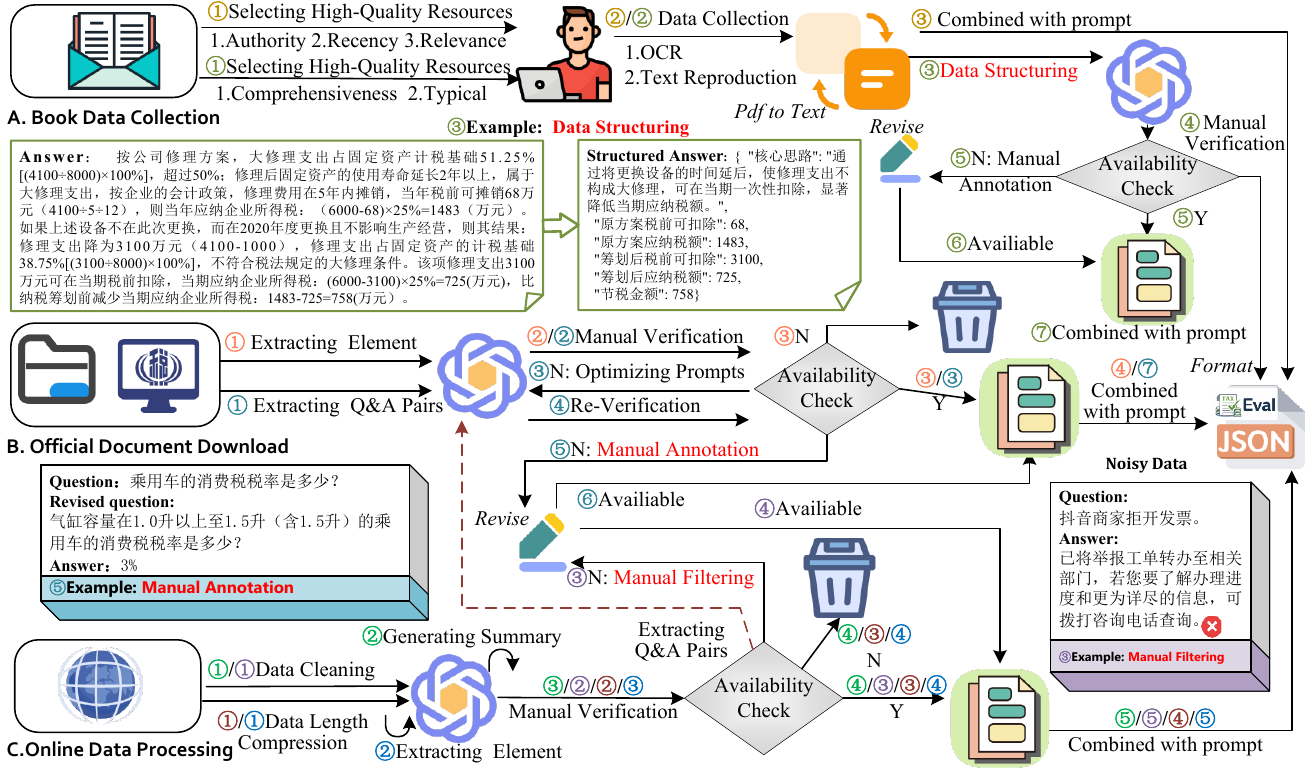}}
    \caption{TaxPraBen's data construction workflow uses 3 methods: (A) Book Data Collection, (B) Official Document Download and (C) Online Data Processing. All data is annotated via the manual "Availability Check". } 
    \label{fig:fig1}
\end{figure*}

\section{Related Work}


\noindent\textbf{(1) Domain-specific LLMs.} LLMs like OpenAI’s GPT~\cite{brown2020language} and Meta’s LLaMA~\cite{touvron2023LLaMa} excel in various NLP tasks but struggle with Chinese-specific applications due to their English-centric training. To address this, models like Gemini~\cite{team2024gemma}, DeepSeek~\cite{guo2025deepseek}, and Qwen~\cite{bai2023qwen} have been developed to better capture Chinese linguistic and cultural nuances. Domain-specific models such as PIXIU~\cite{xie2023pixiu} and HuaTuo~\cite{wang2023huatuo} show promise in fields like auditing~\cite{jiajia2024auditwen}, education~\cite{yu2023taoli}, healthcare~\cite{chen2023bianque}, law~\cite{zhou2024lawgpt}, finance~\cite{hu2024no}, and psychological counseling~\cite{chen2023soulchat}. Research has also examined integrating tax data into the pretraining or fine-tuning of LLMs~\cite{luo2023yayi} in legal and financial contexts. However, it remains uncertain how effectively these models can handle nuanced tax-related knowledge.


\noindent\textbf{(2) Chinese LLM Benchmarks.} Several benchmarks have been developed to evaluate Chinese LLMs. CLEVA~\cite{li2023cleva}, C-Eval~\cite{huang2023c}, and GAOKAO~\cite{zhang2023evaluating} assess general knowledge and reasoning. MMCU~\cite{zeng2023measuring} and Xiezhi~\cite{gu2024xiezhi} focus on fields like medicine and law, while CG-Eval~\cite{zeng2024withdrawn} covers multiple disciplines. Domain-specific benchmarks include FinBen~\cite{xie2024finben} for finance and CMB~\cite{wang2024cmb} for medicine. However, the tax domain is underrepresented, with datasets like LawBench~\cite{fei2024lawbench} and LAiW~\cite{dai2025laiw} offering limited tax datasets, hindering the evaluation of tax reasoning abilities. Although TaxBen~\cite{chen2025taxben} addresses tax-related NLP tasks, it lacks the integration of semantic interpretation and numerical analysis. Thus, it is essential to refine the assessment framework for taxation competence in real-world application scenarios.



\section{Overview design of TaxPraBen}

\subsection{Task Categorization}

Establishing a scientific and systematic evaluation framework is crucial for assessing LLMs in the tax domain. We develop an evaluation system based on Bloom's Cognitive Taxonomy~\cite{fei2024lawbench}, which includes six cognitive levels. To simplify the evaluation, we draw from existing task taxonomies~\cite{xie2024finben}, concentrating on the first three levels to create three core tax capability assessments: 1) Knowledge Memorization (KM): Examines models' ability to accurately recall and recite tax legal provisions, policy regulations, and institutional documents; 2) Knowledge Understanding (KU): Tests models' capacity to identify key information from tax materials and comprehend policy implications; 3) Knowledge Application (KA): Evaluates models' comprehensive ability to apply tax regulations, policy clauses, and computational methods in addressing practical tax issues.

\begin{table*}[htb!]
    \centering
    \scriptsize
    \setlength\tabcolsep{5pt}
    \renewcommand{\arraystretch}{0.9}
   
    \begin{tabular}{cccccccc}
    \toprule
    \multirow{2}{*}{\textbf{Tax  Ability-Hierarchy Task}} & \multirow{2}{*}{\textbf{\makecell{Dataset \\ Name}}} & 
    \multirow{2}{*}{\textbf{Specific Task}}     & \multirow{2}{*}{\textbf{\makecell{Data \\ Scale}}}   & \multirow{2}{*}{\textbf{Evaluation Metric}} & \multirow{2}{*}{\makecell{\textbf{NLP- Task Type}}} & \multicolumn{2}{c}{\textbf{Method}} \\
     &     &   &      &    &  &  HG  &  MC\\
    \hline
    \makecell{Tax Knowledge \\ Memorization (KM)}      &  TaxRecite & Tax Law Recitation    & 200      & BERTScore, BARTScore &Generation (GEN)  &  $\checkmark$   &   \\ 
    \hline
    \multirow{3}{*}{\makecell{Tax Knowledge \\ Understanding (KU)}}  & TaxSum    & Tax News Summarization & 1000  & BERTScore, BARTScore & Generation (GEN)  &  $\checkmark$  &   \\ 
    &TaxTopic  & Tax Topic Classification & 1000   & Accuracy, F1, Macro F1  & Classification (CLS) &    & $\checkmark$  \\ 
    &TaxRead & Tax Reading Comprehension & 1000  & EM Accuracy & Generation (GEN)  &  $\checkmark$  &   \\
    \hline
    \multirow{10}{*}{\makecell{Tax Knowledge \\ Application (KA)}}
    & TaxCalc & Tax Payment Calculation & 500  & EM Accuracy & Reasoning (REA) &   & $\checkmark$ \\ 
    &TaxSCQ & Tax Single-Choice Exam & 700   & Accuracy, F1, Macro F1  & Classification (CLS)  &   & $\checkmark$  \\ 
    &TaxMCQ & Tax Multiple-Choice Exam & 400  & EM Accuracy & Classification (CLS) &   & $\checkmark$  \\ 
    &TaxQA & Tax Knowledge Q\&A & 700   & BERTScore, BARTScore & Generation (GEN)  &  $\checkmark$  & \\ 
    &TaxBoard & Tax Board Q\&A & 500  & BERTScore, BARTScore & Generation (GEN)  &  $\checkmark$  &  \\ 
    &TaxCrime & Tax Law Identification & 200  & Accuracy, F1, Macro F1 & Classification (CLS)  &  $\checkmark$  &  \\ 
    &TaxOpinion & Tax Opinion Summarization & 500  & BERTScore, BARTScore & Generation (GEN)  &  $\checkmark$  &  \\ 
    &TaxRisk & Tax Risk Prevention & 200  & BERTScore & Generation (GEN)  &  $\checkmark$  &  \\ 
    &TaxInspect & Tax Inspection Analysis & 200  & BERTScore, EM Accuracy & Generation (GEN)  &  $\checkmark$  &  \\ 
    &TaxPlan & Tax Strategy Planning & 200  & BERTScore, EM Accuracy & Reasoning (REA)  &  $\checkmark$  &  \\ 
    \bottomrule
    \end{tabular}
     \caption{Overview of tax tasks, datasets, specific tasks, data scale, evaluation metrics, NLP types, and creation methods in TaxPraBen. "MC" and "HC" refer to "Manually Created" and "Human-ChatGPT Collaborative". }
    \label{tab:dataset}
\end{table*}

\subsection{Data Processing}
We construct the TaxPraBen benchmark using a multi-source data fusion strategy, comprising three acquisition pipelines: (A) Book Data Collection, (B) Official Document Download, and (C) Online Data Processing, as illustrated in Figure~\ref{fig:fig1}. These pipelines integrate tax-specific sources, ChatGPT-based processing, and rigorous human validation to ensure data quality, structural consistency, and alignment with real-world tax applications. The manual check process is shown in Appendix~\ref{sec:annotation}.

\textbf{(A) Book Data Collection:} We select tax exam guidebooks and tax planning casebooks as data sources. The guidebooks are filtered based on authority, timeliness, and relevance, while the casebooks are selected according to comprehensiveness and typicality. Using OCR and text reproduction techniques, we extract various types of tax exam questions (\textit{TaxSCQ}, \textit{TaxMCQ} and \textit{TaxCalc}) from PDF files. For tax planning cases, we employ ChatGPT to perform structured information extraction, followed by manual verification (\textit{TaxPlan}).

\textbf{(B) Official Document Download:} We collect tax-related policy documents, regulations, and official announcements from the State Administration of Taxation. By integrating prompt engineering with ChatGPT, we generate question–answer (Q\&A) pairs (\textit{TaxRecite} and \textit{TaxQA}). Each round of generation is followed by manual review, refinement of prompts based on identified issues, and expert validation for factual correction and completion. In addition, we retrieve criminal case verdicts from China Judgments Online and use ChatGPT to extract structured elements related to offense behavior and violated legal articles (\textit{TaxCrime}).

\textbf{(C) Online Data Processing:} First, we crawl tax-related news, risk control reports, and audit cases from the Tax House website. Long-form texts are compressed using ChatGPT to meet input length constraints. For tax news, we extract title-topic pairs (TaxTopic) and article-summary pairs (\textit{TaxSum}). For risk control and inspection cases, we extract structured key elements using ChatGPT (\textit{TaxRisk} and \textit{TaxInspect}). Second, we retrieve public tax-related opinion posts involving tax evasion from an online sentiment monitoring platform. A two-stage ChatGPT process is applied: first to assess tax relevance and clean noise, and second to compress the input and generate abstractive summaries (\textit{TaxOpinion}). Third, we scrape user interactions with officials from the State Administration of Taxation’s message board, using ChatGPT to filter out irrelevant or redundant content (\textit{TaxBoard}). All generated content is manually validated for usability and alignment with tax-specific semantics.

Overall, the data definition and collection details for the three-level tax tasks in the evaluation benchmark TaxPraBen are shown in Appendix~\ref{sec:process}.

\subsection{Instruction Construction}


To evaluate the performance of LLMs in various tax tasks, we collaborate with tax-related experts to annotate instructions for standardized outputs. We form a professional prompt annotation and evaluation team consisting of teachers and students. Evaluators score prompts on a 4-point scale ("Strongly Disagree (0)" to "Strongly Agree (3)"), focusing on accuracy (ACC), naturalness (NAT), and informativeness (INF). We test these prompts on  the ChatALL platform. To ensure scoring reliability, we calculate consistency scores using Fleiss' Kappa and Krippendorff's Alpha, retaining only high-quality prompts with an average score exceeding 2. Detailed annotation processes and consistency analysis results are available in Appendix~\ref{sec:prompt}. The final instructions combine input texts, labels, and validated prompts, evaluated in JSON format.

As shown in Appendix~\ref{sec:statistics}, TaxPraBen features a three-tier taxonomy of tax tasks, with a balanced distribution of 200 to 1000 instances per type. This design enables fair evaluation across cognitive levels, while diverse inputs and prompt lengths enhance task complexity. Overall, TaxPraBen provides a comprehensive and challenging benchmark for assessing LLMs in the Chinese tax practice domain. Unlike most domain-specific benchmarks that rely solely on public datasets~\cite{dai2025laiw,fei2024lawbench}, TaxPraBen integrates expert design and manual annotation to ensure datasets reflect real tax scenarios and meet specific evaluation goals. Table~\ref{tab:dataset} summarizes the task types, dataset statistics, and evaluation metrics in TaxPraBen.

\begin{table*}[htb!]
    \centering
    \scriptsize
    \setlength\tabcolsep{2pt}
    \renewcommand{\arraystretch}{0.95}
    
    \begin{tabular}{ m{1.3cm}<{\centering} |m{6cm}<{\centering} |m{8cm}<{\centering} }
    \toprule
    \textbf{Dataset} & \textbf{Real-World Scenarios} & \textbf{Application Value}  \\ 
    \hline
    TaxCalc & Pre-filing tax calculation for individuals and businesses. & Supports the underlying logic for intelligent tax filing tools.  \\ 
    \hline
    TaxSCQ & Exams for tax system staff or entry qualifications. & Enables training/testing for tax Q\&A systems or AI tutors.  \\ 
    \hline
    TaxMCQ & Complex policy applicability and compliance judgment. & Trains models for regulatory perception and decision-making.  \\ 
    \hline
    TaxQA & Inquiries to tax bureaus (e.g., 12366), tax advisory. & Builds intelligent tax consultation systems.  \\ 
    \hline
    TaxBoard & Message boards and online government services. & Enhances automatic response and assistant capabilities in e-government.  \\ 
    \hline
    TaxCrime & Legal article application in law enforcement. & Supports case legality judgment and transparent law enforcement.  \\ 
    \hline
    TaxOpinion & Public opinion monitoring and sentiment analysis. & Serves risk warning and government response mechanisms.  \\ 
    \hline
    TaxRisk & Financial audits and tax risk control. & Builds AI systems for risk identification and response strategies.  \\ 
    \hline
    TaxInspect & Extraction of key elements in enforcement cases. & Supports casebase development for training and enforcement.  \\ 
    \hline
    TaxPlan & Tax-saving planning in enterprise operations. & Enables AI-assisted tax planning and optimization decisions.  \\ 
    \bottomrule
    \end{tabular}%
    \caption{TaxPraBen subsets for Knowledge Application (KA) task: real-world scenarios and application value.}
    \label{tab:application}
\end{table*}

\subsection{Practical Relevance}

\begin{figure*}[t!] 
    \centering
    \scriptsize
    \vspace{-1em}
    \tcbset{acltaxstyle/.style={colframe=black, colback=white, boxrule=0.4pt, arc=0pt, outer arc=0pt, boxsep=0pt, left=3pt, right=3pt, top=3pt, bottom=3pt}}
    
    \begin{minipage}[t]{0.325\textwidth}
        \begin{tcolorbox}[acltaxstyle]
        {\centering \textbf{\textit{Case 1}: TaxRisk (Tax Risk Prevention)}\par}
        \vspace{2pt}\hrule height 0.5pt \vspace{2pt}
        \raggedright
        \{"\textcolor{blue}{面临的风险}": 买受人无法取得增值税专用发票，影响进项税金抵扣权益，进而引发纠纷，浪费大量行政和司法资源。
        \par\smallskip
        "\textcolor{blue}{对应的解决措施}": 加强税法和行政制度供给，完善增值税抵扣权制度，畅通司法诉讼救济途径，加强司法和行政部门的协作机制。\}
        \end{tcolorbox}
    \end{minipage}
    \hfill 
    \begin{minipage}[t]{0.305\textwidth}
        \begin{tcolorbox}[acltaxstyle]
        {\centering \textbf{\textit{Case 2}: TaxInspect (Tax Inspection Analysis)}\par}
        \vspace{4pt}\hrule height 0.5pt \vspace{5pt}
        \raggedright
        \{"\textcolor{blue}{犯罪行为}": 虚开增值税专用发票，非法抵扣税款。
        \par\smallskip
        "\textcolor{blue}{所犯罪名}": 虚开发票罪。
        \par \smallskip
        "\textcolor{blue}{处罚结果}": 判处有期徒刑一年六个月，缓刑二年，并处罚金人民币13万元。\}
        \end{tcolorbox}
    \end{minipage}
    \hfill
    \begin{minipage}[t]{0.35\textwidth}
        \begin{tcolorbox}[acltaxstyle]
        {\centering \textbf{\textit{Case 3}: TaxPlan (Tax Strategy Planning)}\par}
        \vspace{3pt}\hrule height 0.5pt \vspace{3pt} 
        \raggedright 
        \{"\textcolor{blue}{核心思路}": 时间延后，使修理支出不构成大修理可在当期一次性扣除，显著降低当期应纳税额。
        \par\smallskip 
        "\textcolor{blue}{原方案税前可扣除}": 68, "\textcolor{blue}{原方案应纳税额}": 1483
        \par\smallskip
        "\textcolor{blue}{筹划后税前可扣除}": 3100, "\textcolor{blue}{筹划后应纳税额}": 725
        \par\smallskip
        \textcolor{blue}{节税金额}: 758\} 
        \end{tcolorbox}
    \end{minipage}
     
    \vspace{-0.5em}
    \caption{A unified output format protocol for 3 typical cases of the tax practice scenarios.}
    \label{fig:cases}
    \vspace{-1em}
\end{figure*}


To show the practical relevance of the Knowledge Application (KA) task in TaxPraBen, we map each tax dataset to the real-world scenario in Table~\ref{tab:application}. These tasks not only assess model performance but also simulate essential functions in modern tax workflows, including calculating tax liabilities, determining applicable policies, and developing tax-saving strategies. They support intelligent government systems through automated consultations and risk identification, while assisting regulatory enforcement with compliance assessments. This alignment underscores the value of the TaxPraBen benchmark in model performance evaluation and the development of reliable tax AI systems.

\section{Structured Evaluation for TaxPraBen}

\subsection{Unified Output Protocol}

In tax practice domain, assessing LLM outputs requires semantic accuracy and structural alignment with real-world use cases. TaxPraBen introduces a unified output protocol for various practice areas, enabling structured responses and automatic evaluation across LLMs. As shown in Figure~\ref{fig:cases}, we define standardized JSON output schemas for 3 classic cases: (1)~\textbf{TaxRisk}: Extract key elements from case narratives, including identified risks and solutions, and compute the average BERTScore for all fields. (2)~\textbf{TaxInspect}: Perform a mixed matching using EM Accuracy for criminal charges and BERTScore for violation descriptions and penalties to ensure semantic precision. (3)~\textbf{TaxPlan}: Create a tax planning strategy that integrates text generation and numerical reasoning, and assess semantic similarity and numerical accuracy. The unified output formatting protocol enhances the evaluability and reliability of model outputs by transforming free-form generations into structured data, enabling automatic scoring, cross-model comparisons, and batch evaluations, while also facilitating integration with real-world practice applications.

\subsection{Aligned Field Evaluation}

\begin{table}[t]
    \centering
    \scriptsize
    \setlength\tabcolsep{1pt}
    \renewcommand{\arraystretch}{0.9}
    
    \begin{tabular}{p{0.05\linewidth} | p{0.90\linewidth}}
    \hline
    \multicolumn{1}{c|}{\textbf{Task}} & \multicolumn{1}{c}{\textbf{Prompt Description}} \\
    \hline
    \multirow{9}{*}{\rotatebox{70}{TaxRisk}} &
    你会收到一段模型的输出内容，
    为了精准地匹配，你需要做的是判断模型有没有提取出内容。如果有，则从中提取模型输出的这两部分内容，如果模型本身就没有总结出来，你就留空，不需要替它总结，严格按照以下格式输出，留空的也需要保持格式：\{"面临的风险": , "风险对应的解决方案": \}。模型输出如下：\{raw\_output\}。请只返回JSON对象，不要任何额外解释。\\
    \hline
    \multirow{8}{*}{\rotatebox{70}{TaxInspect}} &
    你会收到一段模型的输出内容，
    为了精准地匹配，你需要做的是判断模型有没有提取出内容。如果有，则从中提取模型输出的这三部分内容，如果模型本身就没有总结出来，你就留空，不需要替它总结，严格按照以下格式输出，留空的也需要保持格式：\{"犯罪行为": , "所犯罪名": , "处罚结果": \}。模型输出如下：\{raw\_output\}。请只返回JSON对象，不要任何额外解释。\\
    \hline
    \multirow{8}{*}{\rotatebox{70}{TaxPlan}} &
    请你从模型输出中提取填空答案，若某一字段没有找到答案，则该字段留空，不要凭空生成答案，并严格按照以下格式输出：\{"核心思路": , "会计利润": , "第一方案所得税": , "第二方案所得税": , "分年捐赠筹划后所得税": , "节税额": \}，除核心思路外，其余字段都只返回数值，不要有任何符号和单位。模型输出如下：\{raw\_output\}。请只返回JSON对象，不要任何额外解释。\\
    \hline
    \end{tabular}
    \caption{Structured output guidance prompts.}
    \label{tab:prompts}
\end{table}

Despite having explicit output templates, most open-source LLMs often still produce inconsistent outputs, such as missing fields, malformed JSON, or mixed text. This hinders automated evaluation for structure-sensitive tax practice scenarios: TaxRisk, TaxInspection, and TaxPlan, as well as cross-model comparisons and batch assessments. 

To tackle this issue and reduce API costs, we leverage ChatGPT-3.5 as a text structure-aware parser. It outperforms fragile regex with strong semantic comprehension and tolerance for formatting noise, requiring no fine-tuning unlike supervised structure predictors. This makes it a lightweight, adaptive solution for normalizing model outputs.  Structured output guidance prompts for the 3 practice tax scenarios are shown in Table~\ref{tab:prompts}. As seen in Appendix~\ref{sec:alignment}, targeted prompting allows ChatGPT to fully parse unstructured content, strip redundant text, and extract all critical fields. We then conduct automated evaluations following the metric calculation methodology in Section~\ref{sec:metrics}.

This strategy reduces evaluation sensitivity to output structure, ensuring that minor deviations do not interfere with results. It is model-agnostic, compatible with both commercial and open-source LLMs, enabling unified cross-system assessments while reducing reliance on brittle rule-based scripts and manual templates, thus making the evaluation pipeline more scalable and maintainable.

\begin{figure}
    \includegraphics[width=0.48\textwidth]{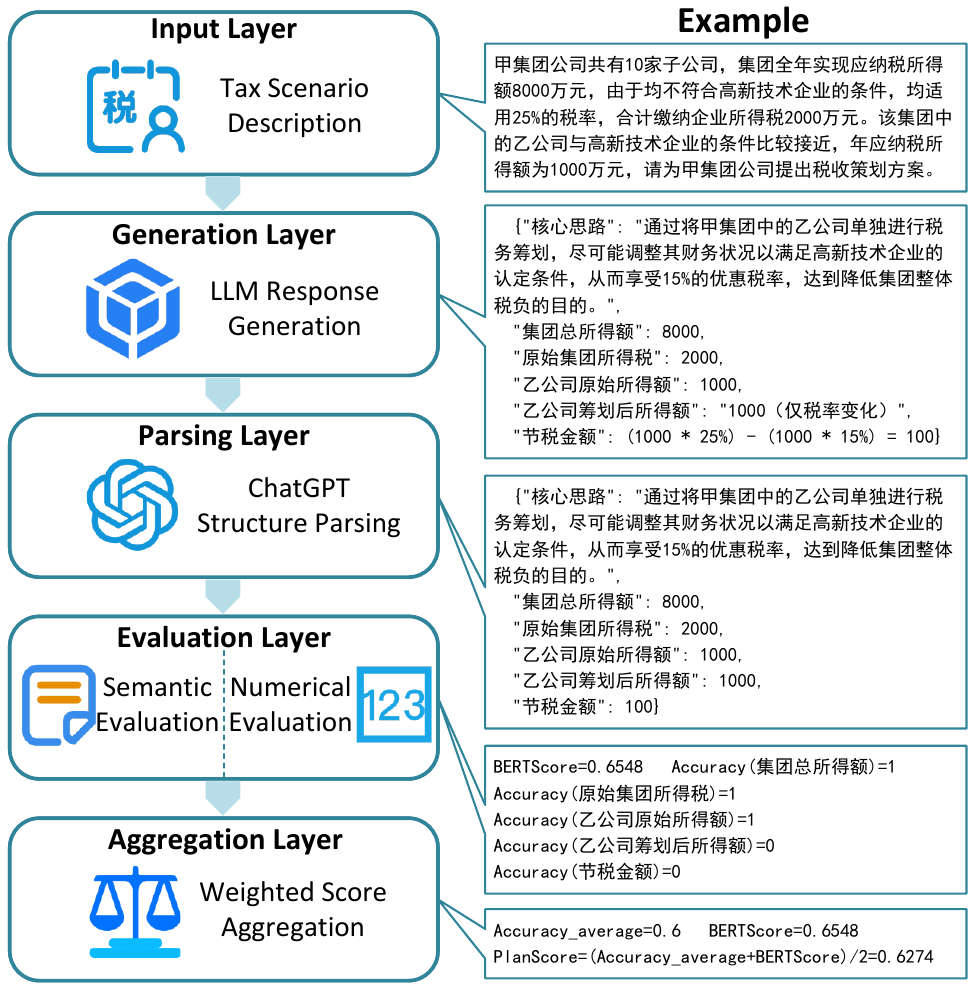}
    \caption{Structured evaluation pipeline for TaxPlan.} 
    \label{fig:pipeline}
\end{figure}

\subsection{Real-World Practice Case}



We take TaxPlan as an example to present our structured evaluation pipeline in Figure~\ref{fig:pipeline}: Feed tax scenario descriptions to target LLMs, with unified output formatting prompts to guide them to generate predefined JSON-style responses.
Since open-source models often fail to follow formatting rules, we use ChatGPT as a structure parser to clean noisy outputs, convert them into valid JSON and extract key fields. Conduct two-dimensional evaluation: use BERTScore to assess text description quality, and EM Accuracy to check numerical matching precision. Calculate the final score via weighted average to fully reflect both reasoning rationality and quantitative accuracy of the model outputs.



\section{Scalable Promotion of TaxPraBen}

Our structured evaluation framework, though originally developed for tax planning tasks, demonstrates significant extensibility across domains. Many real-world scenarios require models to generate hybrid outputs that integrate free-text reasoning with critical numerical data. Traditional evaluation methods, which typically focus solely on textual fluency, lexical overlap, or single-token accuracy, often fail to comprehensively assess such responses in terms of both semantic completeness and numerical correctness. Our proposed structured tax evaluation methodology is, in fact, generalizable to numerous professional domains such as law, finance, and healthcare (as demonstrated in Appendix~\ref{sec:scalable})—all fields that similarly require the generation of hybrid content combining combining "explanatory text + critical numbers". This approach effectively addresses significant gaps in conventional single-metric evaluation pipelines.

\section{Experiments}

\subsection{Baseline Modes}


We assess representative LLMs based on their availability, popularity, and performance across various benchmarks. They are listed in Table~\ref{tab:models}, with detailed descriptions provided in Appendix~\ref{sec:models}.

\begin{table}[htb!]
    \centering
    \scriptsize
    \setlength\tabcolsep{3pt}
    \renewcommand{\arraystretch}{0.9}
    
    \begin{tabular}{c|c|c|c|c|c|c}
    \toprule 
    \textbf{Model}  & \textbf{P} &   \textbf{Pre}  & \textbf{Fin}  & \textbf{Access} & \textbf{Base LLM} &  \textbf{Release} \\ 
    \hline
    \multicolumn{7}{c}{\textbf{Multilingual General LLMs}}  \\ 
    \hline
     ChatGPT & 175B &$\checkmark$ & $\times$  & API   &  --&   11/2022 \\
     GPT-4o & 200B &$\checkmark$ & $\times$  & API   &  --&   05/2024 \\
     MistralV0.3 & 7B & $\checkmark$ & $\times$   & Weights & -- & 07/2024 \\
     Gemma  & 7B & $\checkmark$ & $\times$  & Weights & -- & 07/2024 \\ 
     LLaMA3  & 8B & $\checkmark$ & $\times$   & Weights & -- & 09/2024 \\ 
     Bayling2 &  7B &  $\checkmark$ & $\times$   & Weights & LLaMA2  & 11/2024 \\
     Grok3  &  1200B &  $\checkmark$ & $\times$  & API &  --  &  02/2025\\
     \hline
     \multicolumn{7}{c}{\textbf{Chinese-oriented LLMs}} \\ 
     \hline
    $\text{DeepSeek}_\text{llm}$  & 7B & $\checkmark$ & $\times$  & Weights & -- & 11/2023 \\
    Baichuan2 & 7B & $\checkmark$ & $\times$ & Weights & -- & 01/2024 \\
     Atom &  7B & $\times$  & $\checkmark$  & Weights & LLaMA2-7B & 02/2024 \\
     Qwen2.5 & 7B & $\checkmark$ & $\times$  & Weights & -- & 02/2024 \\
     ChnLLaMA3 & 8B & $\times$  & $\checkmark$  & Weights & LLaMA3-7B & 04/2024 \\
     ERNIE-3.5 & $\sim$ 1000B & $\checkmark$ & $\times$  & API & -- & 07/2024 \\ 
     ChatGLM3  & 6B & $\checkmark$ &$\times$   & Weights & -- & 08/2024 \\ 
     Yi  &  6B & $\checkmark$ & $\times$   & Weights & -- & 11/2024 \\
     GLM4 & 9B & $\checkmark$ &$\times$   & Weights & --  & 11/2024 \\ 
     $\text{DeepSeek}_\text{R1}$  & 7B & $\checkmark$ & $\times$  & Weights & -- & 02/2025 \\
     InternLM2.5 & 7B & $\checkmark$ & $\times$  & Weights & -- & 03/2025\\
     \hline
    \multicolumn{7}{c}{\textbf{Tax-related LLMs}}  \\ 
     \hline
     YaYi2 & 30B & $\checkmark$ & $\times$ & Weights & -- & 03/2024 \\ 
     \bottomrule
    \end{tabular} 
    \caption{Various baseline LLMs, including parameters (P), pre-trained (Pre) or fine-tuned (Fin), access methods (public weights/API calls), base LLM, and release date.}
    \label{tab:models}
\end{table}

\subsection{Evaluation Metrics}
\label{sec:metrics}

\begin{figure*}[htb!]  
    \centering
    \makeatletter
    \def\@captype{figure} 
    \makeatother
    \begin{minipage}{0.45\textwidth}  
        \centering
        \includegraphics[width=\textwidth]{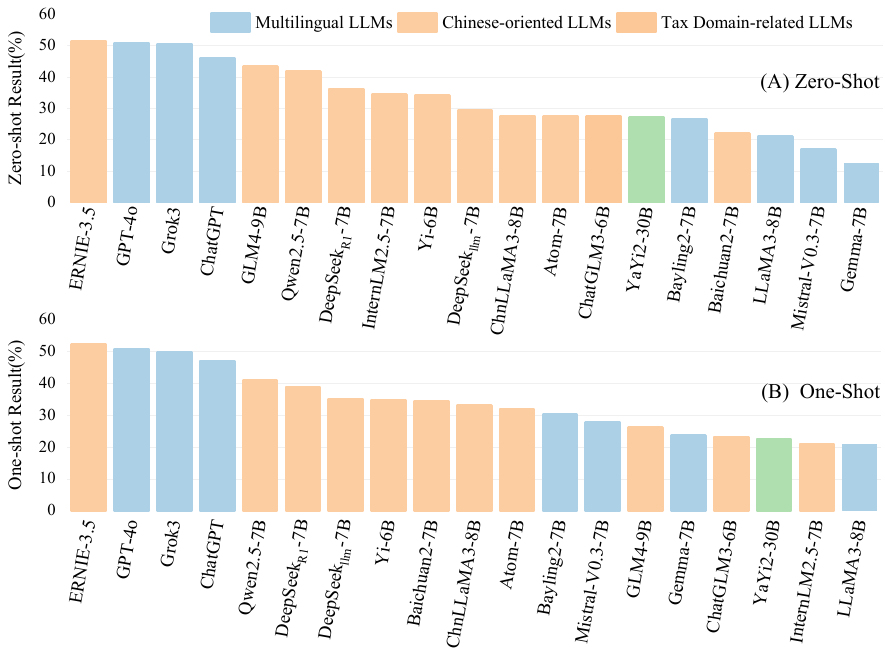}
    \end{minipage}
    \makeatletter
    \def\@captype{figure} 
    \makeatother
    \begin{minipage}{0.53\textwidth}  
        \centering
        \includegraphics[width=\textwidth]{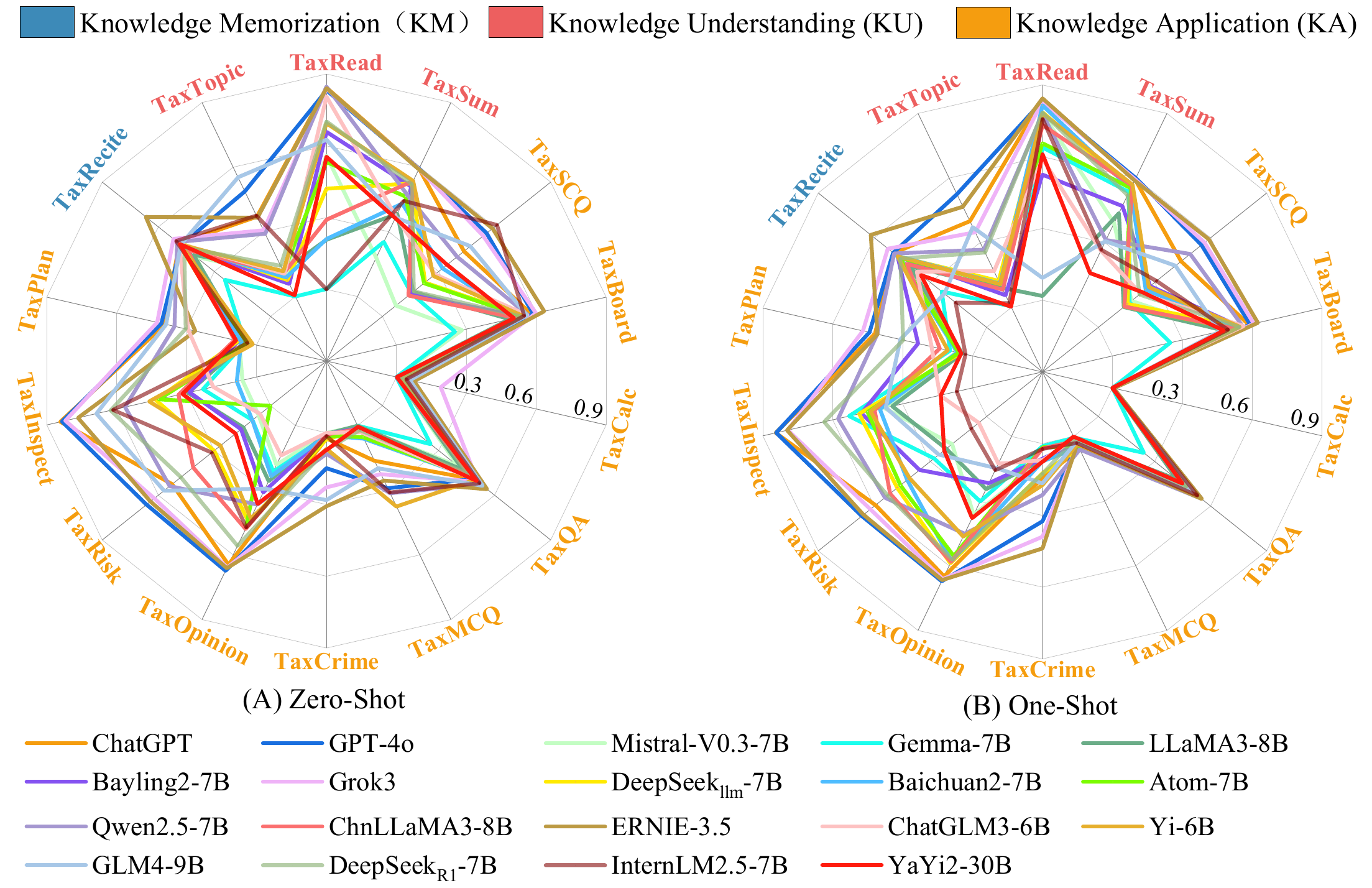}
    \end{minipage}
    \caption{The zero-shot and one-shot overall performance of the 19 popular LLMs evaluated on TaxPraBen.}
    \label{fig:overall}
\end{figure*}

In accordance with~\cite{xie2023pixiu,hu2024no}, we adopt a unified JSON template (see Table~\ref{tab:json}) to organize task-specific evaluation metrics for the 5 output types in TaxPraBen, as detailed below: \textbf{(a)}~\textit{Text Classification Tasks}: Datasets like TaxTopic, TaxSCQ, and TaxCrime involve discrete label prediction. We use Accuracy (ACC), F1 score, and Macro-F1 to measure performance. \textbf{(b)}~\textit{Text Generation Tasks}: Datasets like TaxRecite, TaxSum, TaxQA, TaxBoard, TaxOpinion, and TaxRisk require free-form or structured textual responses. For these, we apply BERTScore and BARTScore to assess semantic similarity. In TaxRisk and TaxInspect, which generate multiple semantic aspects, we compute the average BERTScore. \textbf{(c)}~\textit{Structured Prediction Tasks}: Datasets like TaxRead, TaxCalc, and TaxMCQ require deterministic outputs in fixed formats (e.g., numbers or texts). We evaluate them using Exact Match (EM) Accuracy.  \textbf{(d)}~\textit{Mixed Matching Task}: TaxInspect and TaxPlan combine absolute matches for numerical/text fields and semantic matches for explanatory rationale, averaging EM Accuracy and BERTScore for the final score. To ensure fair assessment across tasks and data types, we compute the overall average of metrics from 5 runs, as defined in Appendix~\ref{sec:computation}. All overall average metrics range from 0 to 1, with higher values indicating better performance.

\begin{table}[htb!]
    \centering
    \small
    
    \setlength\tabcolsep{1pt}
    \renewcommand{\arraystretch}{1}
    \begin{tabular}{|p{7.5cm}|}
    \hline 
    \rowcolor{blue!2}  \{\textbf{Id}:\textcolor{blue}{[data\_id]} \makebox[0.05cm]{}\textbf{query}:\textcolor{blue}{[prompt]}\makebox[0.05cm]{}\textbf{text}:\textcolor{blue}{[text]}\makebox[0.05cm]{}\textbf{answer}:\textcolor{blue}{[response]}\\
    \rowcolor{blue!2}  \makebox[0.2cm]{}  (a) \textbf{choices}:\textcolor{blue}{[...]}  \textbf{gold}:\textcolor{blue}{[...]} \makebox[0.2cm]{} / (b) ---  \makebox[0.2cm]{} / (c) ---   \makebox[0.2cm]{} / (d) --- \}\\
    \hline 
    \end{tabular}
    \caption{The JSON template for each task evaluation}
    \label{tab:json}
\end{table}

\subsection{Inference Settings}

We use the LLM Evaluation Harness~\cite{gao2021framework} to build our inference and evaluation pipelines. For semantic similarity evaluation, we employ the Chinese-supporting models: chinese-xlnet-base for BERTScore and bart-large-cnn for BARTScore. API-based LLMs like ChatGPT, GPT-4o, Grok3, and ERNIE-3.5 are evaluated via the YunWu platform, while open-source LLMs are assessed on a cluster with 4 A100 GPUs (80GB). We set the
input length limit to 2048 tokens. The maximum generation length is 300 tokens for Q\&A and summarization datasets, and 200 tokens for others.

\section{Results}

\subsection{Model Overall Performance}

\begin{table*}[htb!]
    \centering
    \scriptsize
    \setlength\tabcolsep{0.6pt}
    \renewcommand{\arraystretch}{0.9}
    
    \begin{tabular}{ccc|cccccccccccccccccccc}
    \toprule
    \rotatebox[origin=c]{55}{\textbf{Type}} &  \rotatebox[origin=c]{55}{\textbf{Settings}} & \rotatebox[origin=c]{55}{\textbf{Name}} & \rotatebox[origin=c]{55}{ChatGPT}	&\rotatebox[origin=c]{55}{GPT-4o}	& \rotatebox[origin=c]{55}{MistralV0.3} &	\rotatebox[origin=c]{55}{Gemma}	& \rotatebox[origin=c]{55}{LLaMA3} &	\rotatebox[origin=c]{55}{Bayling2} &	\rotatebox[origin=c]{55}{Grok3} &	\rotatebox[origin=c]{60}{$\text{DeepSeek}_\text{llm}$} &	\rotatebox[origin=c]{55}{Baichuan2}	&\rotatebox[origin=c]{55}{Atom} &	\rotatebox[origin=c]{55}{Qwen2.5}	&\rotatebox[origin=c]{55}{ChnLLaMA3} &	\rotatebox[origin=c]{60}{ERNIE-3.5} &	\rotatebox[origin=c]{60}{ChatGLM3}&	\rotatebox[origin=c]{55}{Yi} &	\rotatebox[origin=c]{55}{GLM4} &	\rotatebox[origin=c]{55}{$\text{DeepSeek}_\text{R1}$}	& \rotatebox[origin=c]{55}{InternLM2.5}&	\rotatebox[origin=c]{55}{YaYi2} \\
    \toprule
    \multirow{6}{*}{\rotatebox[origin=c]{70}{\textbf{Tax Task}}} &\multirow{3}{*}{\rotatebox[origin=c]{50}{\textbf{\makecell{Zero-\\Shot}}}} &
        KM &0.488 & 0.478 & 0.400 & 0.244 & 0.412 & 0.466 & 0.519 & 0.494 & 0.470 & 0.474 & 0.499 & 0.445 & \textbf{0.667} & 0.487 & 0.482 & 0.480 & 0.461 & 0.505 & 0.485 \\ 
        & &KU&0.602 & \textbf{0.637} & 0.277 & 0.085 & 0.236 & 0.409 & 0.579 & 0.341 & 0.243 & 0.370 & 0.538 & 0.312 & 0.599 & 0.427 & 0.449 & 0.515 & 0.455 & 0.273 & 0.307 \\ 
        & &KA&0.415 & 0.472 & 0.114 & 0.127 & 0.183 & 0.202 & \textbf{0.482} & 0.258 & 0.188 & 0.225 & 0.375 & 0.247 & 0.475 & 0.206 & 0.295 & 0.404 & 0.324 & 0.349 & 0.239 \\ 
        \cmidrule{2-22}
                                     &\multirow{3}{*}{\rotatebox[origin=c]{50}{\textbf{\makecell{One-\\Shot}}}} &    
        KM &0.497 & 0.499 & 0.413 & 0.239 & 0.407 & 0.475 & 0.527 & 0.474 & 0.480 & 0.467 & 0.494 & 0.426 & \textbf{0.619} & 0.372 & 0.476 & 0.225 & 0.461 & 0.165 & 0.350 \\ 
        & &KU&0.611 & \textbf{0.653} & 0.429 & 0.393 & 0.182 & 0.350 & 0.591 & 0.488 & 0.491 & 0.439 & 0.453 & 0.460 & 0.633 & 0.341 & 0.485 & 0.257 & 0.520 & 0.348 & 0.257 \\ 
        & &KA&0.425 & 0.464 & 0.217 & 0.188 & 0.197 & 0.271 & 0.468 & 0.297 & 0.286 & 0.269 & 0.386 & 0.281 & \textbf{0.477} & 0.186 & 0.293 & 0.265 & 0.340 & 0.172 & 0.203 \\ 
        \hline

    \multirow{6}{*}{\rotatebox[origin=c]{70}{\textbf{NLP Task}}} &\multirow{3}{*}{\rotatebox[origin=c]{50}{\textbf{\makecell{Zero-\\Shot}}}} &    
        CLS &0.248 & 0.370 & 0.042 & 0.042 & 0.072 & 0.055 & 0.325 & 0.088 & 0.120 & 0.100 & 0.273 & 0.067 & \textbf{0.379} & 0.104 & 0.207 & 0.375 & 0.100 & 0.327 & 0.107 \\ 
        & &GEN&0.626 & 0.644 & 0.266 & 0.193 & 0.327 & 0.426 & 0.645 & 0.458 & 0.316 & 0.423 & 0.547 & 0.436 & \textbf{0.670} & 0.388 & 0.489 & 0.515 & 0.542 & 0.428 & 0.410 \\
        & &REA&0.220 & 0.239 & 0.037 & 0.025 & 0.025 & 0.041 & \textbf{0.308} & 0.041 & 0.033 & 0.023 & 0.194 & 0.048 & 0.169 & 0.156 & 0.020 & 0.223 & 0.164 & 0.041 & 0.045 \\ 
        \cmidrule{2-22}
                                     &\multirow{3}{*}{\rotatebox[origin=c]{50}{\textbf{\makecell{One-\\Shot}}}} & 
        CLS &0.261 & 0.359 & 0.094 & 0.047 & 0.077 & 0.106 & 0.335 & 0.114 & 0.122 & 0.076 & 0.258 & 0.065 & \textbf{0.383} & 0.118 & 0.146 & 0.244 & 0.152 & 0.095 & 0.078 \\ 
        & &GEN&0.639 & 0.652 & 0.428 & 0.377 & 0.321 & 0.447 & 0.646 & 0.550 & 0.533 & 0.514 & 0.533 & 0.526 & \textbf{0.668} & 0.325 & 0.522 & 0.295 & 0.565 & 0.315 & 0.348 \\ 
        & &REA&0.210 & 0.223 & 0.034 & 0.048 & 0.023 & 0.117 & \textbf{0.236} & 0.027 & 0.033 & 0.030 & 0.205 & 0.071 & 0.207 & 0.088 & 0.053 & 0.155 & 0.149 & 0.015 & 0.025 \\ 
    \bottomrule
    \end{tabular}
    \caption{The performance of the 19 LLMs on Tax task (KM, KU, KA) and NLP main task (CLS, GEN, REA).}
    \label{tab:task}
\end{table*}

As illustrated in Figure~\ref{fig:overall}, the following findings can be drawn: 
\textbf{(1) Closed-source LLMs Show Superior Performance:} In all settings, closed-source LLMs like ERNIE-3.5, GPT-4o, Grok3, and ChatGPT consistently rank in the Top 3, with ERNIE-3.5 particularly excelling. This is attributed to their large parameter sizes and extensive training datasets. Notably, ERNIE-3.5 benefits from the integration of domain-specific knowledge, which further enhances its performance on tax-related tasks. 
\textbf{(2) Chinese-oriented LLMs Outperform Multilingual LLMs:} Within the open-source group, Chinese-oriented LLMs pretrained predominantly on Chinese corpora consistently outperform multilingual LLMs. This trend reflects the language-specific nature of tax-related tasks, which often involve Chinese regulatory terminology, institutional jargon, and culturally contextual reasoning. As a result, LLMs with strong Chinese pretraining demonstrate superior alignment with task semantics and domain expressions. 
\textbf{(3) Domain Fine-tuning Doesn't Guarantee Strong Results:} Surprisingly, the only tax-related LLM, YaYi2, performs poorly compared to general LLMs. Despite fine-tuning on tax data, it lags behind some smaller open-source LLMs. We believe this is mainly due to two factors: limited quantity and diversity of tax training data, and significant differences between the fine-tuning content and evaluation tasks. Without sufficient data variety and task alignment, domain-specific fine-tuning may still fail. 
\textbf{(4) Zero-Shot and One-Shot Performance Varies Significantly:} Among the 19 models studied, 11 show improvement in the one-shot setting, effectively utilizing in-context learning with task-specific examples. However, LLMs like GLM4 and InternLM2.5 show performance declines, likely due to longer prompts exceeding attention limits and weaker instruction alignment. These results underscore the impact of prompt design and model architecture on few-shot learning, suggesting that one-shot improvements are not universally applicable.

\subsection{Tax Task Discrepancy}

We observe the following insights from Table~\ref{tab:task}: (1)~\textbf{In KM Task:} ERNIE-3.5 performs best due to its knowledge-enhanced pre-training strategy, while other models face significant bottlenecks. This is largely attributed to the high specialization and frequent updates of tax law knowledge, which means most LLMs are trained on static corpora, potentially leading to outdated knowledge. Additionally, many models show decreased performance in one-shot settings, indicating that a single example may disrupt the model's inherent memory capabilities. (2)~\textbf{In KU Task:} GPT-4o, ChatGPT, and ERNIE-3.5 excel, particularly in comprehending tax-related texts. The locally deployed Qwen2.5 performs relatively well, but other models underperform, highlighting challenges in handling complex semantics and specialized terminology. When comparing zero-shot and one-shot settings, most LLMs show improved performance in one-shot scenarios, as this approach helps them focus on key features. (3)~\textbf{In KA task:} Closed-source LLMs perform well but show the weakest overall results. This is because applying tax knowledge requires LLMs to possess stronger reasoning abilities, which poses greater challenges. The complexity of the tax domain necessitates that models accurately apply policies based on specific scenarios and taxpayer identities. This case-by-case application significantly increases difficulty. Compared to zero-shot setting, most LLMs perform better in one-shot scenarios, as examples help clarify output formats and task objectives, thereby enhancing the stability and accuracy of their responses.

\subsection{NLP Task Discrepancy}

Table~\ref{tab:task} presents the following conclusions: \textbf{(1) CLS Task Highlights Specialized Demands:} The CLS task shows a strong need for expertise, underperforming compared to ~\cite{hu2024no,jiajia2024auditwen} on expected simple multiple-choice questions. This is due to complex tax case analyses and calculations. Models like GPT-4o, Grok3, and ERNIE-3.5 excel, with Chinese-focused LLMs such as InternLM2.5 and GLM4 following closely. \textbf{(2) GEN Task Showcases Proficiency:} LLMs demonstrate a relatively strong capability in the GEN task, where ERNIE-3.5 takes the lead, supported by Grok3 and GPT-4o. Among open-source LLMs deployed locally, Qwen2.5 and $\text{DeepSeek}_\text{R1}$ stand out. Their superior performance is attributed to their larger parameter sets and enhanced support for Chinese, making them well-suited for producing tax-related textual content. \textbf{(3) REA Task Reveals Reasoning Limitations:} The REA task presents significant challenges requiring deep understanding, common sense, and multi-step reasoning. Most LLMs perform poorly, with little improvement in one-shot scenarios due to difficulties in adapting calculation patterns. The complexity of multi-step reasoning, especially with tax formulas, remains a hurdle, even for closed-source large-parameter LLMs like Grok3 and ERNIE-3.5.

\section{Conclusion}
In this study, we introduce TaxPraBen, a pioneering practical benchmark for evaluating LLMs in the Chinese tax domain across 3 cognitive levels: Knowledge Memorization, Understanding, and Application. By incorporating 10 diverse real-world practice tasks derived from 14 datasets, TaxPraBen assesses 19 leading LLMs, revealing notable performance gaps. TaxPraBen's contributions include a rare practical dataset, the establishment of a taxonomy, the design of comprehensive metrics, and a structured evaluation approach, positioning it as a vital resource for advancing research on LLM applications in the Chinese tax domain.

\section*{Limitations}

While TaxPraBen offers a valuable benchmark for LLMs in Chinese tax practice, it has limitations: 
(1)~\textbf{Risk of Data Leakage}: Most datasets are sourced from the internet and public books. While this data collection method is convenient, it also increases the risk of test data leakage. Existing LLMs have been trained on vast amounts of internet data, which means they may inadvertently memorize and reproduce content from these datasets. This situation can lead to models achieving exceptionally high scores during evaluations, but such scores do not accurately reflect the models' reasoning abilities or their effectiveness in real-world applications.
(2)~\textbf{Limitations of Model Parameters}: Current evaluations primarily focus on mainstream LLMs with similar parameters, which introduces limitations when comparing the performance of different models. The lack of comparisons with larger open-source LLMs restricts deeper insights into the impact of various parameters. Models of different scales and architectures may exhibit varying capabilities when handling specific tasks, yet these differences have not been thoroughly explored within the current evaluation framework.
(3)~\textbf{Shortcomings of Automated Evaluation}: Although automated semantic similarity assessment metrics for generative tasks offer convenience, they may not accurately reflect human judgments regarding answer quality. Human evaluators consider a broader range of factors, including relevance, accuracy, and contextual suitability, when assessing answers. Therefore, relying solely on these automated metrics can lead to misjudgments about model performance.

\section*{Ethics Statement}

Given the sensitivity of personal privacy and commercial confidentiality in the tax domain, we conduct a comprehensive data review and ethical considerations for the TaxPraBen benchmark study. We implement strict anonymization measures for the collected data, including the removal of company names and personal identifiers to protect privacy. Additionally, we utilize relevant open-source resources to ensure transparency and ethical compliance. Legal experts evaluate our research methods to ensure adherence to data protection standards. We are committed to maintaining the confidentiality of all stakeholders, ensuring that our research respects personal privacy while making a positive contribution to the field of tax practice. Furthermore, the annotation work is completed voluntarily by individuals who receive no compensation. All annotators are informed of the intended use of the data and give consent to participate.

\bibliographystyle{acl_natbib}
\bibliography{custom}

\clearpage
\appendix

\section{Why important}
\label{sec:important}

\subsection{Benchmark Comparison}
\label{sec:benchmark}

In recent years, numerous LLM benchmarks have been introduced for vertical domains like law (LexGLUE~~\cite{chalkidis2022lexglue}, DISC-LawLLM~\cite{yue2023disclawllm}, LawBench~\cite{fei2024lawbench}, LAiW~\cite{dai2025laiw}), finance (Fineval~\citep{guo2025fineval}, CGCE~\citep{zhang2023cgce}, FLARE~\citep{xie2023pixiu}, FinBen~\citep{xie2024finben}, ICE-FLARE~\citep{hu2024no}), and healthcare (DISC-MedLLM~\cite{bao2023disc}, MedMCQA~\cite{pal2022medmcqa}, CMB~\cite{wang2024cmb}, MedBench~\cite{liu2024medbench}) (see Table~\ref{tab:compare}). Most of the focus on traditional NLP tasks (including MCQ, Q\&A, REA, NR) to assess the transfer of general language understanding to specialized fields. However, they primarily evaluate whether LLMs can answer general questions, often neglecting the assessment of real-world problem-solving and complex decision-making abilities.

\begin{table}[htb!]
    \renewcommand{\thetable}{A1}
    \centering
    \footnotesize
    \setlength{\tabcolsep}{1pt}
    \renewcommand{\arraystretch}{1}
    
    \begin{adjustbox}{max width=0.48\textwidth}
    \begin{tabular}{c|c c |c c c c c|c|c}
    \toprule
    
    \multirow{2}{*}{Name}& \multicolumn{2}{c|}{Method} &
    \multicolumn{5}{c|}{NLP Specific Task} &  \multirow{2}{*}{PRA} &
    \multirow{2}{*}{CoMe} \\
    
    \cmidrule(lr){2-3} \cmidrule(lr){4-7} 
     & HC & IPA & MCQ& Q\&A & CLS& REA &NR &  &  \\

     \midrule
     \multicolumn{10}{c}{Law} \\
     \midrule
     LexGLUE & $\times$ & $\checkmark$ &  $\checkmark$& $\times$& $\checkmark$& $\times$& $\times$& $\times$ & $\times$  \\
     DISC-LawLLM & $\checkmark$ & $\checkmark$ &  $\checkmark$& $\times$& $\times$& $\times$& $\times$& $\times$ & $\times$  \\
     LawBench &  $\times$ & $\checkmark$ &  $\checkmark$& $\checkmark$& $\checkmark$& $\checkmark$& $\times$& $\times$ & $\times$  \\
     LAiW &  $\times$ & $\checkmark$ &  $\times$& $\checkmark$& $\checkmark$& $\checkmark$& $\times$& $\times$ & $\times$  \\
     \midrule
      \multicolumn{10}{c}{Finance} \\
      \midrule
      Fineval & $\times$ & $\checkmark$ &  $\checkmark$& $\checkmark$& $\checkmark$& $\checkmark$& $\times$& $\times$ & $\times$ \\
      CGCE & $\times$ & $\checkmark$ &  $\times$& $\checkmark$& $\times$& $\times$& $\checkmark$& $\times$ & $\times$  \\
      FLARE & $\times$ & $\checkmark$ &  $\times$& $\checkmark$& $\checkmark$& $\checkmark$& $\times$& $\times$ &$\times$  \\
      FinBen&$\times$ & $\checkmark$ &  $\checkmark$& $\checkmark$& $\checkmark$& $\checkmark$& $\times$ &  $\times$& $\times$  \\
      ICE-FLARE &$\times$ & $\checkmark$ &  $\checkmark$& $\checkmark$& $\checkmark$& $\checkmark$& $\times$ &  $\times$ & $\times$  \\
      \midrule
     \multicolumn{10}{c}{Medicine} \\
     \midrule
     DISC-MedLLM & $\checkmark$ & $\checkmark$ &  $\checkmark$& $\checkmark$& $\times$& $\times$& $\times$& $\times$ & $\times$  \\
     MedMCQA & $\checkmark$ & $\times$ &  $\checkmark$& $\times$& $\times$& $\times$& $\times$& $\times$ & $\times$  \\
     CMB & $\checkmark$ & $\checkmark$ &  $\checkmark$& $\checkmark$& $\times$& $\checkmark$& $\times$& $\times$ & $\times$  \\
     MedBench & $\checkmark$ & $\checkmark$ &  $\checkmark$& $\checkmark$& $\checkmark$& $\checkmark$& $\times$& $\checkmark$ & $\times$  \\
     \midrule
    \multicolumn{10}{c}{Taxation} \\
    \midrule
     TaxBen & $\checkmark$ & $\times$ &  $\checkmark$& $\checkmark$& $\checkmark$& $\checkmark$ & $\checkmark$& $\times$& $\times$ \\
     \midrule
     TaxPraBen & $\checkmark$ & $\times$ & $\checkmark$& $\checkmark$& $\checkmark$& $\checkmark$ & $\checkmark$& $\checkmark$& $\checkmark$  \\
    \bottomrule
    \end{tabular}
    \end{adjustbox}
    \caption{Comparing different domain-specific benchmarks, including data construction methods (human construction, HC; improved public data, IPA), task coverage across NLP specific tasks (multiple-choice questions, MCQ; question and answer, Q\&A; text reasoning, REA; numerical reasoning, NR), practical application task (PRA) and the use of combined metrics (CoMe). “$\checkmark$” indicates presence, and “$\times$” indicates absence.}
    \label{tab:compare}
\end{table}

In contrast, TaxPraBen is the first benchmark designed for Chinese tax practice, marking a fundamental shift from task-oriented to capability-oriented evaluation. Its main innovation is expanding the focus from isolated language tasks to practice-oriented, integrated tasks that cover key skills such as legal interpretation, numerical reasoning, and risk assessment, making it much more aligned with real-world professional needs. For instance, traditional TaxBen's NR task is limited to basic questions like "\textit{What is the tax rate for income bracket X}?". In contrast, TaxPraBen's PRA task require a comprehensive question: "\textit{Calculating a taxpayer's final tax liability and identifying potential audit risks based on income sources, deductions, and regional regulations}", thus replicating the full professional workflow and greatly enhancing practical relevance. It highlights TaxPraBen’s innovative use of a combined metric to jointly assess the accuracy of tax amount calculations and the semantic similarity of audit risk identification. Moreover, TaxPraBen introduces a breakthrough in data construction. Unlike most benchmarks that rely on public datasets, TaxPraBen is built entirely on authentic cases from tax professionals, combining manual curation with LLM-based prompt engineering. This ensures high-quality, realistic scenarios and strengthens its practical relevance. 

In summary, as the first truly comprehensive benchmark for Chinese tax practice, TaxPraBen stands out by shifting from task- to capability-oriented evaluation, filling the gap in high-quality tax data, and redefining structured assessment systems. This not only advances the use of LLMs in tax but also provides valuable insights for developing benchmarks in other specialized domains.

\subsection{TaxBen Contrast}
\label{sec:contrast}

In the field of tax artificial intelligence, the application of LLMs is gaining attention. However, due to their inherent opacity, tax experts remain cautious about their real-world applicability. While TaxBen advances benchmarking in this domain, it focuses narrowly on isolated NLP tasks-such as Q\&A, text generation \& classification, and numerical reasoning—neglecting to  evaluate models' practical capabilities in real-world scenarios.

\begin{table}[htb!] 
   \renewcommand{\thetable}{A2}
    \centering
    \footnotesize
    \setlength{\tabcolsep}{1pt}
    \renewcommand{\arraystretch}{1}
  
   \begin{adjustbox}{max width=0.48\textwidth}
   \begin{tabular}{c|c|c|c|c|c} 
    \toprule 
    \multirow{2}{*}{\makecell{\textbf{LLM} \\ \textbf{Type}}}&\multirow{2}{*}{\textbf{Name}} & \multicolumn{2}{c|}{\textbf{Zero-Shot}} & \multicolumn{2}{c}{\textbf{One-Shot}} \\ 
    \cmidrule{3-4} \cmidrule{5-6}
     && \textbf{TaxBen} &\textbf{TaxPraBen} & \textbf{TaxBen} &\textbf{TaxPraBen} \\ 
    \midrule 
    \multirow{3}{*}{$\text{GLS}_{\text{LLM}}$}&ERNIE-3.5& 1 & 1 & 1 & 1   \\ 
    &Grok-3 & 2 & 2  & 2 & 2 \\ 
    &ChatGPT & 3 & 3  & 3 & 3  \\ 
    \midrule 
    \multirow{3}{*}{$\text{COS}_{\text{LLM}}$}&ChatGLM3& 8 &  12 & 13  & 15   \\ 
    & $\text{DeepSeek}_\text{llm}$ & 12 & 9 & 8  & 6 \\  
    & Atom & 10 & 11 & 9  & 10 \\ 
    \midrule 
    $\text{Tax}_{\text{LLM}}$&YaYi2& 13 & 13 & 16 & 16 \\
    \bottomrule 
  \end{tabular} 
  \end{adjustbox}
   \vskip 2pt
    \begin{flushleft}
     \scriptsize
     \textit{\textbf{Note}: To ensure fair model rankings, we consider only jointly evaluated  LLMs.}
     \end{flushleft}
  \caption{Comparison of the evaluation ranking results of various LLMs between TaxPraBen and TaxBen. General closed-source LLMs: $\text{GLS}_{\text{LLM}}$, Chinese open-source LLMs: $\text{COS}_{\text{LLM}}$, Tax-related LLM: $\text{TAX}_{\text{LLM}}$.} 
  \label{tab:taxben}
\end{table}

As shown in Table~\ref{tab:taxben}, apart from closed-source large-parameter LLMs (including ERNIE‑3.5, Grok‑3, and ChatGPT), several LLMs like ChatGLM3 and Atom—exhibit inflated rankings on the simpler NLP tasks covered by TaxBen,  but see significant declines in TaxPraBen, which includes real-world application tasks. The notable discrepancy in model rankings between TaxPraBen and TaxBen highlights the limitations of TaxBen in addressing genuine practical tasks. While TaxBen does include numerical reasoning and question-answering tasks, can the practical capabilities of the models be adequately reflected solely by averaging the metrics from these two tasks? Considering that real-world practical tasks require both semantic matching and computational reasoning abilities, the answer is clearly no. Further analysis, as illustrated in Table~\ref{tab:differ}, supports this conclusion. For instance, models like InternLM2.5, Yi, and ERNIE‑3.5 perform excellently and rank highly on the numerical‑reasoning (NR data) and semantic‑matching (Q\&A data) tasks in TaxBen. Yet, on the TaxPlan task in TaxPraBen—which demands the integration of both capabilities—their metrics decline markedly, with most rankings dropping to 16th-18th ranking positions, showing an average decline of 5–11 positions (a total of 18). This clearly demonstrates that the practical ability of LLMs cannot be represented simply by weighting the evaluation metrics of individual NLP subtasks. 

\begin{table}[htb!] 
   \renewcommand{\thetable}{A3}
    \centering
    \footnotesize
    \setlength{\tabcolsep}{1pt}
    \renewcommand{\arraystretch}{1}
    
   \begin{adjustbox}{max width=0.48\textwidth}
  \begin{tabular}{c|cc|cc|c|cc|c} 
  \toprule 
    \multirow{3}{*}{\textbf{LLM}}&\multicolumn{5}{c|}{\textbf{TaxBen}} & \multicolumn{3}{c}{\textbf{TaxPraBen}} \\ 
    \cmidrule{2-6} \cmidrule{7-9} 
    & \multicolumn{2}{c|}{\textbf{NR Data}} &\multicolumn{2}{c|}{\textbf{Q\&A Data}} & \textbf{Avg} &\multicolumn{2}{c|}{\textbf{TaxPlan}} & \multirow{2}{*}{R($\downarrow$)}\\ 
    &  \textbf{Avg.Val} &\textbf{R}  & \textbf{Avg.Val} & \textbf{R}  & \textbf{.R} &\textbf{Avg.Val} & \textbf{R}  &  \\
    \midrule
    \multicolumn{9}{c}{Zero-Shot} \\
    \midrule 
    InternLM2.5 & 0.042 & 5 &  0.520 & 4  & 5 & 0.039 & 16& 11\\
    Yi & 0.024 &7  & 0.495 & 9  & 8 & 0.016 & 18  & 10 \\
    ERNIE-3.5 & 0.076 & 2  &  0.559 & 1  & 2 &0.263 & 7& 5 \\
    \midrule
    \multicolumn{9}{c}{One-Shot} \\
    \midrule
    InternLM2.5 & 0.000& 7  &0.529 & 4  & 6 & 0.031 & 18 & 12 \\
    ERNIE-3.5 & 0.002 & 2  & 0.487 & 9  & 6 &0.052 & 15 & 9 \\
    Qwen2.5& 0.000 & 7  &0.421 & 17  & 12 & 0.045 & 17 & 5 \\
   \bottomrule 
  \end{tabular} 
  \end{adjustbox}
  \vskip 2pt
    \begin{flushleft}
     \scriptsize
     \textit{\textbf{Note}: To ensure fair model rankings, we consider only jointly evaluated  LLMs and round the averag of the metric rankings (Avg.R) for NR and Q\&A data.}
     \end{flushleft}
    \caption{Comparison of metric rankings on TaxBen (numerical reasoning (NR data)+semantic matching (Q\&A data)) and TaxPraBen (tax practice (TaxPlan)). Metric Ranking: R, Average Value. "Avg.R" denotes average of the metric rankings for NR and Q\&A data.}   
  \label{tab:differ}
\end{table}

These findings further reveal that existing domain‑specific benchmarks, with their emphasis on NLP‑task evaluation, exhibit significant shortcomings in assessing real‑world applicability. Thus, we propose TaxPraBen, the first benchmark designed to evaluate LLMs in Chinese tax‑practice scenarios. TaxPraBen incorporates 10 datasets (see Table~\ref{tab:application}) covering real‑world tasks such as tax‑risk prevention, tax‑inspection analysis, and tax‑strategy planning, thereby enabling a systematic assessment of models’ comprehensive applied ability in complex tax‑practice settings.

\section{Taxonomy Motivation}
\label{sec:Taxonomy}

TaxPraBen employs Bloom's Taxonomy as its evaluation system, systematically assessing the mastery of tax knowledge in LLMs through a hierarchical cognitive ability structure. This framework categorizes tax competencies into 3 progressive dimensions: \textit{Knowledge Memorization (KM)}, K\textit{nowledge Understanding (KU)}, and \textit{Knowledge Application (KA)}. In the specialized tax domain, this tiered approach first requires models to accurately recall legal provisions (e.g., TaxRecite tests verbatim clause reproduction), then interpret policy implications (e.g., TaxRead evaluates reading comprehension of regulations), and finally apply knowledge practically (e.g., TaxPlan examines tax planning strategy design). 
Bloom's Taxonomy offers 3 key advantages: (1) Clear hierarchical standards enable precise quantification of knowledge depth; (2) Reveal inter-level relationships helps diagnose fundamental skill gaps; (3) Adapt to tax practice complexity, it identifies cross-level weaknesses. Compared to traditional one-dimensional difficulty grading, this structured method provides targeted optimization guidance (e.g., enhancing legal memorization or tax reasoning abilities). Its adoption marks a paradigm shift from evaluating task completion to knowledge mastery.

\begin{table*}[htb!] 
    \renewcommand{\thetable}{C2}
    \centering
    \footnotesize
    \setlength{\tabcolsep}{15pt}
    \renewcommand{\arraystretch}{1.1}
    
    \begin{tabular}{c|l}
    \toprule 
    \textbf{Dataset Source} & \multicolumn{1}{c}{\textbf{URL}} \\ 
     
    \hline 
    Tax House website& \url{https://www.shui5.cn/}  \\ 
    \hline 
    Tax Practice Cases & \url{https://wxredian.com/art?id=ecf5c4f6a9986a6eda47f48ba7ed5e74}  \\
    \hline 
    Certified Tax Agent Exam & \url{http://www.canet.com.cn/sws/zhinan/kaoshijiaocai.html}  \\ 
    \hline 
    Opinion Monitoring System & \url{https://open-yuqing.stonedt.com/}  \\
    \hline 
    Taxation Administration Website & \url{https://www.chinatax.gov.cn/}  \\
    \bottomrule 
    \end{tabular} 
    \caption{The original source along with the URL of our used datasets.} 
    \label{tab:source}
\end{table*} 

\begin{table*}[htb!] 
    \renewcommand{\thetable}{C3}
    \centering
    
    \vspace{-5pt}
    \makeatletter
    \def\@captype{figure} 
    \makeatother
    \begin{minipage}{0.33\textwidth}  
        \centering
        \raisebox{-10pt}{\includegraphics[width=\textwidth]{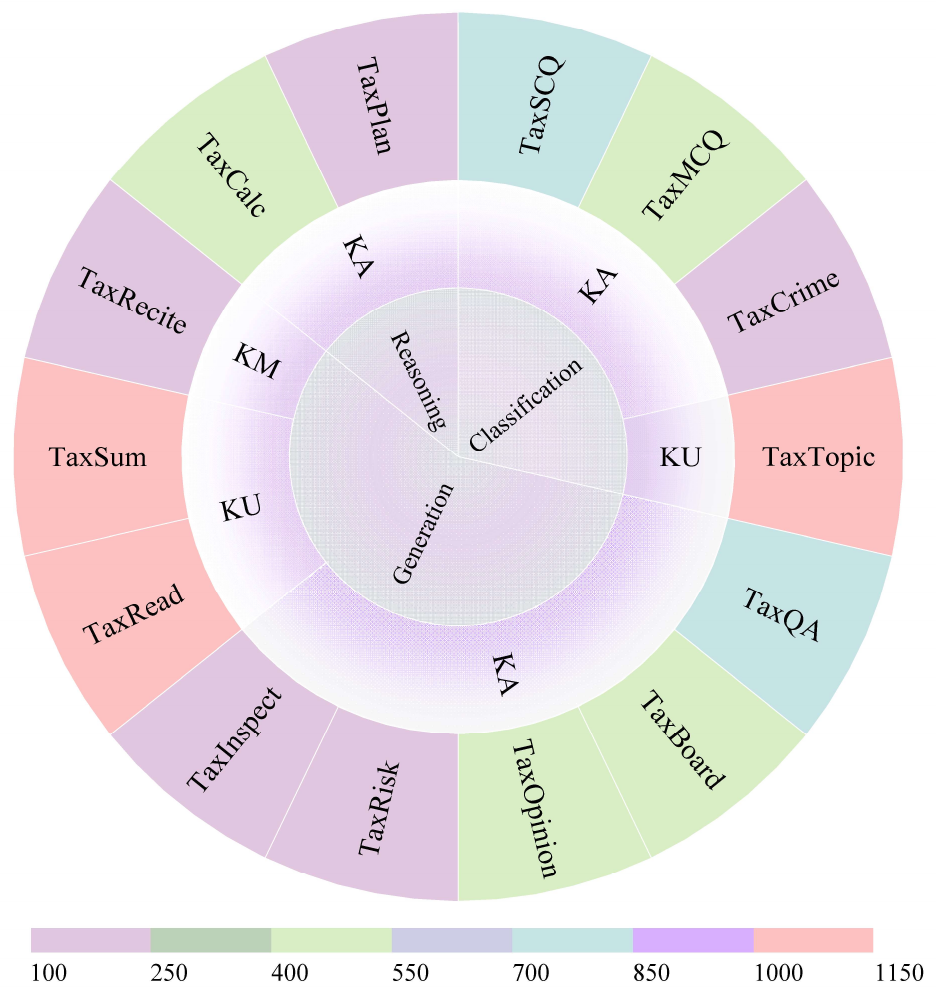}}
        \label{fig:datasets}
    \end{minipage}
    \makeatletter
    \def\@captype{table} 
    \makeatother
    \begin{minipage}{0.65\textwidth}  
        \centering
        \setlength\tabcolsep{0.5pt} 
        \renewcommand{\arraystretch}{0.9} 
        \resizebox{\textwidth}{!}{
        \begin{tabular}{c|ccc|ccc|ccc|ccc}
        \toprule
        \multirow{2}{*}{\textbf{Dataset}} & \multicolumn{3}{c|}{\textbf{Zero-Shot (Prompt)}} & \multicolumn{3}{c|}{\textbf{One-Shot (Prompt)}} & \multicolumn{3}{c|}{\textbf{Input Text}} & \multicolumn{3}{c}{\textbf{Label}} \\
        & \textbf{Max} & \textbf{Min} & \textbf{Avg} & \textbf{Max} & \textbf{Min} & \textbf{Avg} & \textbf{Max} & \textbf{Min} & \textbf{Avg} & \textbf{Max} & \textbf{Min} & \textbf{Avg} \\
        \hline
        TaxRecite & 42 & 37 & 39.06 & 592 & 109 & 206.80 & 69 & 33 & 40.11 & 514 & 17 & 115.05 \\
        \hline
        TaxSum & 38 & 35 & 36.37 & 1469 & 273 & 573.60 & 1335 & 130 & 405.50 & 236 & 14 & 115.12 \\
        TaxTopic & 130 & 125 & 128.02 & 207 & 120 & 150.50 & 100 & 6 & 37.26 & 1 & 1 & 1.00 \\
        TaxRead & 56 & 49 & 53.82 & 1812 & 221 & 566.30 & 2066 & 125 & 521.85 & 286 & 1 & 15.44 \\
        \hline
        TaxCalc & 86 & 65 & 69.80 & 1383 & 327 & 636.40 & 1266 & 216 & 535.48 & 9 & 4 & 5.84 \\
        TaxSCQ & 68 & 55 & 59.93 & 494 & 111 & 218.50 & 328 & 35 & 129.10 & 1 & 1 & 1.00 \\
        TaxMCQ & 146 & 136 & 141.14 & 587 & 153 & 319.50 & 469 & 42 & 140.94 & 4 & 2 & 3.15 \\
        TaxQA & 38 & 29 & 34.75 & 724 & 60 & 126.40 & 149 & 7 & 30.03 & 532 & 2 & 46.78 \\
        TaxBoard & 48 & 41 & 45.28 & 1081 & 109 & 299.10 & 474 & 23 & 99.58 & 814 & 11 & 143.32 \\
        TaxCrime & 219 & 201 & 208.72 & 521 & 303 & 387.13 & 299 & 24 & 140.60 & 2 & 1 & 1.13 \\
        TaxOpinion & 53 & 38 & 43.82 & 2235 & 825 & 1386.05 & 1646 & 502 & 987.79 & 445 & 163 & 298.24 \\
        TaxRisk & 112 & 68 & 84.88 & 1873 & 329 & 1110.60 & 1597 & 153 & 923.00 & 302 & 40 & 110.76 \\
        TaxInspect & 198 & 171 & 179.13 & 1603 & 411 & 930.92 & 1149 & 156 & 627.77 & 213 & 32 & 105.22 \\
        TaxPlan & 354 & 83 & 150.65 & 1071 & 590 & 719.80 & 432 & 57 & 150.15 & 394 & 65 & 141.91 \\
        \bottomrule
        \end{tabular}%
        }
    \end{minipage}
    \caption{TaxPraBen includes 3 NLP and tax tasks across 14 datasets ranging from 200 to 1,000 entries, along with statistics on the maximum (Max), minimum (Min), and average (Avg) lengths of prompts and input/labels.}
    \label{tab:count}
\end{table*}

\section{Instruction Details}
\label{sec:Dataset}

\subsection{Collection Process}
\label{sec:process}

The definition and data collection details of the datasets for the three-level tax tasks in our proposed TaxPraBen benchmark are shown in Table~\ref{tab:collection}. These datasets span three core tax task levels (KM: Knowledge Memorization, KU: Knowledge Understanding, KA: Knowledge Application) with 14 specific datasets. Each dataset details its task definition and collection sources, official tax documents, the website of the State Taxation Administration, Tax House articles, certified tax agent exam materials, and real-world tax practice cases, with some leveraging ChatGPT-3.5 as an auxiliary tool for text compression or question-answer generation. For the convenience of researchers, Table~\ref{tab:source} lists the original sources of our reconstructed dataset.

\subsection{Dataset Refinement}
\label{sec:annotation}

In Table~\ref{tab:dataset}, aside from the TaxTopic, TaxCalc, TaxSCQ, and TaxMCQ datasets—which have definitive answers—all other tax-related datasets constructed via a human–ChatGPT collaborative approach require manual verification. Unlike other benchmarks that directly use data generated with LLM assistance, we recognize that LLM-based prompt engineering during the construction phase can introduce hallucinations~\cite{farquhar2024detecting}, such as fabricated figures or descriptions, which may add noise or even incorrect labels to the data. This issue is particularly evident in the TaxRisk, TaxInSpect, and TaxPlan datasets. Despite multiple attempts to refine the prompts, obtaining structured outputs that meet our requirements remains challenging. Consequently, we manually review and correct all data generated with LLM involvement.

The refinement process is carried out by a dedicated annotation team of 50 students. While similar in composition to the prompt design team—which consists mainly of graduate students—this group also includes a significant number of undergraduates due to the scale of the annotation task. All team members are native Chinese speakers with diverse academic backgrounds in computer science, finance, taxation, statistics, and related fields. Computer science specialists are responsible for explaining annotation guidelines and conducting random sampling checks, while other professionals, leveraging their domain expertise in taxation, perform the actual data labeling. This collaborative approach ensures both technical rigor and domain accuracy, thereby enhancing the robustness and reliability of the dataset. Each annotator initially contributes 20 distinct data entries. After the first round, the computer science team randomly samples 5 entries from each annotator to verify compliance with the requirements. Through several rounds of feedback and guidance, all annotators eventually meet the quality standards. For annotation, we use the EasyDataSet~\footnote{\url{https://docs.easy-dataset.com/}}~\cite{miao2025easy} platform, which is designed for constructing instruction sets for LLMs. The platform supports manual correction, allows precise Q\&A labeling, and helps reduce label duplication. We provide the annotation demonstration examples in Figure~\ref{fig:screenshot}.

\subsection{Dataset Statistics}
\label{sec:statistics}

We present detailed statistical information of our TaxPraBen dataset in Table~\ref{tab:count}. Specifically, we analyze the data length distribution. The maximum input is 2,066 characters (≈510–520 tokens), and the longest target is 814 characters (≈200–300 tokens)—values that closely align with our experimental settings. Accordingly, we cap the maximum input at 2,048 tokens. For Q\&A and summarization datasets, we allow up to 300 generated tokens; for all others, the limit is 200 tokens, with prompts explicitly requesting concise outputs. These choices are well-founded: (1) virtually every sample fits within the model’s context window, so no information is lost to truncation; (2) for generative tasks, the required outputs rarely exceed 300 tokens, confirming that the limits strike a balance between informativeness and brevity. In short, the statistics corroborate the soundness of our configuration.

\subsection{Dataset Instance}
To enhance transparency and ensure reproducibility, we present the instance formats and representative prompts utilized in TaxPraBen. For each dataset, we provide an example table that illustrates the evaluation input template—such as zero-shot or one-shot settings—along with a concrete example. Each example includes the input text, candidate options (or output constraints), and the gold label. To maintain clarity, examples are grouped by dataset and presented in separate tables, as shown in Tables~\ref{tab:TaxRecite}–\ref{tab:TaxPlan}. These datasets are carefully curated by domain experts, are cost-effective, and have no commercial usage restrictions. 

\textbf{\textit{Knowledge Memorization (KM)}}: contains 1 dataset. (1)~\textbf{TaxRecite}: consists of "provision cue–statutory text" pairs, designed to evaluate the model’s ability to reproduce tax-law provisions verbatim and maintain clause-level consistency. \textbf{\textit{Knowledge Understanding (KU)}}: contains 3 datasets. (1)~\textbf{TaxSum}: includes tax-related news articles and reference summaries; the model is required to generate a summary of no more than 300 words. (2)~\textbf{TaxTopic}: a single-label classification task based on news titles. The label space contains 10 categories: A. Financial literacy news; B. Tax and finance forms; C. Local regulations; D. Regulation interpretation; E. Taxable income adjustments; F. Tax assessment; G. Tax incentives; H. Tax planning; I. Tax Q\&A; J. National regulations. (3)~\textbf{TaxRead}: comprises article–question–reference answer triples, focusing on extractive reading comprehension grounded in the provided text. \textbf{\textit{Knowledge Application (KA)}}: contains 10 datasets. (1)~\textbf{TaxCalc}: features tax computation scenarios and requires outputs with constrained key numerical fields. (2)~\textbf{TaxSCQ}: single-choice questions with four candidate options; the output is restricted to a single letter in \{A, B, C, D\}. (3)~\textbf{TaxMCQ}: multiple-choice questions with five candidate options; the output is a set of letters drawn from \{A, B, C, D, E\} (at least two options, reported in alphabetical order). (4) TaxQA: tax consultation question–answer samples, requiring the model to generate standardized responses. (5)~\textbf{TaxBoard}: government message-board Q\&A; the model must provide an answer and state the corresponding policy basis. (6)~\textbf{TaxCrime}: a Criminal Law article classification task grounded in tax-related criminal fact descriptions. The label space contains 13 categories: A. Article 201 of the Criminal Law; B. Article 203 of the Criminal Law; C. Article 204 of the Criminal Law; D. Article 205 of the Criminal Law; E. Article 206 of the Criminal Law; F. Article 207 of the Criminal Law; G. Article 208 of the Criminal Law; H. Article 209 of the Criminal Law; I. Article 210 of the Criminal Law; J. Article 211 of the Criminal Law; K. Article 212 of the Criminal Law; L. Article 227 of the Criminal Law; M. Article 264 of the Criminal Law. (7)~\textbf{TaxOpinion}: includes tax-related public opinion texts and reference summaries; the model generates a summary of approximately 300 words. (8)~\textbf{TaxRisk}: risk management and control texts with structured extraction targets, namely "Risks Encountered" and "Corresponding Mitigation Measures". (9)~\textbf{TaxInspect}: an information extraction task over tax inspection cases, requiring the model to output "Criminal Act", "Charged Offense", "Penalty Outcome", the Charged Offense is restricted to one of seven categories:  Crime of holding forged invoices; Crime of illegally selling special VAT invoices; Crime of illegally purchasing special VAT invoices; Crime of purchasing forged special VAT invoices; Crime of fraudulently obtaining export tax refunds; Crime of tax evasion; Crime of falsely issuing invoices. (10) \textbf{TaxPlan}: A structured tax-planning generation task where only the core rationale is textual and all other key fields are numeric, enabling automatic evaluation.

\begin{table*}[htb]
    \renewcommand{\thetable}{C18}
    \centering
    \footnotesize
    \setlength\tabcolsep{1.5pt}
    \renewcommand{\arraystretch}{1}
    
    \begin{tabular}{c|m{13cm}<{\centering}}
    \toprule
    \textbf{Metrics} & \multicolumn{1}{c}{\textbf{Description}} \\
    \hline
    Accuracy (ACC) & Measure if the response meets the target answer provided by the query, avoiding errors and bias. \\
    \hline
    Naturalness (NAT) & Assesses whether the response description is fluent, natural, and consistent with human expression. \\ 
    \hline
    Informativeness (INF) & Evaluate if the response description includes enough key information and fully answers the question. \\
    \bottomrule
    \end{tabular} 
    \caption{Briefly description of the three manual metrics (ACC,NAT,INF) for prompt quality evaluation.}
    \label{tab:questionnaire}
\end{table*}

\begin{table}[htb!]
  \renewcommand{\thetable}{C22}
  \centering
   \footnotesize
    \setlength\tabcolsep{0.5pt}
    \renewcommand{\arraystretch}{1}  
    
    \begin{adjustbox}{max width=0.48\textwidth}
    \begin{tabular}{cc|c|c|c|c}
    \toprule
    \multirow{2}{*}{\textbf{Task}} &\multirow{2}{*}{\textbf{Dataset}}  & \multicolumn{4}{c}{\textbf{Prompt Consistency Analysis}} \\ 
    \cmidrule(lr){3-6}
     && \textbf{Total} & \textbf{Retained} & \textbf{Fleiss's $\kappa$} & \textbf{Krippendorff's $\alpha$} \\
    \midrule
    KM &TaxRecite & 20 & 10 & 0.842 & 0.867 \\
    \hline
    \multirow{4}{*}{KU} &TaxSum & 20 & 12 & 0.824 & 0.856  \\
    &TaxSum & 20 & 13 & 0.833 & 0.872  \\
    &TaxTopic & 20 & 11 & 0.843 &0.891  \\
    &TaxRead & 20 & 16 & 0.837 &0.861 \\
    \hline
    \multirow{10}{*}{KA} &TaxCalc & 20 & 13 & 0.856 &0.872  \\
    &TaxSCQ & 20 & 15 & 0.828 & 0.865 \\
    &TaxMCQ & 20 & 16 & 0.853 &0.857  \\
    &TaxQA & 20 & 13 & 0.819 &0.842  \\
    &TaxBoard & 20 & 15 & 0.832 & 0.873 \\
    &TaxCrime & 20 & 14 & 0.851 &0.863 \\
    &TaxOpinion & 20 & 16 & 0.876 &0.881  \\
    &TaxRisk & 20 & 18 & 0.846 &0.858  \\
    &TaxInspect & 20 & 15 & 0.862 &0.871 \\
    &TaxPlan & 20 & 14 & 0.846 & 0.893  \\
    \bottomrule
    \end{tabular}%
    \end{adjustbox}
    \caption{Prompt evaluation consistency analysis: average scores for Fleiss' $\kappa$ and Krippendorff's $\alpha$.}
    \label{tab:agreement}
\end{table}

All datasets are converted into instruction sets in JSONL format, with the following structure:
\begin{lstlisting}
{
 id:[integer] unique sample ID
 query:[string] input question & prompt
 text:[string] input question content
 answer:[string] expected response
}
\end{lstlisting}

In addition, for all other classification datasets (including TaxSCQ, TaxTopic, and TaxCrime), the instructions contain two extra fields.
\begin{lstlisting}
{
 choices:[list] a list of responses
 gold:[integer] the ideal target response
}
\end{lstlisting}

\subsection{Prompt Annotation}
\label{sec:prompt}


To guide LLMs in producing outputs in a specified format for evaluation, a rigorous process of prompt selection, adaptation, and design is required. Given the specialized knowledge and complexity involved in designing prompts for the tax domain, expertise in the field is essential, and it is recognized that different LLMs exhibit significant variation in their adaptability to prompts. Therefore, a customized set of prompt samples is developed for each dataset to ensure the generalizability and comparability of prompt performance across different models. This procedure consists of two phases: annotation and evaluation.
(1)~\textbf{Annotation Phase}: We assemble a team of five instructors from the faculties of finance and economics, who specialize in auditing, tax administration, tax planning, tax agency, tax inspection, and public administration. For each dataset, at least two instructors collaborate—one is responsible for drafting diverse prompts, and the other reviews them to ensure the professionalism of the tax scenario descriptions and the accuracy of terminology. The team provides 20 prompt examples for each dataset. To ensure consistency in the quality of prompt annotation, we address the issue of variability in LLM adaptability caused by differences in how experts understand the data and their individual annotation styles.
(2)~\textbf{Evaluation Phase}: We form a professional annotation and evaluation team consisting of eight undergraduate students and two graduate students. These students have backgrounds in economics and taxation and have demonstrated strong academic performance, ensuring both expertise and accountability. Each participant evaluates 5 prompts using a questionnaire that assesses 3 metrics: Accuracy (ACC), Naturalness (NAT), and Informativeness (INF), as shown in Table~\ref{tab:questionnaire}. The evaluation options are divided into four levels with corresponding scores: "Strongly Disagree (0)", "Disagree (1)", "Agree (2)", and "Strongly Agree (3)". Examples of these manual scores are shown in Tables~\ref{tab:prompt1}-\ref{tab:prompt3}. All written prompts are tested via the ChatALL~\footnote{\url{https://github.com/ai-shifu/ChatALL}} platform using APIs of randomly selected LLMs, like OpenAI's ChatGPT, and Alibaba's Qwen.

To ensure the reliability and quality of our evaluation annotation process, we calculate two key inter-annotator agreement scores in each evaluation round: Fleiss' Kappa ($\kappa$) and Krippendorff's Alpha ($\alpha$)~\citep{yu2025open,peng2025plutus}. Fleiss' Kappa measures the consistency among multiple raters, while Krippendorff's Alpha accounts for imbalances in category distribution. We  individually assess the Kappa and Alpha scores for 3 manual metrics (ACC, NAT, and INF), averaging these scores to derive the final inter-annotator agreement score. Regular training sessions and discussions on scoring guidelines help minimize subjective bias. As summarized in Table~\ref{tab:agreement}, after 5 rounds of cross-evaluation, the average inter-annotator agreement scores for each prompt set across different raters range from 0.8 to 1, demonstrating the robustness and quality of the manual assessments. To ensure consistency in prompt evaluation, we retain only those prompts with average scores exceeding 2 (AVG>2) as high-quality paired prompts. Examples of human-annotated prompts and their corresponding English translations for all datasets are presented in Table~\ref{tab:prompt}.

To enhance the adaptability of LLMs to prompts, each data sample is paired with a randomly selected  high-quality prompt during evaluation.

\section{Scalable Promotion}
\label{sec:scalable}

Our TaxPraBen  proposed the structured evaluation approach can be naturally extended to specialized fields like law, finance, and medicine.
(1)~\textbf{Legal}: For tasks like verdict summarization and sentencing prediction, our framework pairs BERTScore to validate legal reasoning soundness with accuracy metrics to verify numerical outcomes including fines or sentence lengths, enabling reliable, controllable legal AI evaluation.
(2)~\textbf{Finance}: For audit reports and risk analysis workflows, it validates the rationality of qualitative risk assessments while cross-checking the accuracy of transaction amounts and discrepancy ratios, standardizing performance comparisons across models for high-stakes financial automation. 
(3)~\textbf{Medicine}: It supports granular assessment of hybrid clinical outputs, evaluating diagnostic narrative quality via semantic metrics while validating the precision of vitals, lab values and other quantitative biomedical measurements, ideal for discharge summary automation and diagnostic support tools. This structured paradigm establishes TaxPraBen as scalable evaluation framework for compliance-critical domains, enabling holistic assessment of both interpretive reasoning quality and numerical precision for LLM hybrid outputs. 
Table~\ref{tab:fileds} presents adapted practice case studies for these extended domains.

\section{Benchmark Leaderboard}

To enable transparent and standardized comparison of tax-oriented LLMs, we build the TaxPraBen open leaderboard, which supports model submission, automatic evaluation, and result visualization. The TaxPraBen open leaderboard~\footnote{\url{https://huggingface.co/spaces/hzkkk/TaxPraBen}} and its representative interface are shown in Figure~\ref{fig:leaderboard}.

\textbf{(1) Platform design.} We adopt a modular architecture to continuously add tasks and metrics while keeping results comparable across versions, and provide an end-to-end pipeline from submission to display.
\textbf{(2) Submission protocol.} Submissions should include necessary metadata and reproducible settings, and comply with privacy and governance requirements, avoiding sensitive data and improper behaviors.
\textbf{(3) Leaderboard Evaluation.} The leaderboard covers the main tasks and metrics of TaxPraBen, reporting overall and per-task scores with filtering and sorting for quick diagnosis of strengths and weaknesses.
\textbf{(4) Community Collaboration.} The platform accepts new models and benchmark extensions, and we maintain versioned tasks and rules to ensure long-term traceability and comparability.

Overall, TaxPraBen offers a scalable, open, and transparent platform for evaluating LLMs in Chinese tax practice. It enables model submissions, interactive performance comparisons, community contributions, and iterative improvements, providing a reliable reference for researchers and practitioners while bridging the gap between academic research and real-world applications in tax domain.

\section{Evaluation Details}
\label{sec:details}

\subsection{Structured Alignment}
\label{sec:alignment}

To evaluate the three typical practical cases (TaxRisk, TaxInSpect, and TaxPlan), we require LLMs to produce JSON responses that strictly follow a predefined structure with specified fields. However, even with carefully designed prompts that show high consistency in human evaluations, it remains challenging to guide all LLMs to produce fully compliant outputs. Thus, we further employ ChatGPT, combined with the prompts shown in Table~\ref{tab:prompts}, to process the non-compliant outputs: removing redundant text and extracting all key fields. Examples of raw outputs and their standardized structured versions are provided in Table~\ref{tab:convert}.

\subsection{Models Introduction}
\label{sec:models}

We evaluate 19 LLMs, organized into three groups: \textbf{(1) 7 Multilingual General LLMs}, namely ChatGPT-3.5-Turbo-1106\&GPT-4o~~\citep{brown2020language} (OpenAI), Mistral-v0.3~\cite{jiang2023mistral7b} (Mistral AI), Gemma~\cite{team2024gemma} (Google), BayLing2~\cite{zhang2024bayling} (Chinese Academy of Sciences), LLaMA3~\cite{dubey2024llama} (Meta), and Grok-3~\footnote{\url{https://grok.com/}} (xAI), covering both open-source and commercial models; \textbf{(2) 11 Chinese-oriented LLMs}, including Baichuan2~\cite{yang2023baichuan} (Baichuan AI), DeepSeek llm\&R1~\cite{guo2025deepseek} (DeepSeek), Qwen2.5~\cite{yang2024qwen2} (Alibaba), ChatGLM3\&GLM4~\cite{glm2024chatglm} (Zhipu AI), ChineseLLaMA3~\cite{cui2023alpaca} (Community), YI~\cite{young2024yi} (01 AI), ERNIE-3.5-8K (Baidu), Atom~\cite{zhao2024atom} (AtomEcho), and InternLM2.5~\cite{cai2024internlm2} (Shanghai AI Lab), pretrained on Chinese text and generally surpass multilingual LLMs in Chinese tasks; \textbf{(3) 1 Tax-related LLM}, YaYi2~\cite{luo2023yayi} (Zhongke Wenge), fine-tuned with tax-related corpora to enhance its understanding of Chi-nese tax knowledge.  Their respective descriptions are as follows. 
\textbf{ChatGPT-3.5}: Developed by OpenAI with the Turbo-1106 version, this LLM  features 175B parameters and excels in conversational skills, widely used in NLP tasks. 
\textbf{GPT-4o}: A multimodal LLM  by OpenAI, capable of text, image, and voice processing, known for fast responses and strong conversational abilities, accessible via API.
\textbf{Mistral-V0.3}: Developed by Mistral AI, this 7B parameter LLM  emphasizes efficient reasoning and is suitable for local deployment. \textbf{Gemma}: A 7B parameter model by Google, designed for multilingual processing in resource-constrained environments, lightweight and practical. 
\textbf{LLaMA3}: Developed by Meta, this LLM  features 8B parameters and excels in multilingual generation for complex tasks. \textbf{Bayling2}: Based on LLAMA2 architecture, this 7B parameter LLM  from the Chinese Academy of Sciences focuses on multilingual alignment, particularly for low-resource languages. \textbf{Grok3}: Developed by xAI, this model boasts 1200B parameters and is API-accessible, capable of handling complex tasks.
\textbf{DeepSeek-llm}: A 7B parameter model by DeepSeek, optimized for Chinese processing and suitable for localized applications.
\textbf{Baichuan2}: Developed by Baichuan AI, this 7B parameter LLM  specializes in Chinese content generation with stable performance and open weight access.
\textbf{Atom}: Fine-tuned from LLAMA2-7B, this 7B parameter LLM is optimized for Chinese tasks by a Chinese research team.
\textbf{Qwen2.5}: Developed by Alibaba Cloud, this 7B parameter LLM supports Chinese and multilingual processing with excellent performance and weight access.
\textbf{ChineseLLaMA3 (ChnLLaMA3)}: A community-developed variant fine-tuned on Llama-3-8B, featuring 8B parameters and multilingual capabilities.
\textbf{ERNIE-3.5-8K}: Developed by Baidu, this model excels in Chinese language understanding with about 1,000B parameters, ideal for search and Q\&A tasks, accessible via API.
\textbf{ChatGLM3}: Optimized for Chinese conversation, this 6B parameter model by Zhipu AI offers weight access for chat applications.
\textbf{YI}: A 6B parameter model from 01.AI, excelling in Chinese processing and code generation, suitable for development scenarios with weight access.
\textbf{GLM4}: Developed by Zhipu AI, this 9B parameter model emphasizes high-precision performance in Chinese tasks for professional applications.
\textbf{DeepSeek-R1}: A 7B parameter LLM by DeepSeek, known for powerful reasoning capabilities, addressing complex problem-solving needs.
\textbf{InternLM2.5}: Developed by the Shanghai Artificial Intelligence Laboratory, this 7B parameter model excels in both Chinese and multilingual tasks, applicable across various scenarios.
\textbf{YaYi2}: A 30B parameter LLM by Zhongke WenGe, offering weight access, designed as a general-purpose LLM with some tax-related fine-tuning for professional applications.

\subsection{Metric Computation}
\label{sec:computation}

Below is the detailed explanations of our used evaluation metrics. 

  \textbf{Accuracy (ACC)}~\cite{goutte2005probabilistic}：It measures text classification performance by the proportion of correctly predicted samples. Defined as:
  \[
    \mathrm{ACC}=\frac{TP+TN}{TP+TN+FP+FN}
  \]
  where $TP$ (True Positive) is the count of correctly identified positive samples, $TN$ (True Negative) is the count of correctly identified negatives, $FP$(False Positive) is the count of negatives misclassified as positives, and $FN$ (False Negative)  is the count of positives misclassified as negatives.
  
  \textbf{F1 Score}~\cite{goutte2005probabilistic}: It is the harmonic mean of Precision and Recall. It is useful for evaluating model performance in sentiment analysis, especially with class imbalance. Defined as:
  \[
    \mathrm{F1}=\frac{2\cdot \mathrm{Precision}\cdot \mathrm{Recall}}{\mathrm{Precision}+\mathrm{Recall}}
  \]
  where Precision and Recall are defined as:
  \[
    \mathrm{Precision}=\frac{TP}{TP+FP}
  \]
  \[
    \mathrm{Recall}=\frac{TP}{TP+FN}
  \]

  \begin{figure*}[t]
    \centering
    \begin{equation*}
    \mathrm{MCC}=
    \frac{
      \sum_{i=1}^{K}\sum_{j=1}^{K} C_{ij}\left(\delta_{ij}-\frac{C_{i+}C_{+j}}{N^{2}}\right)
    }{
      \sqrt{
        \left(\sum_{i=1}^{K} C_{i+}\right)
        \left(\sum_{j=1}^{K} C_{+j}\right)
        \left(\sum_{i=1}^{K}\sum_{j=1}^{K}\frac{C_{i+}C_{+j}}{N^{2}}\right)
      }
    }
    \end{equation*}
    \end{figure*}
  
  \textbf{Macro F1}: It is the average of the F1 scores  across classes, reflecting performance consistency in multi-class label  analysis. Defined as:
  \[
    \mathrm{Macro~F1}=\frac{1}{N}\sum_{i=1}^{N}\mathrm{F1}_i
  \]
  where $N$ is the number of classes and $\mathrm{F1}_i$ is the F1 Score for class $i$.

  \textbf{Matthews Correlation Coefficient（MCC)}~\citep{matthews1975comparison}：It is a comprehensive classification metric for imbalanced data in sentiment analysis, as defined in the formula, where $K$ is the total number of classes ($K=2$ for binary classification and $K>2$ for multi-class classification); $C_{ij}$ is the entry in the confusion matrix at row $i$ and column $j$ (the number of samples whose true class is $i$ and predicted class is $j$); $C_{i+}$ is the sum of row $i$ (the number of samples with true class $i$); $C_{+j}$ is the sum of column $j$ (the number of samples predicted as class $j$); and $N$ is the total number of samples, i.e., $N=\sum_{i=1}^{K}C_{i+}=\sum_{j=1}^{K}C_{+j}$.
   
    \textbf{Exact Match (EM) Accuracy}~~\cite{rajpurkar2016squad}： It requires exact matches between predicted and true labels, defined as follows:
    \[
    \mathrm{EM~Accuracy} = \frac{\text{Number of matches}}{\text{Total samples}}
    \]
    
    \textbf{BERTScore}~\cite{zhang2019bertscore}：It is the harmonic mean of Precision and Recall, computed as follows:
    \[
    \mathrm{BERTScore} = \frac{2 \cdot (\text{Precision} \cdot \text{Recall})}{\text{Precision} + \text{Recall}}
    \]
    
    \textbf{BARTScore}~\cite{yuan2021bartscore}: It evaluates generated text quality by calculating reconstruction error with the BART model, defined as follows:
    \[
    \mathrm{BARTScore} =  \text{BART}(Ref, Gen)
    \]
    where \text{Ref} and \text{Gen} denote reference and generated texts, respectively. \text{BART}($Ref$, $Gen$) is the reconstruction error calculated by the BART model. 
    
    \textbf{RiskScore}: It averages BERTScore for semantic matching between "risks encountered" and "matching solutions", detailed as follows:
    \[
    \mathrm{RiskScore} = \frac{1}{2} \sum_{i=1}^{2} \text{BERTScore}(Ref_i, Gen_i)
     \]
    where ${Ref}_i$ and ${Gen}_i$ denote reference and generated texts.
    
    \textbf{InspectScore}: It combines numerical matching for "offense charge" and semantic matching for "criminal conduct" and "penalty result", defined as:
    \begin{align*}
    \mathrm{InspectScore}&=\frac{1}{2}(\text{EM~Accuracy}(Ref_o, Gen_o)\\
                         &\hspace{-2em}+\frac{1}{2}\sum_{i=1}^{2}\text{BERTScore}(Ref_i, Gen_i))
    \end{align*}
    where ${Ref}_i$ and ${Gen}_i$ denote the reference and prediction for two semantic matching fields, while  ${Ref}_o$ and ${Gen}_o$ denote those for precise numerical (exact) matching.
    
    \textbf{PlanScore}: It combines semantic matching for "core idea" and precise numerical matching for multiple fields (as each data entry varies, e.g., "pre-tax deductible amount of original plan", "taxable amount of original plan", "pre-tax deductible amount after planning", and "tax savings amount"), defined as follows:
    \begin{align*}
    \mathrm{PlanScore}&=\frac{1}{2}(\text{BERTScore}(Ref_c, Gen_c)\\
                         &\hspace{-2em}+\frac{1}{p}\sum_{i=1}^{p}\text{EM~Accuracy}(Ref_i, Gen_i))
    \end{align*}
where ${Ref}_c$ and ${Gen}_c$ represent the reference and prediction for semantic matching, while ${Ref}_i$ and ${Gen}_i$ denote those for exact matching in $p$ fields.

\subsection{Overall Metrics}
Besides the TaxRisk, TaxInspection, and TaxPlan datasets, our custom-built RiskScore, InspectScore, and PlanScore can be applied as integrated evaluation metrics. For the other datasets, we assess the model's overall performance across datasets using the weighted average of all specialized metrics. Specifically, in the TaxSCQ, TaxTopic, and TaxCrime datasets, the final overall score is calculated as the average of ACC, F1, and Macro F1 metrics, defined as follows：
\[
 \mathrm{S_{Overall}}=\frac{\text{ACC} + \text{F1} + \text{Macro\ F1}}{3}. 
 \]
For the TaxKQA, TaxBoard, TaxRecite, TaxSum and TaxOpinion datasets, the overall score is calculated by transforming BARTScore to be greater than 0 with a weight of 10, and then averaging it with BERTScore, ensuring normalization to the range [0, 1], defined as follows：
\[
 \mathrm{S_{Overall}}=\frac{\text{BERTScore}+\frac{\text{BARTScore} + 10}{10}}{2}. 
 \]

\section{Detailed Results}

The detailed results of these models are presented in Table~\ref{tab:details}. Unless otherwise specified, the issues discussed are relevant to both zero-shot and one-shot scenarios. We conduct a thorough analysis of certain notable commonalities or differences observed in each task.

\textbf{TaxRecite.} The TaxRecite dataset requires models to faithfully reproduce the original wording of given tax provisions, which is highly challenging for models that have not explicitly encountered the corresponding regulations. Closed-source large models (e.g., ERNIE-3.5) perform best on this task, often generating article texts that closely match the reference answers. In contrast, most open-source models, lacking explicit memorization of tax statutes, tend to produce incomplete or paraphrased outputs. Typical errors include omitting critical legal phrasing, summarizing provisions in free-form language instead of verbatim recitation, and adding unnecessary explanations. We also observe that, while some models can recover partial snippets in the zero-shot setting, providing a demonstration may interfere with recall for certain models: they rigidly imitate the example format and deviate from the exact wording, leading to performance degradation under the one-shot setting.

\textbf{TaxSum.} In TaxSum, models are required to read long policy news articles and generate concise summaries. This task evaluates both language understanding and summarization, as well as the ability to capture tax-related key information. Results show that closed-source models generally exhibit clear advantages in identifying salient points and maintaining factual consistency, whereas many smaller open-source models produce overly generic summaries, miss critical details (e.g., dates, numbers, or policy names), or even introduce content inconsistent with the source. A plausible explanation is that tax policy news contains abundant domain-specific terminology and complex sentence structures, making comprehensive understanding difficult for weaker models.

\textbf{TaxTopic.} TaxTopic is a topic classification task where models must determine the tax theme of a news article, testing keypoint extraction and topic recognition. Results indicate that open-source models can sometimes capture key terms and contextual cues to identify the relevant tax category. Under the zero-shot setting, GLM4 outperforms closed-source models; however, beyond this case, many open-source models achieve near-random performance and struggle on certain categories. This discrepancy likely stems from the fact that tax news often involves specialized terminology and implicit domain context, requiring substantial tax knowledge to classify accurately.

\textbf{TaxRead.} TaxRead provides relevant laws or official notices as context and asks models to extract answers accordingly, resembling an open-book question answering scenario. Most models achieve strong performance, likely because answers are explicitly present in the provided materials and can be retrieved via pattern matching, reducing reliance on external knowledge. Even smaller models can locate correct spans through surface matching. Moreover, providing demonstrations helps models better follow the expected answer format and reduce verbosity, leading to improved accuracy for most models in the one-shot setting.

\textbf{TaxCalc.} All models perform poorly on TaxCalc, making it one of the most challenging datasets. This task requires computing tax liabilities or tax differences given a scenario, involving numerical operations, logical reasoning, and tax-rule application. Such requirements exceed the strengths of current LLMs: even top models such as GPT-4o and ERNIE-3.5 frequently fail on complex tax planning computations, with accuracy close to zero. While a few large models occasionally produce near-correct answers for simple cases, almost no model can solve more complex instances (e.g., multi-step progressive calculations or comparative scenario reasoning).

\textbf{TaxSCQ.} TaxSCQ is a single-choice multiple-choice task where models must select exactly one correct option. It evaluates detailed tax knowledge and basic reasoning in an exam-like format. Closed-source models perform relatively well. Among open-source models, InternLM2.5 and GLM4 achieve competitive performance with closed-source models under the zero-shot setting. However, most open-source models remain only slightly above random guessing (25\%). We also observe that some models simply restate the prompt or generate responses unrelated to the options, revealing failures in aligning with the multiple-choice answering format.

\textbf{TaxMCQ.} TaxMCQ is a more difficult multi-choice format where models must select all correct options. Compared to single-choice questions, this task demands more comprehensive judgment: models must identify all correct points while avoiding distractors. Results show that most models cannot correctly select all correct options simultaneously. The best-performing models are Yi, Qwen2.5, and InternLM2.5, all of which surpass closed-source models under the zero-shot setting. A typical error pattern is a combination of omissions and false positives: models often identify only the most salient correct option while missing other correct ones, and they also guess incorrect options due to uncertainty.

\textbf{TaxQA.} TaxQA requires models to generate free-form answers to tax questions, assessing domain knowledge application and explanation quality. Overall, model differences are not pronounced and performance is generally moderate. Among open-source models, Qwen2.5 and GLM4 perform relatively well in the zero-shot setting. We also find that knowledge-limited models may fabricate non-existent tax rules or provide incorrect tax rates, producing factually unreliable answers.

\textbf{TaxBoard.} TaxBoard requires models to produce official-style responses to taxpayers’ inquiries, typically demanding concise answers supported by legal bases. Results confirm that models with strong instruction-following capabilities perform well on this dataset. ERNIE-3.5 can emulate the style of official tax authority replies: it tends to answer directly, respond clearly, and cite relevant statutes or regulatory references to enhance credibility. In contrast, some open-source models produce poorly structured answers lacking legal support, and weaker models may deviate from the question altogether.

\textbf{TaxCrime.} TaxCrime asks models to determine which tax law or regulation is violated given a case description, requiring both statute-level knowledge and case-to-law mapping ability. Results show that most models struggle to identify the correct legal provisions. ERNIE-3.5 consistently outperforms other models; this advantage may be associated with its knowledge-enhanced paradigm and broader Chinese knowledge coverage. In comparison, open-source models generally lack statute-level Chinese tax knowledge injection and therefore fail to map case details to specific provisions.

\textbf{TaxOpinion.} TaxOpinion requires models to summarize informal online public opinion texts, which are often colloquial and noisy (e.g., comments on tax evasion cases). Results exhibit substantial divergence across models. Some open-source models can extract main viewpoints from lengthy posts and generate coherent summaries aligned with the original intent. By contrast, weaker models frequently miss the main point: they either paraphrase fragmented sentences without distilling the central idea, or ignore the stance and sentiment expressed in the post, resulting in unfocused summaries.

\textbf{TaxRisk.} TaxRisk asks models to read tax risk-control articles and extract risk points along with corresponding mitigation measures, outputting them in a predefined JSON structure. This task combines information extraction and structured generation, representing a typical knowledge-application scenario. Results vary markedly across models. In addition to closed-source models, GLM4 and Qwen2.5 perform well under the zero-shot setting. Inspection of the outputs shows that models may capture only major risks while missing secondary ones, or provide overly generic mitigation strategies.

\textbf{TaxInspect.} TaxInspect requires models to read tax inspection cases and extract key elements in a structured format, emphasizing both fine-grained factual extraction and strict adherence to output schemas. Results show that only closed-source models and a few open-source models (e.g., GLM4, DeepSeek-R1) perform well. For most open-source models, outputs either largely copy the original case text without extracting fields, or contain severe misalignment where background information is incorrectly filled into target fields.

\textbf{TaxPlan.} TaxPlan is among the most demanding datasets, requiring models to propose tax planning solutions for enterprise scenarios, compute tax liabilities and savings under different strategies, and output results in a structured format. This task integrates scenario understanding, quantitative reasoning, and plan generation, making it extremely challenging. Results show that ERNIE-3.5 performs slightly worse, while GLM4 is the best among open-source models. Overall performance remains unsatisfactory across models: few can fully cover all elements in the reference answers, and accurate numerical computation is particularly rare. Qualitatively, models may propose broadly reasonable planning ideas, but they frequently fail when precise numerical calculations are required.

\begin{table*}[htb!]
   \renewcommand{\thetable}{C1}
    \centering
    \footnotesize
    \setlength\tabcolsep{1pt}
    \renewcommand{\arraystretch}{1.2}
    
        \begin{tabular}{m{0.7cm}<{\centering}|m{1cm}<{\centering} |m{5cm}| m{9cm} }
        \toprule
        \textbf{Task}  & \textbf{Dataset} & \multicolumn{1}{c}{\textbf{Definition}} & \multicolumn{1}{c}{\textbf{Collection Details}}  \\ 
        \hline
        \multirow{2}{*}{\textbf{KM}} & \rotatebox{70}{\textbf{TaxRecite}}& \textbf{Tax Law Recitation:} Provide the section number of the tax law, and require the model to accurately recite the original description. & Download legal and regulatory documents from the official website of the State Taxation Administration（国家税务总局）, extract a portion of the original text of the provisions, and process them into a question-and-answer format.  \\ 
        \hline
        \multirow{9}{*}{\textbf{KU}} & \rotatebox{70}{\textbf{TaxSum}} & \textbf{Tax News Summarization:} Provide the news report and require the model to extract key information and generate a summary. & Select tax-related news articles from the Tax House website, retain the manually annotated summaries from the site, and compress the original text while preserving its main points to optimize the input length. \\ 
        \cline{2-4}
        & \rotatebox{70}{\textbf{TaxTopic}} & \textbf{Tax Topic Classification:}  Provide the news title  and require the model to determine its corresponding thematic category. & Select tax-related articles from the Tax House website, extract their titles, and record the categories they belong to on the website as labels for thematic identification. \\ 
        \cline{2-4}
        & \rotatebox{70}{\textbf{TaxRead}} & \textbf{Tax Reading Comprehension:} Provide the article along with relevant questions, and require the model to extract relevant content for answer. & Select tax-related articles from the Tax House (税屋) website, compress the original text using ChatGPT while retaining its main points to optimize input length, and utilize ChatGPT to extract questions and corresponding answers from the compressed articles. \\
        \hline
         \multirow{42}{*}{\textbf{KA}} & \rotatebox{70}{\textbf{TaxCalc}} & \textbf{Tax Payment Calculation:} Provide a corporate tax case and require the model to calculate and provide the accurate amount. & Carefully select calculation problems from the exercise book for the Certified Tax Agent exam, ensuring the questions cover multiple tax types to comprehensively address and encompass the tax domain. \\ 
         \cline{2-4}
        & \rotatebox{70}{\textbf{TaxSCQ}} & \textbf{Tax Single-Choice Exam:} Provide the question along with four options, requiring the model to identify one correct answer. & Carefully select single-choice questions from the exercise book for the Certified Tax Agent Exam (注册税务师考试）. To ensure broad coverage, we extract relevant exam questions from both real exam papers and mock exam papers, encompassing various knowledge areas and key exam topics.\\  
        \cline{2-4}
        & \rotatebox{70}{\textbf{TaxMCQ}} & \textbf{Tax Multiple-Choice Exam:} Provide the question along with five options, requiring the model to identify all correct answers. &  Carefully select multiple-choice questions from the exercise book for the Certified Tax Agent exam. To ensure broad coverage, we extract relevant exam questions from both real exam papers and mock exam papers, encompassing various knowledge areas and key exam topics.\\  
        \cline{2-4}
        & \rotatebox{70}{\textbf{TaxQA}} & \textbf{Tax Knowledge Q\&A:} Provide a question that asks the model to give semantically similar answers with limited length. & Download legal regulations, tax service guides, and tax incentive documents from the official website of the State Taxation Administration, as well as tax knowledge documents from Baidu Wenku（百度文库）. Then, extract tax-related questions and their answers from these files. \\ 
        \cline{2-4}
        & \rotatebox{70}{\textbf{TaxBoard}} & \textbf{Tax Board Q\&A:} Provide an official user inquiry question, requiring the model to give a response with limited length. & Collect real user questions and official responses from the message board of the State Taxation Administration's website, simulating authentic user consultation scenarios to provide users with more comprehensive references and guidance. \\
        \cline{2-4}
        & \rotatebox{70}{\textbf{TaxCrime}} & \textbf{Tax Law Identification:} Provide tax-related crime facts and ask the model to identify the violated law article from the options. & Download tax-related criminal case judgments from China Judgments Online, extract criminal behaviors and violated law articles, and use ChatGPT to label the articles for a classification task that identifies the violated laws based on the behaviors.  \\
        \cline{2-4}
        & \rotatebox{70}{\textbf{TaxOpinion}} & \textbf{Tax Opinion Summarization:} Provide tax-related public opinion articles and require the model to generate corresponding summaries. & Retrieve public opinions on tax evasion from the StoneDT （思通数科）Opinion system, use ChatGPT to assess relevance to taxation, clean and compress the tax-related content by removing emoticons and ensuring fluent, logical sentences, and finally generate summaries with ChatGPT as dataset answers.  \\
        \cline{2-4}
        & \rotatebox{70}{\textbf{TaxRisk}} & \textbf{Tax Risk Prevention:} Provide articles on fiscal and tax risk control and ask the model to extract the mentioned risks and their solutions. & Download articles on fiscal and tax risk control from Tax House (税屋) website, compress them using ChatGPT while retaining key information as input text, then extract the faced risks and corresponding  solutions information using ChatGPT as answers. \\
        \cline{2-4}
        & \rotatebox{70}{\textbf{TaxInspect}} & \textbf{Tax Inspection Analysis:} Provide tax inspection cases and extract mentioned criminal behaviors, charges, and penalties. & Download tax inspection case articles from Tax House website, compress them using ChatGPT while retaining key information as input text, then extract "criminal behaviors", "crime charges", and "punishment results" information using ChatGPT as answers.  \\
        \cline{2-4}
        & \rotatebox{70}{\textbf{TaxPlan}} & \textbf{Tax Strategy Planning:} Provide specific tax scenarios and require the model to perform tax planning to achieve tax-saving effects. & Extract tax planning cases from tax planning case books, including scenario descriptions and tax planning solutions, use ChatGPT to extract structured information as answers, replacing the original descriptive tax planning solutions with structured answers to form the dataset. \\
        \bottomrule
        \end{tabular}
    \caption{Description of the data definitions and collection details for the three-level tax tasks in TaxPraBen.}
    \label{tab:collection}
\end{table*}

\begin{figure*}[htb!] 
    \renewcommand{\thefigure}{C1}
    \centering
    \begin{subfigure}
    \centering
    \fbox{\includegraphics[width=\textwidth]{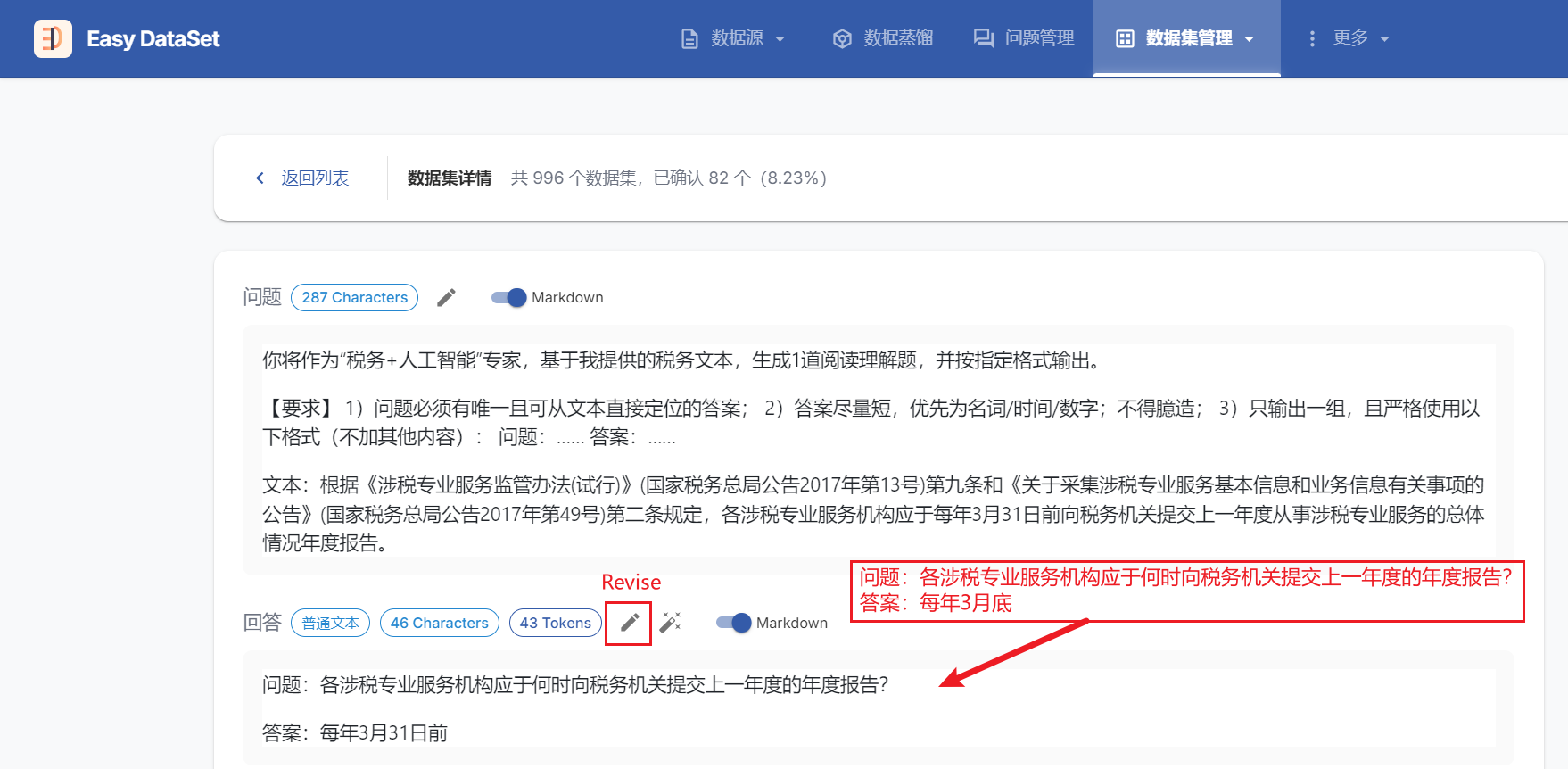}}
    \end{subfigure}

    \vspace{5em}

    \begin{subfigure}
    \centering
    \fbox{\includegraphics[width=\textwidth]{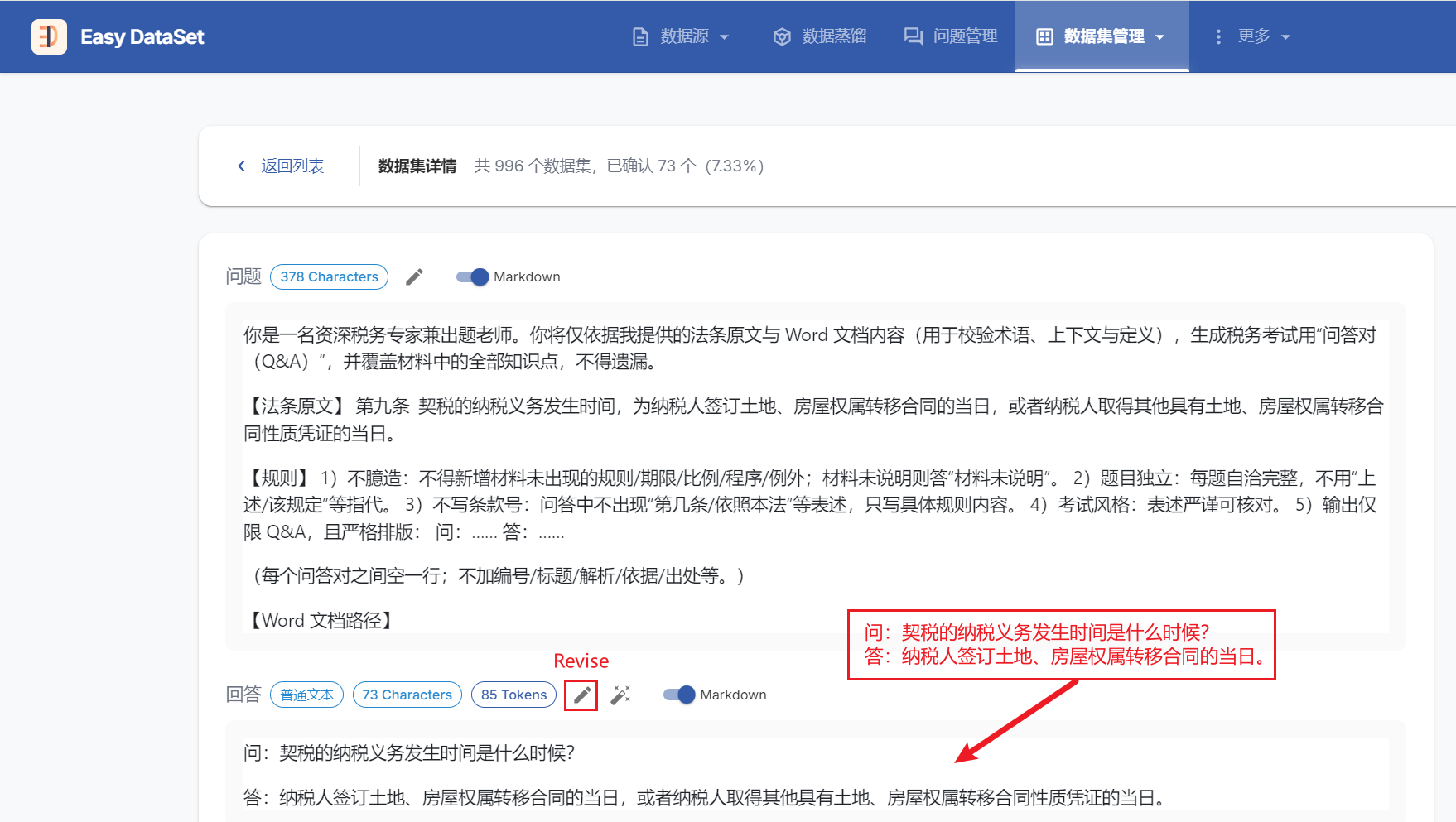}}
    \end{subfigure}
     \vskip 2pt
     \begin{flushleft}
     \footnotesize
     \textit{\textbf{Note}: Our annotation process on the EasyDataset platform goes beyond simple question(问题)\&answer(回答）interface-based matching. We treat each question as an "annotation requirement + original document" serving as a cross-reference annotation guideline, with each answer, including the Q\&A pairs generated by ChatGPT, requiring proofreading by domain experts.}
     \end{flushleft}
    \caption{An example screenshot demonstrating manual adjustment of the TaxRead and TaxQA datasets.}
    \label{fig:screenshot}
\end{figure*}

\begin{table*}[t]
   \renewcommand{\thetable}{C4}
    \centering
    \footnotesize
    \setlength\tabcolsep{0pt}
    \renewcommand{\arraystretch}{2}
    
    \begin{tabular}{m{0.98\textwidth}}
    
    \toprule
    
    \textbf{Prompt:} 根据问题需求，请提供法条的原文部分，不必对其含义做出解释，也不需要指明是第几条。(\textit{According to the question requirements, please provide the verbatim excerpt of the relevant legal provisions. Do not explain their meaning, and do not indicate article numbers.})
    \\ \hline
    
    \multicolumn{1}{c}{\textbf{Zero-Shot}}
    \\ \hline
    
    \textbf{Query:} \textcolor{blue}{\{Prompt\}}\\
    \textbf{Text:} 2007年12月6日通过的《中华人民共和国企业所得税法实施条例》中第一百一十一条的内容是什么？ (\textit{What is the content of Article 111 of the Implementation Regulations of the Enterprise Income Tax Law of the People's Republic of China, adopted on December 6, 2007?})
    \\ \hline
    
    \textbf{Answer:} 企业所得税法第四十一条所称合理方法，包括：(一)可比非受控价格法，是指按照没有关联关系的交易各方进行相同或者类似业务往来的价格进行定价的方法；(二)再销售价格法，是指按照从关联方购进商品再销售给没有关联关系的交易方的价格，减除相同或者类似业务的销售毛利进行定价的方法；(三)成本加成法，是指按照成本加合理的费用和利润进行定价的方法；(四)交易净利润法，是指按照没有关联关系的交易各方进行相同或者类似业务往来取得的净利润水平确定利润的方法；(五)利润分割法，是指将企业与其关联方的合并利润或者亏损在各方之间采用合理标准进行分配的方法；(六)其他符合独立交易原则的方法。 (\textit{The ``reasonable methods'' referred to in Article 41 of the Enterprise Income Tax Law include: (1) the Comparable Uncontrolled Price (CUP) method, i.e., determining the price by reference to the prices applied in identical or similar transactions between parties without an associated relationship; (2) the Resale Price method, i.e., determining the price by reference to the price at which goods purchased from an associated party are resold to an unrelated party, minus the gross margin for identical or similar transactions;(3) the Cost Plus method, i.e., determining the price based on cost plus reasonable expenses and profit; (4) the Transactional Net Margin method (TNMM), i.e., determining profit based on the net profit level realized in identical or similar transactions between unrelated parties; (5) the Profit Split method, i.e., allocating the combined profit or loss between an enterprise and its associated parties according to reasonable allocation keys; and (6) other methods consistent with the arm's-length principle.})
    \\ \hline
    
    \multicolumn{1}{c}{\textbf{One-Shot}}
    \\ \hline
    
    \textbf{Query:} \textcolor{blue}{\{Prompt\}}\\
    \textbf{Example-Text:} 2010年2月10日通过的《税务行政复议规则》中第二十五条的内容是什么？ (\textit{What is the content of Article 25 of the Rules on Tax Administrative Reconsideration, adopted on February 10, 2010?})\\
    \textbf{Example-Answer:} 同一行政复议案件申请人超过5人的，应当推选1至5名代表参加行政复议。 (\textit{In an administrative reconsideration case where more than five applicants are involved, the applicants shall designate one to five representatives to participate in the administrative reconsideration proceedings.} )\\
    \textbf{Text:} 2007年12月6日通过的《中华人民共和国企业所得税法实施条例》中第一百一十一条的内容是什么？ (\textit{What is the content of Article 111 of the Implementation Regulations of the Enterprise Income Tax Law of the People's Republic of China, adopted on December 6, 2007?})
    \\ \hline
    
    \textbf{Answer:} 企业所得税法第四十一条所称合理方法，包括：(一)可比非受控价格法，是指按照没有关联关系的交易各方进行相同或者类似业务往来的价格进行定价的方法；(二)再销售价格法，是指按照从关联方购进商品再销售给没有关联关系的交易方的价格，减除相同或者类似业务的销售毛利进行定价的方法；(三)成本加成法，是指按照成本加合理的费用和利润进行定价的方法；(四)交易净利润法，是指按照没有关联关系的交易各方进行相同或者类似业务往来取得的净利润水平确定利润的方法；(五)利润分割法，是指将企业与其关联方的合并利润或者亏损在各方之间采用合理标准进行分配的方法；(六)其他符合独立交易原则的方法。 (\textit{The ``reasonable methods'' referred to in Article 41 of the Enterprise Income Tax Law include: (1) the Comparable Uncontrolled Price (CUP) method, i.e., determining the price by reference to the prices applied in identical or similar transactions between parties without an associated relationship; (2) the Resale Price method, i.e., determining the price by reference to the price at which goods purchased from an associated party are resold to an unrelated party, minus the gross margin for identical or similar transactions;(3) the Cost Plus method, i.e., determining the price based on cost plus reasonable expenses and profit; (4) the Transactional Net Margin method (TNMM), i.e., determining profit based on the net profit level realized in identical or similar transactions between unrelated parties; (5) the Profit Split method, i.e., allocating the combined profit or loss between an enterprise and its associated parties according to reasonable allocation keys; and (6) other methods consistent with the arm's-length principle.})
    \\
    \bottomrule
    \end{tabular}
    \caption{An instance format with paired prompts for the TaxRecite dataset in the KM task.}
\label{tab:TaxRecite}
\end{table*}

\begin{table*}[t]
    \renewcommand{\thetable}{C5}
    \centering
    \footnotesize
    \setlength\tabcolsep{0pt}
    \renewcommand{\arraystretch}{1}
    
    \begin{tabular}{m{0.98\textwidth}}
    
    \toprule
    
    \textbf{Prompt:} 请根据以下文本编写一个简短的概要，字数不超过300字，重点在于提炼核心内容。 (\textit{Please write a concise summary of the following text in no more than 300 words, focusing on extracting the key points.})
    \\ \hline
    
    \multicolumn{1}{c}{\textbf{Zero-Shot}}
    \\ \hline
    
    \textbf{Query:} \textcolor{blue}{\{Prompt\}}\\
    \textbf{Text:} 为配合个人所得税专项附加扣除工作，依据相关法律法规，三部门联合发布通知，明确个人所得税住房贷款利息专项附加扣除相关信息归集事项。信息归集范围为1989年1月1日（含）后发放的商业性个人住房贷款，不包括个人商用房贷款。数据项涵盖借款人姓名、证件号码、贷款银行、合同编号、是否为首套住房贷款等11项内容。住房公积金贷款由主管部门负责归集。各商业银行于2019年3月1日前向征信系统报送信息。....。 (\textit{To facilitate the work related to the special additional deductions for individual income tax, the three departments have jointly issued a notice clarifying the information collection matters related to the special additional deduction for housing loan interest. The scope of information collection includes commercial personal housing loans issued after January 1, 1989 (inclusive), excluding personal commercial property loans. The data items cover 11 contents, including the borrower's name, identification number, loan bank, contract number, and whether it is a first home loan, among others. Housing provident fund loans are to be collected by the relevant management department. All commercial banks must report this information to the credit reporting system by March 1, 2019. ... .})
    \\ \hline
    
    \textbf{Answer:} 信息归集工作时间紧、任务重、要求高，各商业银行要高度重视，充分认识贯彻施行《个人所得税专项附加扣除暂行办法》的重要性和紧迫性，切实加强组织领导，务必在规定时间内做好各项工作。 (\textit{Given the tight timelines, heavy workload, and stringent requirements associated with information consolidation, all commercial banks should attach great importance to this effort. They should fully recognize the significance and urgency of implementing the \emph{Interim Measures for Special Additional Deductions of Individual Income Tax}, strengthen organizational leadership, and ensure that all related tasks are completed within the prescribed time frame.})
    \\ \hline
    
    \multicolumn{1}{c}{\textbf{One-Shot}}
    \\ \hline
    
    \textbf{Query:} \textcolor{blue}{\{Prompt\}}\\
    \textbf{Example-Text:} 国家税务总局厦门市税务局通知，自即日起，通过12366纳税服务平台受理网上咨询以提升服务品质。纳税人、缴费人可点击网站“我要咨询”，使用智能咨询、网上留言、在线咨询和众包互助服务。原有智能咨询服务平台继续用于社保自助咨询；历史留言可在“办税服务-历史咨询”查看。智能咨询页面功能如下：系统根据IP地址默认显示地区，用户可修改。1. 智能咨询：录入问题自助查询。2. 在线咨询：工作时间可与客服对话。3. 网上留言：新用户需注册登录后提交留言。4. 众包互助：登录后可查看或帮助回复问题。感谢纳税人对12366纳税咨询服务工作的支持。 (\textit{The Xiamen Municipal Tax Service of the State Taxation Administration (STA) announces that, effective immediately, online inquiries will be handled through the 12366 Tax Service Platform to enhance service quality. Taxpayers and contribution payers may click ``I Want to Consult'' on the platform website and use the Intelligent Consultation, Online Message, Live Chat, and Crowdsourced Mutual-Help services. The original Intelligent Consultation platform will remain available for self-service inquiries related to social insurance, and historical messages can be accessed via ``Tax Service''--``Inquiry History''. The Intelligent Consultation page provides the following functions: the system automatically displays the user’s region based on the IP address, which can be modified manually. (1) Intelligent Consultation: users enter questions for self-service retrieval; (2) Live Chat: during working hours, users may communicate with customer service staff in real time; (3) Online Message: new users must register and log in before submitting messages; and (4) Crowd sourced Mutual Help: after logging in, users may browse and assist in answering questions. The Xiamen Municipal Tax Service of the STA appreciates taxpayers’ continued support for the 12366 tax consultation services.} )\\
    \textbf{Example-Answer:} 为更好地服务纳税人和缴费人，提升网上咨询服务品质，我局网站自即日起，通过国家税务总局12366纳税服务平台受理网上咨询。点击首页“我要咨询”，即可使用智能咨询、网上留言、在线咨询和众包互助等咨询服务。 (\textit{To better serve taxpayers and fee payers and to enhance the quality of online consultation services, our bureau will, effective immediately, accept online inquiries via the State Taxation Administration’s 12366 Tax Service Platform. Users may click ``I want to consult'' on the homepage to access consultation services including intelligent consultation, online messaging, live online consultation, and crowdsourced mutual assistance.})\\
    \textbf{Text:} 为配合个人所得税专项附加扣除工作，依据相关法律法规，三部门联合发布通知，明确个人所得税住房贷款利息专项附加扣除相关信息归集事项。信息归集范围为1989年1月1日（含）后发放的商业性个人住房贷款，不包括个人商用房贷款。数据项涵盖借款人姓名、证件号码、贷款银行、合同编号、是否为首套住房贷款等11项内容。住房公积金贷款由主管部门负责归集。各商业银行于2019年3月1日前向征信系统报送信息。....。 (\textit{To facilitate the work related to the special additional deductions for individual income tax, the three departments have jointly issued a notice clarifying the information collection matters related to the special additional deduction for housing loan interest. The scope of information collection includes commercial personal housing loans issued after January 1, 1989 (inclusive), excluding personal commercial property loans. The data items cover 11 contents, including the borrower's name, identification number, loan bank, contract number, and whether it is a first home loan, among others. Housing provident fund loans are to be collected by the relevant management department. All commercial banks must report this information to the credit reporting system by March 1, 2019. ... .}.)
    
    \\ \hline
    
    \textbf{Answer:} 信息归集工作时间紧、任务重、要求高，各商业银行要高度重视，充分认识贯彻施行《个人所得税专项附加扣除暂行办法》的重要性和紧迫性，切实加强组织领导，务必在规定时间内做好各项工作。 (\textit{Given the tight timelines, heavy workload, and stringent requirements associated with information consolidation, all commercial banks should attach great importance to this effort. They should fully recognize the significance and urgency of implementing the \emph{Interim Measures for Special Additional Deductions of Individual Income Tax}, strengthen organizational leadership, and ensure that all related tasks are completed within the prescribed time frame.})
    \\
    \bottomrule
    \end{tabular}
    \caption{An instance format with paired prompts for the TaxSum dataset in the KU task.}
\label{tab:TaxSum}
\end{table*}

\begin{table*}[t]
   \renewcommand{\thetable}{C6}
    \centering
    \footnotesize
    \setlength\tabcolsep{0pt}
    \renewcommand{\arraystretch}{1}
    
    \begin{tabular}{m{0.98\textwidth}}
    
    \toprule
    
    \textbf{Prompt:} 阅读标题以了解文章主题，并在提供的选项中选择最适合表达这一主题的一个，选项有：A.财商资讯\ \ B.财税表单\ \ C.地方法规\ \ D.法规解读\ \ E.纳税调整\ \ F.纳税评估\ \ G.税收优惠\ \ H.税收筹划\ \ I.税务问答\ \ J.中央法规。(R\textit{ead the title to identify the main topic of the article, and select the single option that best matches this topic from the following candidates: A.~Financial literacy news; B.~Tax and finance forms; C.~Local regulations; D.~Regulation interpretation; E.~Taxable income adjustments; F.~Tax assessment; G.~Tax incentives; H.~Tax planning; I.~Tax Q\&A; J.~National regulations.})
    \\ \hline
    
    \multicolumn{1}{c}{\textbf{Zero-Shot}}
    \\ \hline

    \textbf{Query:} \textcolor{blue}{\{Prompt\}}\\
    \textbf{Text:} 浙江省国家税务局公告2014年第10号\ 浙江省国家税务局关于企业资产损失税前扣除申报有关问题的公告 (\textit{Announcement No.~10 (2014) of the Zhejiang Provincial Office of the State Administration of Taxation: On Issues Concerning the Pre-tax Deduction Declaration for Enterprise Asset Losses.})
    \\ \hline
    
    \textbf{Answer:} C
    \\ \hline
    
    \textbf{Choices:} [``A'', ``B'', ``C'', ``D'', ``E'', ``F'', ``G'', ``H'', ``I'', ``J'']
    \\ \hline
    
    \textbf{Gold:} 2
    \\ \hline
    
    \multicolumn{1}{c}{\textbf{One-Shot}}
    \\ \hline

    \textbf{Query:} \textcolor{blue}{\{Prompt\}}\\
    \textbf{Example-Text:} 税务总局解答热点财税问题 (\textit{Hotspot Q\&A on Tax and Fiscal Issues Issued by China’s State Taxation Administration})\\
    \textbf{Example-Answer:} I\\
    \textbf{Text:} 浙江省国家税务局公告2014年第10号\ 浙江省国家税务局关于企业资产损失税前扣除申报有关问题的公告 (\textit{Announcement No.~10 (2014) of the Zhejiang Provincial Office of the State Administration of Taxation: On Issues Concerning the Pre-tax Deduction Declaration for Enterprise Asset Losses.})
    
    \\ \hline
    
    \textbf{Answer:} C
    \\ \hline
    
    \textbf{Choices:} [``A'', ``B'', ``C'', ``D'', ``E'', ``F'', ``G'', ``H'', ``I'', ``J'']
    \\ \hline
    
    \textbf{Gold:} 2
    \\
    \bottomrule
    \end{tabular}
    \caption{An instance format with paired prompts for the TaxTopic dataset in the KU task.}
\label{tab:TaxTopic}
\end{table*}

\begin{table*}[t]
   \renewcommand{\thetable}{C7}
    \centering
    \footnotesize
    \setlength\tabcolsep{0pt}
    \renewcommand{\arraystretch}{1}
    
    \begin{tabular}{m{0.98\textwidth}}
    
    \toprule
    
    \textbf{Prompt:} 关于下面提出的问题，请仔细考虑四个可能的答案，并选出你认为正确的那个，仅需指出“A”、“B”、“C”或“D”，不需要解释。(\textit{Please carefully consider the four candidate answers to the question below and select the one you deem correct. Return only the letter ``A'', ``B'', ``C'', or ``D'' without providing any explanation.})
    \\ \hline

    \multicolumn{1}{c}{\textbf{Zero-Shot}}
    \\ \hline
    
    \textbf{Query:} \textcolor{blue}{\{Prompt\}}\\
    \textbf{Text:} 关于退还关税，下列说法正确的是()。   A.纳税人自缴纳税款之日起3年内，可以申请退还关税   B.海关发现实际征收税款多于应收税款，可以不用通知纳税义务人办理退还手续   C.已征出口关税的货物，因故未装运出口申请退关的，可以申请退还关税   D.海关应当自受理退税申请之日起60日内查实并通知纳税义务人办理退还手续 (\textit{Regarding the refund of customs duties, which of the following statements is correct()? A.Within three years from the date on which the taxpayer pays the duty, the taxpayer may apply for a refund of customs duties. B.If the Customs authorities discover that the duties actually collected exceed the duties payable, they may refund the overpaid amount without notifying the taxpayer or tax obligor to complete the refund procedures. C.For goods on which export duties have already been levied, if the goods are not shipped for certain reasons and a duty refund is requested, an application may be filed for a refund of the customs duties. D.The Customs authorities shall, within 60 days from the date of accepting the refund application, verify the relevant facts and notify the taxpayer or tax obligor to complete the refund procedures.})
    \\ \hline
    
    \textbf{Answer:} C
    \\ \hline
    
    \textbf{Choices:} [``A'', ``B'', ``C'', ``D'']
    \\ \hline
    
    \textbf{Gold:} 2
    \\ \hline

    \multicolumn{1}{c}{\textbf{One-Shot}}
    \\ \hline
    
    \textbf{Query:} \textcolor{blue}{\{Prompt\}}\\
    \textbf{Example-Text:} 下列各课税要素，与纳税期限的选择密切相关的是（）。   A.计税依据   B.纳税环节   C.课税对象的性质   D.纳税人 (\textit{Which of the following tax elements is most closely associated with the selection of the tax payment period ( )? A.Tax base B.Taxable stage C.Nature of the taxable object D.Taxpayer})\\
    \textbf{Example-Answer:} C\\
    \textbf{Text:} 关于退还关税，下列说法正确的是()。   A.纳税人自缴纳税款之日起3年内，可以申请退还关税   B.海关发现实际征收税款多于应收税款，可以不用通知纳税义务人办理退还手续   C.已征出口关税的货物，因故未装运出口申请退关的，可以申请退还关税   D.海关应当自受理退税申请之日起60日内查实并通知纳税义务人办理退还手续 (\textit{Regarding the refund of customs duties, which of the following statements is correct()? A.Within three years from the date on which the taxpayer pays the duty, the taxpayer may apply for a refund of customs duties. B.If the Customs authorities discover that the duties actually collected exceed the duties payable, they may refund the overpaid amount without notifying the taxpayer or tax obligor to complete the refund procedures. C.For goods on which export duties have already been levied, if the goods are not shipped for certain reasons and a duty refund is requested, an application may be filed for a refund of the customs duties. D.The Customs authorities shall, within 60 days from the date of accepting the refund application, verify the relevant facts and notify the taxpayer or tax obligor to complete the refund procedur}es.)\\ \hline
    
    \textbf{Answer:} C
    \\ \hline
    
    \textbf{Choices:} [``A'', ``B'', ``C'', ``D'']
    \\ \hline
    
    \textbf{Gold:} 2
    \\
    \bottomrule
    \end{tabular}
    \caption{An instance format with paired prompts for the TaxSCQ dataset in the KA task.}
\label{tab:TaxSCQ}
\end{table*}

\begin{table*}[t]
   \renewcommand{\thetable}{C8}
    \centering
    \footnotesize
    \setlength\tabcolsep{2pt}
    \renewcommand{\arraystretch}{1.8}
    
    \begin{tabular}{m{0.98\textwidth}}
    \toprule
    
    \textbf{Prompt:} 在阅读完下面的文本之后，请根据其内容回答问题，确保回复是从文中摘录且言简意赅，不要重复问题，答案中不要出现句号。(\textit{After reading the following text, please answer the question based on its content. Ensure that your response is a concise excerpt from the text, without repeating the question, and do not include periods in your answer.})
    \\ \hline

    \multicolumn{1}{c}{\textbf{Zero-Shot}}
    \\ \hline
    
    \textbf{Query:} \textcolor{blue}{\{Prompt\}}\\
    \textbf{Text:} 为综合反映与科学评价资产评估机构整体实力和专业胜任能力，推动资产评估行业信用体系建设，提升行业公信力和影响力，促进高质量发展，中国资产评估协会修订并发布了《资产评估机构综合评价办法》。该办法旨在通过系统化的评估体系，全面衡量资产评估机构的业务水平、服务质量、风险管理等多方面的能力，从而引导行业健康发展。新办法已经第六届常务理事会第三次会议审议通过，于2025年2月18日正式发布。问题：《资产评估机构综合评价办法》何时正式发布？ (\textit{To comprehensively reflect and rigorously assess the overall strength and professional competence of asset appraisal firms, promote the development of an industry-wide credit system, enhance credibility and influence, and foster high-quality growth, the China Appraisal Society (CAS) revised and released the \textit{Measures for the Comprehensive Evaluation of Asset Appraisal Firms}. The revised measures aim to establish a systematic evaluation framework that holistically assesses firms' capabilities across multiple dimensions, including operational performance, service quality, and risk management, thereby guiding the healthy development of the industry. The new measures were reviewed and approved at the Third Meeting of the Third Executive Council of the Sixth Standing Council and were formally released on February 18, 2025. \textbf{Question:} When was the Measures for the Comprehensive Evaluation of Asset Appraisal Firms formally released?})
    \\ \hline
    
    \textbf{Answer:} 2025年2月18日 (\textit{February 18, 2025})
    \\ \hline
    
    \multicolumn{1}{c}{\textbf{One-Shot}}
    \\ \hline
    
    \textbf{Query:} \textcolor{blue}{\{Prompt\}}\\
    \textbf{Example-Text:} 国家税务总局平潭综合实验区税务局通知，已委托中国邮政集团公司福建省平潭县分公司为符合条件的自然人代开增值税普通发票。为便于工作开展，自即日起启用“国家税务总局平潭综合实验区税务局委托代开发票专用章”印章。附件为该印章印模。通知时间为2019年8月21日。问题：国家税务总局平潭综合实验区税务局委托哪家公司为符合条件的自然人代开增值税普通发票？ (\textit{The Pingtan Comprehensive Experimental Zone Taxation Bureau of the State Taxation Administration (STA) issued a notice stating that it has entrusted the Pingtan County Branch of China Post Group Co., Ltd. (Fujian Province) to issue VAT ordinary invoices on behalf of eligible natural persons. To facilitate implementation, the bureau has, effective immediately, put into use the seal entitled ``Special Seal for Entrusted Invoice Issuance of the Pingtan Comprehensive Experimental Zone Taxation Bureau of the State Taxation Administration.'' An impression specimen of the seal is provided in the appendix. The notice was issued on August~21,~2019. \textbf{Question:} Which company was entrusted by the Pingtan Comprehensive Experimental Zone Taxation Bureau of the STA to issue VAT ordinary invoices on behalf of eligible natural persons?})\\
    \textbf{Example-Answer:} 中国邮政集团公司福建省平潭县分公司 (\textit{Pingtan County Branch, Fujian Province, China Post Group Corporation.})\\
    \textbf{Text:} 为综合反映与科学评价资产评估机构整体实力和专业胜任能力，推动资产评估行业信用体系建设，提升行业公信力和影响力，促进高质量发展，中国资产评估协会修订并发布了《资产评估机构综合评价办法》。该办法旨在通过系统化的评估体系，全面衡量资产评估机构的业务水平、服务质量、风险管理等多方面的能力，从而引导行业健康发展。新办法已经第六届常务理事会第三次会议审议通过，于2025年2月18日正式发布。问题：《资产评估机构综合评价办法》何时正式发布？ (\textit{To comprehensively reflect and rigorously assess the overall strength and professional competence of asset appraisal firms, promote the development of an industry-wide credit system, enhance credibility and influence, and foster high-quality growth, the China Appraisal Society (CAS) revised and released the \textit{Measures for the Comprehensive Evaluation of Asset Appraisal Firms}. The revised measures aim to establish a systematic evaluation framework that holistically assesses firms' capabilities across multiple dimensions, including operational performance, service quality, and risk management, thereby guiding the healthy development of the industry. The new measures were reviewed and approved at the Third Meeting of the Third Executive Council of the Sixth Standing Council and were formally released on February 18, 2025. \textbf{Question:} When was the Measures for the Comprehensive Evaluation of Asset Appraisal Firms formally released?})
    \\ \hline
    
    \textbf{Answer:} 2025年2月18日 (\textit{February 18, 2025})
    \\
    \bottomrule
    
    \end{tabular}
    \caption{An instance format with paired prompts for the TaxRead dataset in the KA task.}
\label{tab:TaxRead}
\end{table*}

\begin{table*}[t]
    \renewcommand{\thetable}{C9}
    \centering
    \footnotesize
    \setlength\tabcolsep{2pt}
    \renewcommand{\arraystretch}{1}
    
    \begin{tabular}{m{0.98\textwidth}}
    \toprule
    
    \textbf{Prompt:} 针对题目，请直接给出xx处的答案，并将答案带入到问题中复述一遍，例如："业务（1）应缴纳消费税为xx万元"，保证输出字符在30个字符以内。(\textit{For the question, please directly provide the answer for \texttt{xx}, and restate it in the question, such as: 'For transaction (1), the consumption tax payable is \texttt{xx} yuan,' ensuring the output is within 30 characters.})
    \\ \hline

    \multicolumn{1}{c}{\textbf{Zero-Shot}}
    \\ \hline

    \textbf{Query:} \textcolor{blue}{\{Prompt\}}\\
    \textbf{Text:} 位于某市区国家重点扶持的高新技术企业为增值税一般纳税人（制造业企业），2023年销售产品取得不含税收入6500万元，另外取得投资收益320万元，全年发生产品销售成本和相关费用共计5300万元。缴纳的税金及附加339万元，发生的营业外支出420万元，12月末企业自行计算的全年会计利润总额761万元，预缴企业所得税96万元。2024年1月经聘请的税务师事务所审核，发现以下问题：（1）8月中旬以预收款方式销售一批产品，预收含税价款234万元，并收存银行。12月下旬将该批产品发出，但未将预收款转作收入。（不考虑相关成本）... 问题为：新产品研发费应调整的应纳税所得额为xx万元？ (\textit{A nationally supported key high-tech enterprise located in the urban district of a certain city is a general VAT taxpayer (manufacturing enterprise). In 2023, it generated RMB 65.00 million in product sales revenue (VAT-exclusive) and additionally earned investment income of RMB 3.20 million. For the year, the total cost of goods sold and related expenses amounted to RMB 53.00 million. Taxes and surcharges paid were RMB 3.39 million, and non-operating expenses were RMB 4.20 million. As of the end of December, the enterprise calculated an annual accounting profit before tax of RMB 7.61 million, and it had prepaid corporate income tax (CIT) of RMB 0.96 million. In January 2024, after an audit by a hired tax agent firm, the following issues were identified:(1) In mid-August, the enterprise sold a batch of products under an advance-receipt arrangement, receiving an advance payment of RMB 2.34 million (VAT-inclusive), which was deposited in the bank. In late December, the batch of products was delivered, but the advance receipt was not reclassified as revenue. (Relevant costs are ignored.) ... \textbf{Question:} The adjustment to taxable income attributable to the new product R\&D expenses should be \texttt{xx} 10,000 RMB.})
    \\ \hline
    
    \textbf{Answer:} -72.75
    \\ \hline
    
    \multicolumn{1}{c}{\textbf{One-Shot}}
    \\ \hline
    
    \textbf{Query:} \textcolor{blue}{\{Prompt\}}\\
    \textbf{Example-Text:} 我国某居民企业在甲国设立一家分公司，在乙国设立一家持股80\%的子公司，2023年该企业计算的境内利润总额4000万元。相关涉税资料如下：（1）甲国分公司按我国税法确认的销售收入300万元，销售成本500万元。（2）收到乙国子公司投资收益1900万元，子公司已在乙国缴纳企业所得税1000万元，子公司当年税后利润全部按持股比例分配，乙国预提所得税税率5\%。（其他相关资料：该居民企业适用25\%的企业所得税税率，无纳税调整金额，境外已纳税额选择“分国不分项”抵免方式）问题为：2023年该居民企业来源于子公司投资收益的境外实际缴纳税额是多少万元？ (\textit{A Chinese resident enterprise establishes a branch in Country A and an 80\% owned subsidiary in Country B. In 2023, the enterprise reports a domestic total profit of RMB 40 million. The relevant tax information is as follows: (1) The branch in Country A has sales revenue of RMB 3 million and sales cost of RMB 5 million as recognized under Chinese tax law. (2) The enterprise receives RMB 19 million in investment income (dividends) from the subsidiary in Country B. The subsidiary has paid RMB 10 million in corporate income tax in Country B. The subsidiary distributes all of its after-tax profit for the year in proportion to shareholding, and the withholding income tax rate in Country B is 5\%. (Other information: the resident enterprise is subject to a 25\% corporate income tax rate in China, with no tax adjustments, and adopts the ``per-country, no-itemization'' foreign tax credit method.) \textbf{Question:} In 2023, what is the amount of foreign taxes actually paid attributable to the investment income from the subsidiary (in RMB million)?})\\
    \textbf{Example-Answer:} 2023年该居民企业来源于子公司投资收益的境外实际缴纳税额是900.00万元 (\textit{In 2023, the resident enterprise incurred foreign income taxes actually paid amounting to RMB~9.00~million on investment income received from its subsidiaries.})\\
    \textbf{Text:} 位于某市区国家重点扶持的高新技术企业为增值税一般纳税人（制造业企业），2023年销售产品取得不含税收入6500万元，另外取得投资收益320万元，全年发生产品销售成本和相关费用共计5300万元。缴纳的税金及附加339万元，发生的营业外支出420万元，12月末企业自行计算的全年会计利润总额761万元，预缴企业所得税96万元。2024年1月经聘请的税务师事务所审核，发现以下问题：（1）8月中旬以预收款方式销售一批产品，预收含税价款234万元，并收存银行。12月下旬将该批产品发出，但未将预收款转作收入。（不考虑相关成本）...问题为：新产品研发费应调整的应纳税所得额为xx万元？ (\textit{A nationally supported key high-tech enterprise located in the urban district of a certain city is a general VAT taxpayer (manufacturing enterprise). In 2023, it generated RMB 65.00 million in product sales revenue (VAT-exclusive) and additionally earned investment income of RMB 3.20 million. For the year, the total cost of goods sold and related expenses amounted to RMB 53.00 million. Taxes and surcharges paid were RMB 3.39 million, and non-operating expenses were RMB 4.20 million. As of the end of December, the enterprise calculated an annual accounting profit before tax of RMB 7.61 million, and it had prepaid corporate income tax (CIT) of RMB 0.96 million. In January 2024, after an audit by a hired tax agent firm, the following issues were identified:(1) In mid-August, the enterprise sold a batch of products under an advance-receipt arrangement, receiving an advance payment of RMB 2.34 million (VAT-inclusive), which was deposited in the bank. In late December, the batch of products was delivered, but the advance receipt was not reclassified as revenue. (Relevant costs are ignored.)...\textbf{Question:} The adjustment to taxable income attributable to the new product R\&D expenses should be \texttt{xx} 10,000 RMB.})
    \\ \hline
    
    \textbf{Answer:} -72.75
    \\
    \bottomrule
    \end{tabular}
    \caption{An instance format with paired prompts for the TaxCalc dataset in the KA task.}
\label{tab:TaxCalc}
\end{table*}

\begin{table*}[t]
   \renewcommand{\thetable}{C10}
    \centering
    \footnotesize
    \setlength\tabcolsep{0pt}
    \renewcommand{\arraystretch}{0.8}
    
    \begin{tabular}{m{0.98\textwidth}}

    \toprule
    
    \textbf{Prompt:} 针对以下问题，请分析五个选项，并选择最适合的两个或更多，然后按字母顺序提供“A”、“B”、“C”、“D”或“E”作为你的答案，输出格式为字母组合，字母之间不要有任何其他字符，且必须是顺序的，例如：“AB”或“ABC”。 (\textit{For the following question, analyze the five options and select the two or more most appropriate ones. Then provide your answer as a sequence of letters chosen from ``A'', ``B'', ``C'', ``D'', and ``E'', listed in alphabetical order (e.g., ``AB'', or ``ABC''). })
    \\ \hline

    \multicolumn{1}{c}{\textbf{Zero-Shot}}
    \\ \hline

    \textbf{Query:} \textcolor{blue}{\{Prompt\}}\\
    \textbf{Text:} 关于不动产租赁服务的增值税处理，正确的有()。   A.个体工商户异地出租不动产，在不动产所在地预缴的增值税款，可在当期增值税税款中抵减   B.以经营租赁方式将土地出租给他人使用，按不动产经营租赁服务缴纳增值税   C.一般纳税人出租其2016年5月1日前取得的不动产可选择简易计税方法计税   D.其他个人异地出租不动产，向不动产所在地预缴税款，向居住所在地申报纳税   E.纳税人向其他个人出租不动产，可以开具增值税专用发票 (\textit{Regarding the VAT treatment of real estate leasing services, which of the following statements are correct?  A. Individual businesses leasing real estate in a different location can offset the prepaid VAT at the property’s location against their current VAT payable. B. Leasing land to others under an operating lease is subject to VAT as real estate leasing services. C. General taxpayers renting out real estate acquired before May 1, 2016, can choose the simplified tax calculation method. D. Individuals leasing real estate in a different location should prepay VAT at the property’s location and declare taxes at their place of residence. E. Taxpayers leasing real estate to individuals can issue a VAT special invoice.})
    \\ \hline
    
    \textbf{Answer:} ABC
    \\ \hline

    \multicolumn{1}{c}{\textbf{One-Shot}}
    \\ \hline

    \textbf{Query:} \textcolor{blue}{\{Prompt\}}\\
    \textbf{Example-Text:} 下列业务产生的收入中，可以作为计算业务招待费、广告费和业务宣传费税前扣除限额基数的有（）。   A.让渡无形资产使用权   B.让渡固定资产所有权   C.从事债权投资   D.接受捐赠   E.将资产转移至境外分支机构用于销售 (\textit{Among the following types of income, which can be included in the base used to compute the pre-tax deduction limits for business entertainment expenses, advertising expenses, and business promotion expenses? A.Granting the right to use intangible assets B.Transferring ownership of fixed assets C.Engaging in debt investment D.Receiving donations E.Transferring assets to an overseas branch for sale})\\
    \textbf{Example-Answer:} AE\\
    \textbf{Text:} 关于不动产租赁服务的增值税处理，正确的有()。   A.个体工商户异地出租不动产，在不动产所在地预缴的增值税款，可在当期增值税税款中抵减   B.以经营租赁方式将土地出租给他人使用，按不动产经营租赁服务缴纳增值税   C.一般纳税人出租其2016年5月1日前取得的不动产可选择简易计税方法计税   D.其他个人异地出租不动产，向不动产所在地预缴税款，向居住所在地申报纳税   E.纳税人向其他个人出租不动产，可以开具增值税专用发票 (\textit{Regarding the VAT treatment of real estate leasing services, which of the following statements are correct?  A. Individual businesses leasing real estate in a different location can offset the prepaid VAT at the property’s location against their current VAT payable. B. Leasing land to others under an operating lease is subject to VAT as real estate leasing services. C. General taxpayers renting out real estate acquired before May 1, 2016, can choose the simplified tax calculation method. D. Individuals leasing real estate in a different location should prepay VAT at the property’s location and declare taxes at their place of residence. E. Taxpayers leasing real estate to individuals can issue a VAT special invoice.})
    \\ \hline
    
    \textbf{Answer:} ABC
    \\
    \bottomrule
    \end{tabular}
    \caption{An instance format with paired prompts for the TaxMCQ dataset in the KA task.}
\label{tab:TaxMCQ}
\end{table*}

\begin{table*}[t]
   \renewcommand{\thetable}{C11}
    \centering
    \footnotesize
    \setlength\tabcolsep{0pt}
    \renewcommand{\arraystretch}{0.8}
    
    \begin{tabular}{m{0.98\textwidth}}

    \toprule
    
    \textbf{Prompt:} 对于下面提出的税务问题，请准备简要的答案，注意每项答复应保持在300字之内。 (\textit{For the tax-related question below, please provide concise answers. Each response should be limited to no more than 300 words.})
    \\ \hline
    
    \multicolumn{1}{c}{\textbf{Zero-Shot}}
    \\ \hline
    
    \textbf{Query:} \textcolor{blue}{\{Prompt\}}\\
    \textbf{Text:} 转让债权如何确定其应予减除的财产原值和合理费用？ (\textit{How should the original value of the property to be deducted and the reasonable expenses be determined when transferring a debt?})
    \\ \hline
    
    \textbf{Answer:} 采用“加权平均法”确定其应予减除的财产原值和合理费用。即以纳税人购进的同一种类债券买入价和买进过程中缴纳的税费总和，除以纳税人购进的该种类债券数量之和，乘以纳税人卖出的该种类债券数量，再加上卖出的该种类债券过程中缴纳的税费。 (\textit{he "weighted average method" is used to determine the deductible original value and reasonable expenses of the asset. Specifically, the total purchase cost and related taxes and fees for the same type of bonds are divided by the total quantity purchased to obtain the average cost per bond. This average is then multiplied by the quantity sold, and any taxes and fees paid during the sale are added.})
    \\ \hline

    \multicolumn{1}{c}{\textbf{One-Shot}}
    \\ \hline
    
    \textbf{Query:} \textcolor{blue}{\{Prompt\}}\\
    \textbf{Example-Text:} 人均耕地超过二亩但不超过三亩的地区的耕地占用税每平方米是多少？ (\textit{What is the cultivated land use tax (per square meter) in regions where the per-capita cultivated land area exceeds 2 mu but does not exceed 3 mu?})\\
    \textbf{Example-Answer:} 六元至三十元。 (\textit{from six to thirty yuan.})\\
    \textbf{Text:} 转让债权如何确定其应予减除的财产原值和合理费用？ (\textit{How should the original value of the property to be deducted and the reasonable expenses be determined when transferring a debt?})
    \\ \hline
    
    \textbf{Answer:} 采用“加权平均法”确定其应予减除的财产原值和合理费用。即以纳税人购进的同一种类债券买入价和买进过程中缴纳的税费总和，除以纳税人购进的该种类债券数量之和，乘以纳税人卖出的该种类债券数量，再加上卖出的该种类债券过程中缴纳的税费。 (\textit{The "weighted average method" is used to determine the deductible original value and reasonable expenses of the asset. Specifically, the total purchase cost and related taxes and fees for the same type of bonds are divided by the total quantity purchased to obtain the average cost per bond. This average is then multiplied by the quantity sold, and any taxes and fees paid during the sale are added.})
    \\
    \bottomrule
    \end{tabular}
    \caption{An instance format with paired prompts for the TaxQA dataset in the KA task.}
\label{tab:TaxKQA}
\end{table*}

\begin{table*}[t]
   \renewcommand{\thetable}{C12}
    \centering
    \footnotesize
    \setlength\tabcolsep{0pt}
    \renewcommand{\arraystretch}{1.8}
    
    \begin{tabular}{m{0.98\textwidth}}
    
    \toprule
    
    \textbf{Prompt:} 对于由群众提出的税务问题，请提供确切的解答，并指出相关的法律依据，确保每个回答都在300字以内。 (\textit{For tax-related inquiries raised by the general public, please provide precise answers and cite the relevant legal basis. Ensure that each response is no more than 300 Chinese characters in length.})
    \\ \hline
    
    \multicolumn{1}{c}{\textbf{Zero-Shot}}
    \\ \hline
    
    \textbf{Query:} \textcolor{blue}{\{Prompt\}}\\
    \textbf{Text:} 标题：文化事业建设费今年还有减半优惠吗？ 内容：文化事业建设费今年还有减半优惠吗？ (\textit{Title: Is there still a 50\% reduction policy for the Cultural Undertaking Construction Fee this year? Content: Is there still a 50\% reduction policy for the Cultural Undertaking Construction Fee this year?})
    \\ \hline
    
    \textbf{Answer:} 根据《财政部关于延续实施文化事业建设费优惠政策的通知》(财税〔2025〕7号)规定，自2025年1月1日至2027年12月31日，对归属中央收入的文化事业建设费，将按应缴费额的50\%减征；对归属地方收入的文化事业建设费，各省可以结合当地经济发展和文化事业需求，在50\%的幅度内减征。若已征收的费用符合减免条件，可予以退还或抵减以后应缴费用。 (\textit{According to the \emph{Notice of the Ministry of Finance on Extending Preferential Policies for the Cultural Undertakings Construction Fee} (Caishui [2025] No.~7), from January~1,~2025 to December~31,~2027, the cultural undertakings construction fee attributable to central government revenue shall be reduced by 50\% of the payable amount. For the portion attributable to local government revenue, each province may, in light of local economic development and cultural-sector needs, grant a reduction of up to 50\%. If fees already collected meet the eligibility criteria for such reductions or exemptions, they may be refunded or credited against future payable amounts.})
    \\ \hline
    
    \multicolumn{1}{c}{\textbf{One-Shot}}
    \\ \hline
    
    \textbf{Query:} \textcolor{blue}{\{Prompt\}}\\
    \textbf{Example-Text:} 标题：企业录用脱贫人口可以享受什么优惠？ 内容：企业录用脱贫人口可以享受什么优惠？ (\textit{Title: What preferential tax policies are available to enterprises that employ individuals lifted out of poverty? Content: What preferential tax policies are available to enterprises that employ individuals lifted out of poverty?})\\
    \textbf{Example-Answer:} 根据《关于进一步支持重点群体创业就业有关税收政策的公告》(苏财税〔2023〕23号)第二条的规定，自2023年1月1日至2027年12月31日，企业招用脱贫人口，以及在人力资源社会保障部门公共就业服务机构登记失业半年以上且持《就业创业证》或《就业失业登记证》的人员，与其签订1年以上期限劳动合同并依法缴纳社会保险费的，自签订劳动合同并缴纳社会保险当月起，在3年内按实际招用人数，每人每年可享受7800元的定额扣减，适用于增值税、城市维护建设税、教育费附加、地方教育附加和企业所得税。 (\textit{According to Article~2 of the \emph{Announcement on Further Supporting Tax Policies for Entrepreneurship and Employment of Key Groups} (Su Cai Shui [2023] No.~23), from January~1,~2023 to December~31,~2027, enterprises that hire (i) individuals from alleviated-poverty households or (ii) persons who have been registered as unemployed for more than six months with public employment service agencies of the human resources and social security authorities and who hold an \emph{Employment and Entrepreneurship Certificate} or an \emph{Employment and Unemployment Registration Certificate}, and that sign labor contracts with a term of at least one year and pay social insurance premiums in accordance with the law, are eligible, starting from the month in which the labor contract is executed and the social insurance premiums are paid, to claim a fixed-amount tax credit of RMB~7{,}800 per eligible employee per year, based on the actual number of hires, for up to three years. This credit may be used to offset value-added tax, urban maintenance and construction tax, education surcharge, local education surcharge, and corporate income tax}.)\\
    \textbf{Text:} 标题：文化事业建设费今年还有减半优惠吗？ 内容：文化事业建设费今年还有减半优惠吗？ (\textit{Title: Is there still a 50\% reduction policy for the Cultural Undertaking Construction Fee this year? Content: Is there still a 50\% reduction policy for the Cultural Undertaking Construction Fee this year?})
    \\ \hline
    
    \textbf{Answer:} 根据《财政部关于延续实施文化事业建设费优惠政策的通知》(财税〔2025〕7号)规定，自2025年1月1日至2027年12月31日，对归属中央收入的文化事业建设费，将按应缴费额的50\%减征；对归属地方收入的文化事业建设费，各省可以结合当地经济发展和文化事业需求，在50\%的幅度内减征。若已征收的费用符合减免条件，可予以退还或抵减以后应缴费用。 (\textit{According to the \emph{Notice of the Ministry of Finance on Extending Preferential Policies for the Cultural Undertakings Construction Fee} (Caishui [2025] No.~7), from January~1,~2025 to December~31,~2027, the cultural undertakings construction fee attributable to central government revenue shall be reduced by 50\% of the payable amount. For the portion attributable to local government revenue, each province may, in light of local economic development and cultural-sector needs, grant a reduction of up to 50\%. If fees already collected meet the eligibility criteria for such reductions or exemptions, they may be refunded or credited against future payable amounts.})
    \\
    \bottomrule
    \end{tabular}
    \caption{An instance format with paired prompts for the TaxBoard dataset in the KA task.}
\label{tab:TaxBoard}
\end{table*}

\begin{table*}[t]
   \renewcommand{\thetable}{C13}
    \centering
    \footnotesize
    \setlength\tabcolsep{0pt}
    \renewcommand{\arraystretch}{1.8}
    
    \begin{tabular}{m{0.98\textwidth}}
    
    \toprule
    
    \textbf{Prompt:} 阅读下面的犯罪事实，请判断其主要违反了以下哪一法律条款：A. 刑法第二百零一条 B. 刑法第二百零三条 C. 刑法第二百零四条 D. 刑法第二百零五条 E. 刑法第二百零六条 F. 刑法第二百零七条 G. 刑法第二百零八条 H. 刑法第二百零九条 I. 刑法第二百一十条 J. 刑法第二百一十一条 K. 刑法第二百一十二条 L. 刑法第二百二十七条 M. 刑法第二百六十四条。只需输出对应字母，不需要解释原因。 (\textit{Read the following statement of criminal facts and determine which of the following legal provisions is primarily violated: A. Article 201 of the Criminal Law; B. Article 203 of the Criminal Law; C. Article 204 of the Criminal Law; D. Article 205 of the Criminal Law; E. Article 206 of the Criminal Law; F. Article 207 of the Criminal Law; G. Article 208 of the Criminal Law; H. Article 209 of the Criminal Law; I. Article 210 of the Criminal Law; J. Article 211 of the Criminal Law; K. Article 212 of the Criminal Law; L. Article 227 of the Criminal Law; M. Article 264 of the Criminal Law. Output only the corresponding letter; no explanation is required.})
    \\ \hline
    
    \multicolumn{1}{c}{\textbf{Zero-Shot}}
    \\ \hline
    
    \textbf{Query:} \textcolor{blue}{\{Prompt\}}\\
    \textbf{Text:} 被告人叶俊在2017年4月11日10时许，在北京市丰台区七里庄“速八酒店”门前向樊某出售北京市增值税普通发票，现场被北京市公安局丰台分局东铁匠营派出所民警当场查获，起获增值税普通发票19张，票面金额共计人民币1875921.23元，且经鉴定均为真发票。 (\textit{The defendant, Ye Jun, was apprehended at approximately 10:00 on April 11, 2017, while selling Beijing value-added tax ordinary invoices to an individual surnamed Fan in front of the Super 8 Hotel in Qilizhuang, Fengtai District, Beijing. Officers from the Dongtiejianying Police Station of the Fengtai Branch of the Beijing Municipal Public Security Bureau seized 19 VAT ordinary invoices with a total face value of RMB~1,875,921.23; an appraisal confirmed that all of the invoices were genuine.})
    \\ \hline
    
    \textbf{Answer:} H
    \\ \hline
    
    \textbf{Choices:} [``A'', ``B'', ``C'', ``D'', ``E'', ``F'', ``G'', ``H'', ``I'', ``J'', ``K'', ``L'', ``M'']
    \\ \hline
    
    \textbf{Gold:} 7
    \\ \hline
    
    \multicolumn{1}{c}{\textbf{One-Shot}}
    \\ \hline
    
    \textbf{Query:} \textcolor{blue}{\{Prompt\}}\\
    \textbf{Example-Text:} 2015年1月至2016年7月，被告人金斌锋在没有实际业务往来的情况下，以收取一定比例的费用差价为条件，介绍上海欣原纸塑包装材料有限公司虚开增值税专用发票。共虚开27份，销货单位为上海江岚电子科技有限公司和上海泉易电子科技有限公司，价税合计人民币240.1万余元，税额人民币34.8万余元。上述发票已由欣原公司入账并向税务机关申报抵扣税款。金斌锋在案发后如实供述了犯罪事实。 (\textit{From January 2015 to July 2016, the defendant, Jin Binfeng, without any genuine underlying business transactions, arranged for Shanghai Xinyuan Paper \& Plastic Packaging Materials Co., Ltd.\ to issue value-added tax (VAT) special invoices by charging a service fee at a certain percentage of the invoice amount. In total, 27 invoices were falsely issued, with Shanghai Jianglan Electronic Technology Co., Ltd.\ and Shanghai Quanyi Electronic Technology Co., Ltd.\ listed as the sellers. The total amount (including VAT) was approximately RMB 2.401 million, including VAT of approximately RMB 0.348 million. These invoices were recorded in Xinyuan’s accounts and used to declare input VAT credits to the tax authorities. After the case was uncovered, Jin truthfully confessed to the criminal facts.})\\
    \textbf{Example-Answer:} D\\
    \textbf{Text:} 被告人叶俊在2017年4月11日10时许，在北京市丰台区七里庄“速八酒店”门前向樊某出售北京市增值税普通发票，现场被北京市公安局丰台分局东铁匠营派出所民警当场查获，起获增值税普通发票19张，票面金额共计人民币1875921.23元，且经鉴定均为真发票。 (\textit{The defendant, Ye Jun, was apprehended at approximately 10:00 on April 11, 2017, while selling Beijing value-added tax ordinary invoices to an individual surnamed Fan in front of the Super 8 Hotel in Qilizhuang, Fengtai District, Beijing. Officers from the Dongtiejianying Police Station of the Fengtai Branch of the Beijing Municipal Public Security Bureau seized 19 VAT ordinary invoices with a total face value of RMB~1,875,921.23; an appraisal confirmed that all of the invoices were genuine.})
    \\ \hline
    
    \textbf{Answer:} H
    \\ \hline
    
    \textbf{Choices:} [``A'', ``B'', ``C'', ``D'', ``E'', ``F'', ``G'', ``H'', ``I'', ``J'', ``K'', ``L'', ``M'']
    \\ \hline
    
    \textbf{Gold:} 7
    \\
    \bottomrule
    \end{tabular}
    \caption{An instance format with paired prompts for the TaxCrime dataset in the KA task.}
\label{tab:TaxCrime}
\end{table*}

\begin{table*}[t]
    \renewcommand{\thetable}{C14}
    \centering
    \footnotesize
    \setlength\tabcolsep{2pt}
    \renewcommand{\arraystretch}{1}
   
    \begin{tabular}{m{0.98\textwidth}}
    
    \toprule
    
    \textbf{Prompt:} 请将下述税务舆情概括为一段约300字的摘要，要求突出主要事件、观点或问题核心。 (\textit{Please summarize the following tax-related public opinion into an approximately 300-word abstract, highlighting the main events and the core viewpoints or issues.})
    \\ \hline
    
    \multicolumn{1}{c}{\textbf{Zero-Shot}}
    \\ \hline
    
    \textbf{Query:} \textcolor{blue}{\{Prompt\}}\\
    \textbf{Text:} 近日，网红二驴涉嫌商业欺诈及逃税事件引发舆论关注。多名受害人和合作方纷纷揭露其不当行为，特别是一位女老板称自己被二驴骗走3000万元。...。更有消息称，二驴的主持人华少也曾为其花费40万元包机，整场活动的费用累计超过3000万元。在这一事件的背后，二驴的税务问题也受到关注。有报道称，二驴通过其父亲并玉坤的账户进行大额收款，意图规避税务责任。这一做法引发了外界对其是否存在逃税行为的质疑。值得注意的是，二驴的妻子曾因涉嫌偷税漏税而被处罚，因此人们对二驴的税务问题更加关注。此举是否涉及逃税，以及这些收入是否依法缴纳税款，目前尚无官方回应。...。 (\textit{Recently, the influencer Er Lu has come under public scrutiny due to allegations of commercial fraud and tax evasion. Numerous victims and partners have come forward to expose his misconduct, particularly a female business owner who claims to have been defrauded of 30 million yuan. ... Furthermore, reports indicate that Er Lu's host, Hua Shao, spent 400,000 yuan on a charter flight for him, with the total expenses for the event exceeding 30 million yuan. Behind this incident, Er Lu's tax issues have also attracted attention. It has been reported that he received large sums through his father Bing Yukun's account in an apparent attempt to evade tax responsibilities. This practice has raised questions about whether he is engaging in tax evasion. Notably, Er Lu's wife has previously been penalized for tax evasion, which has drawn even more scrutiny to his tax matters. It remains unclear whether these actions involve tax evasion or if the income has been properly taxed, as there has been no official response yet. ...})
    \\ \hline
    
    \textbf{Answer:} 近日，网红二驴因涉嫌商业欺诈及逃税引发广泛关注。一名女老板曝光称，与二驴合作举办演唱会时投入了3000万元，但二驴未完成合同约定的销售目标，且未投入任何资金。女老板还提供了聊天记录和转账截图，揭露二驴虚假承诺，甚至收取了其他商家的抗位费。...。 (\textit{Recently, Chinese livestream celebrity Erlv  has attracted widespread attention following allegations of commercial fraud and tax evasion. A female business owner claimed that she invested RMB 30 million to co-organize a concert with Erlv; however, Erlv allegedly failed to meet the contractual sales targets and did not contribute any capital. The business owner further provided purported chat logs and bank-transfer screenshots, alleging that Erlv made false promises and even collected so-called booth (slotting) fees from other merchants.  ...}.)
    \\ \hline

    \multicolumn{1}{c}{\textbf{One-Shot}}
    \\ \hline

    \textbf{Query:} \textcolor{blue}{\{Prompt\}}\\
    \textbf{Example-Text:} 记者从市市场监管委获悉，为期一年的加油机作弊专项治理行动已接近尾声。截至目前，我市已完成对全市加油站的全面检查，立案查处9起案件，规范了成品油零售市场秩序，有力遏制了加油机作弊、计量作弊和偷逃税等违法行为。...。下一步，市市场监管委将继续加大执法力度，将专项治理与常态化执法检查相结合，推动建立健全加油机计量监管的长效机制，进一步规范成品油零售市场秩序。 (\textit{The reporter learned from the Municipal Market Supervision Administration that the year-long special campaign against fuel dispenser fraud is nearing its conclusion. So far, comprehensive inspections of all gas stations in the city have been completed, resulting in the filing and investigation of nine cases, thereby standardizing the retail market order for refined oil and effectively curbing illegal activities such as fuel dispenser fraud, measurement cheating, and tax evasion. ... Going forward, the market regulation authority will continue to intensify enforcement, integrate the special campaign with routine inspections, and promote the establishment of a long-term mechanism for metrological supervision of fuel dispensers, thereby further standardizing the refined-oil retail market.})\\
    \textbf{Example-Answer:} 自去年8月起，市市场监管委联合多部门开展为期一年的加油机作弊专项治理行动，....。接下来，市市场监管委将加强执法力度，推动建立长期有效的加油机计量监管机制，进一步规范市场秩序。 (\textit{Since August of last year, the Municipal Market Supervision Administration, in collaboration with multiple departments, has launched a year-long special campaign against fuel dispenser fraud. ... Going forward, the Administration will further strengthen enforcement and promote the establishment of a long-term, effective regulatory mechanism for fuel-dispenser metrology to better standardize market order.})\\
    \textbf{Text:} 近日，网红二驴涉嫌商业欺诈及逃税事件引发舆论关注。多名受害人和合作方纷纷揭露其不当行为，特别是一位女老板称自己被二驴骗走3000万元。...。更有消息称，二驴的主持人华少也曾为其花费40万元包机，整场活动的费用累计超过3000万元。在这一事件的背后，二驴的税务问题也受到关注。有报道称，二驴通过其父亲并玉坤的账户进行大额收款，意图规避税务责任。这一做法引发了外界对其是否存在逃税行为的质疑。值得注意的是，二驴的妻子曾因涉嫌偷税漏税而被处罚，因此人们对二驴的税务问题更加关注。此举是否涉及逃税，以及这些收入是否依法缴纳税款，目前尚无官方回应。.... (\textit{Recently, the influencer Er Lu  has come under public scrutiny due to allegations of commercial fraud and tax evasion. Numerous victims and partners have come forward to expose his misconduct, particularly a female business owner who claims to have been defrauded of 30 million yuan. ... Furthermore, reports indicate that Er Lu's host, Hua Shao, spent 400,000 yuan on a charter flight for him, with the total expenses for the event exceeding 30 million yuan. Behind this incident, Er Lu's tax issues have also attracted attention. It has been reported that he received large sums through his father Bing Yukun's account in an apparent attempt to evade tax responsibilities. This practice has raised questions about whether he is engaging in tax evasion. Notably, Er Lu's wife has previously been penalized for tax evasion, which has drawn even more scrutiny to his tax matters. It remains unclear whether these actions involve tax evasion or if the income has been properly taxed, as there has been no official response yet. ... })
    \\ \hline
    
    \textbf{Answer:} 近日，网红二驴因涉嫌商业欺诈及逃税引发广泛关注。一名女老板曝光称，与二驴合作举办演唱会时投入了3000万元，但二驴未完成合同约定的销售目标，且未投入任何资金。女老板还提供了聊天记录和转账截图，揭露二驴虚假承诺，甚至收取了其他商家的抗位费。...。 (\textit{Recently, Chinese livestream celebrity Erlv has attracted widespread attention following allegations of commercial fraud and tax evasion. A female business owner claimed that she invested RMB 30 million to co-organize a concert with Erlv; however, Erlv allegedly failed to meet the contractual sales targets and did not contribute any capital. The business owner further provided purported chat logs and bank-transfer screenshots, alleging that Erlv made false promises and even collected so-called booth (slotting) fees from other merchants. ...}.)
    \\
    \bottomrule
    \end{tabular}
     \caption{An instance format with paired prompts for the TaxOpinion dataset in the KA task.}
\label{tab:TaxOpinion}
\end{table*}

\begin{table*}[t]
    \renewcommand{\thetable}{C15}
    \centering
    \footnotesize
    \setlength\tabcolsep{2pt}
    \renewcommand{\arraystretch}{1}
    \begin{tabular}{m{0.98\textwidth}}
    
    \toprule
    
    \textbf{Prompt:} 阅读下列财税风险控制相关内容，提炼出“面临的风险”及“风险对应的解决方案”，只需直接输出如下格式内容：\{"面临的风险": , "风险对应的解决方案": \} (\textit{Read the following content related to fiscal and tax risk control, and extract (i) the ``Risks Encountered'' and (ii) the ``Corresponding Mitigation Measures.'' Output \emph{only} the following JSON format:
\{"Risks Encountered": , "Corresponding Mitigation Measures": \}})
    \\ \hline
    
    \multicolumn{1}{c}{\textbf{Zero-Shot}}
    \\ \hline
    
    \textbf{Query:} \textcolor{blue}{\{Prompt\}}\\
    \textbf{Text:} 建筑企业存在较大的税收风险，主要涉及错误适用税目税率、简易计税进项税额未转出、合同金额与结算金额不一致。针对这些风险，建议企业要加强对税务风险的防控。对于总承包业务未分别核算的风险，建议按照不同税率核算销售额，避免高税率缴纳；对于简易计税进项税额未转出的风险，要注意用于简易计税项目的进项税额不可抵扣；对于合同金额与结算金额不一致的风险，应根据实际情况及时进行印花税补缴。企业在实际操作中需谨慎遵守政策规定，避免风险的发生。 (\textit{Construction enterprises face substantial tax risks, which mainly stem from (i) misapplication of taxable items and applicable tax rates, (ii) failure to reverse input VAT credits when adopting the simplified taxation method, and (iii) inconsistencies between contract amounts and settlement amounts. To mitigate these risks, enterprises should strengthen tax risk management and internal controls. Regarding the risk that general contracting activities are not accounted for separately, firms are advised to segregate sales by the corresponding tax rates to avoid being taxed at a higher rate. With respect to the failure to reverse input VAT credits under the simplified method, firms should note that input VAT attributable to simplified-taxation projects is not creditable. For discrepancies between contract and settlement amounts, enterprises should, where appropriate, promptly pay any additional stamp duty based on the actual circumstances. In practice, enterprises should adhere strictly to relevant tax policies and regulations to minimize the likelihood of non-compliance.})
    \\ \hline
    
    \textbf{Answer:} \{"面临的风险": "错误适用税目税率、简易计税进项税额未转出、合同金额与结算金额不一致。", "对应的解决措施": "加强税务风险防控、按照不同税率核算销售额、注意简易计税项目进项税额不可抵扣、及时进行印花税补缴。"\} (\textit{\{"Risks Encountered": "Misapplication of tax items and rates; failure to reverse input VAT under the simplified taxation regime; inconsistencies between the contract amount and the settlement amount.", "Corresponding Mitigation Measures": "Strengthen tax risk control; account for sales revenue by applicable tax rates; note that input VAT for simplified taxation items is non-creditable; promptly settle any outstanding stamp duty through supplementary payment."\}}.)
    \\ \hline

    \multicolumn{1}{c}{\textbf{One-Shot}}
    \\ \hline

    \textbf{Query:} \textcolor{blue}{\{Prompt\}}\\
    \textbf{Example-Text:} 国家金融监管局数据显示，截至2023年末，我国普惠型小微企业贷款余额29.06万亿元，同比增长23.27\%。普惠型涉农贷款余额12.59万亿元，同比增长20.34\%。根据税法，金融企业需确保涉农和中小企业贷款损失的税前扣除符合条件。部分金融企业未正确提供税前扣除所需的完整证据，例如司法追索记录，导致税务风险。税务检查发现，甲银行未能提供完整的追索和资产清偿证明，致使部分贷款损失未获得税前扣除。政策规定要求金融企业对不同金额的贷款损失提供具体的追索记录和资产证明。建议金融机构遵守相关法规，确保贷款损失税前扣除的合规性和透明度。 (\textit{According to data released by the National Financial Regulatory Administration, by the end of 2023 China’s inclusive micro- and small-enterprise loan balance reached 29.06 trillion RMB, representing a year-on-year increase of 23.27\%. Inclusive agriculture-related loans totaled 12.59 trillion RMB, up 20.34\% year on year. Under the tax law, financial institutions must ensure that the pre-tax deduction of credit losses on agriculture-related and small- and medium-sized enterprise (SME) loans satisfies the applicable eligibility requirements. However, some institutions fail to furnish complete supporting documentation required for such deductions (e.g., records of judicial recovery efforts), thereby increasing tax compliance risk. In tax audits, Bank A was unable to provide comprehensive evidence of recovery actions and asset liquidation, resulting in the denial of pre-tax deductions for a portion of its loan losses. Relevant policies require financial institutions to submit specific recovery records and asset documentation for loan losses at different loss amounts. We therefore recommend that financial institutions strictly comply with the relevant regulations to improve the compliance and transparency of pre-tax deductions for loan losses.})\\
    \textbf{Example-Answer:} \{"面临的风险": "部分金融企业未正确提供税前扣除所需的完整证据，导致税务风险", "对应的解决措施": "建议金融机构遵守相关法规，确保贷款损失税前扣除的合规性和透明度。"\} (\textit{\{"Risks Encountered": "Some financial institutions may fail to provide complete and appropriate documentation required for pre-tax deductions, thereby increasing tax compliance risks.", "Corresponding Mitigation Measures": "It is recommended that financial institutions comply with relevant regulations and ensure the compliance and transparency of pre-tax deductions for loan losses."\}})\\
    \textbf{Text:} 建筑企业存在较大的税收风险，主要涉及错误适用税目税率、简易计税进项税额未转出、合同金额与结算金额不一致。针对这些风险，建议企业要加强对税务风险的防控。对于总承包业务未分别核算的风险，建议按照不同税率核算销售额，避免高税率缴纳；对于简易计税进项税额未转出的风险，要注意用于简易计税项目的进项税额不可抵扣；对于合同金额与结算金额不一致的风险，应根据实际情况及时进行印花税补缴。企业在实际操作中需谨慎遵守政策规定，避免风险的发生。 (\textit{Construction enterprises face substantial tax risks, which mainly stem from (i) misapplication of taxable items and applicable tax rates, (ii) failure to reverse input VAT credits when adopting the simplified taxation method, and (iii) inconsistencies between contract amounts and settlement amounts. To mitigate these risks, enterprises should strengthen tax risk management and internal controls. Regarding the risk that general contracting activities are not accounted for separately, firms are advised to segregate sales by the corresponding tax rates to avoid being taxed at a higher rate. With respect to the failure to reverse input VAT credits under the simplified method, firms should note that input VAT attributable to simplified-taxation projects is not creditable. For discrepancies between contract and settlement amounts, enterprises should, where appropriate, promptly pay any additional stamp duty based on the actual circumstances. In practice, enterprises should adhere strictly to relevant tax policies and regulations to minimize the likelihood of non-compliance.})
    \\ \hline
    
    \textbf{Answer:} \{"面临的风险": "错误适用税目税率、简易计税进项税额未转出、合同金额与结算金额不一致。", "对应的解决措施": "加强税务风险防控、按照不同税率核算销售额、注意简易计税项目进项税额不可抵扣、及时进行印花税补缴。"\} (\textit{\{"Risks Encountered": "Misapplication of tax items and rates; failure to reverse input VAT under the simplified taxation regime; inconsistencies between the contract amount and the settlement amount.", "Corresponding Mitigation Measures": "Strengthen tax risk control; account for sales revenue by applicable tax rates; note that input VAT for simplified taxation items is non-creditable; promptly settle any outstanding stamp duty through supplementary payment."\}}.)
    \\
    \bottomrule
    \end{tabular}
    \caption{An instance format with paired prompts for the TaxRisk dataset in the KA task.}
\label{tab:TaxRisk}
\end{table*}

\begin{table*}[t]
   \renewcommand{\thetable}{C16}
    \centering
    \footnotesize
    \setlength\tabcolsep{2pt}
    \renewcommand{\arraystretch}{1.2}
    
    \begin{tabular}{m{0.98\textwidth}}
    
    \toprule
    
    \textbf{Prompt:} 请提取以下税务稽查文章中的关键信息，内容包括：犯罪行为、所犯罪名与处罚结果。所犯罪名仅可在以下罪名中选择其一：持有伪造的发票罪，非法出售增值税专用发票罪，非法购买增值税专用发票罪，购买伪造的增值税专用发票罪，骗取出口退税罪，逃税罪，虚开发票罪。直接按照以下格式作答，无需添加解释说明：\{"犯罪行为": , "所犯罪名": , "处罚结果": \} (\textit{Please extract the key information from the following tax audit case report, including: criminal act, the charged offense, and the penalty outcome. The charged offense must be selected from exactly one of the following options: \textit{Crime of holding forged invoices}, \textit{Crime of illegally selling special VAT invoices}, \textit{Crime of illegally purchasing special VAT invoices}, \textit{Crime of purchasing forged special VAT invoices}, \textit{Crime of fraudulently obtaining export tax refunds}, \textit{Crime of tax evasion}, \textit{Crime of falsely issuing invoices}. Provide the answer strictly in the following format without any additional explanation: \{"Criminal Act": , "Charged Offense": , "Penalty Outcome": \}})
    \\ \hline
    
    \multicolumn{1}{c}{\textbf{Zero-Shot}}
    \\ \hline
    
    \textbf{Query:} \textcolor{blue}{\{Prompt\}}\\
    \textbf{Text:} 税务局检查发现，A石油化工有限公司在2017年至2019年期间存在多起偷税行为。具体违法事实包括将购进的非应征消费税产品变为应征消费税产品进行销售，涉及煤油和普柴的数量分别为2017年590.7吨和2018年1927.6吨，以及2019年将有机化工原料变为汽油的6700.842吨。根据相关法律规定，该行为构成偷税行为，导致公司未缴纳的消费税总额达到1769.1万元。对此，税务局依法对A公司分别处以总计8845.5万元的罚款。 (\textit{During a tax inspection, the tax authority found that Company A (Petrochemical) committed multiple acts of tax evasion between 2017 and 2019. The specific violations included converting purchased non--excise-taxable products into excise-taxable products for sale. The quantities involved were 590.7 tons of kerosene and 1,927.6 tons of diesel in 2017 and 2018, respectively; additionally, in 2019, 6,700.842 tons of organic chemical feedstock were converted into gasoline. Under the relevant statutory provisions, these actions constitute tax evasion and resulted in unpaid excise tax totaling RMB 17.691 million. Accordingly, the tax authority imposed a total fine of RMB 88.455 million on Company A in accordance with law.})
    \\ \hline
    
    \textbf{Answer:} \{"犯罪行为": "A石油化工有限公司将购进的非应征消费税产品变为应征消费税产品进行销售，涉及煤油、普柴及有机化工原料，未缴纳消费税总额达到1769.1万元。", "所犯罪名": "逃税罪", "处罚结果": "对A公司分别处以总计8845.5万元的罚款。"\} (\textit{\{"Criminal Act": "Company A (a petroleum and petrochemical enterprise) sold products initially purchased as non-excise-taxable goods after converting them into excise-taxable products, involving kerosene, ordinary diesel, and organic chemical raw materials, and evaded a total of RMB 17.691 million in excise tax.", "Charged Offense": "Tax Evasion", "Penalty Outcome": "Company A was fined a total of RMB 88.455 million."\}}.)
    \\ \hline

    \multicolumn{1}{c}{\textbf{One-Shot}}
    \\ \hline

    \textbf{Query:} \textcolor{blue}{\{Prompt\}}\\
    \textbf{Example-Text:} 2021年6月，曾某因涉嫌虚开增值税专用发票罪被捕，并于2021年12月接受了东莞市第三人民法院的公开审理。检方指控曾某利用瑞航塑胶公司名义，通过虚构交易方式获取5224059.35元的税款，并获得32份虚假发票。法院认定曾某系公司实际经营者，直接负责且主观恶性较轻。尽管曾某在庭审中未否认指控，但其辩护人提出其初犯、自愿认罪等情节。最终，法院依据《中华人民共和国刑法》第二百零五条，判处曾某有期徒刑11年。 (\textit{In June 2021, Mr.~Zeng was arrested on suspicion of the crime of falsely issuing special value-added tax (VAT) invoices, and his case was publicly tried by the Third People’s Court of Dongguan in December 2021. The prosecution alleged that Mr.~Zeng, acting in the name of Ruihang Plastics Co., Ltd., obtained RMB~5{,}224{,}059.35 in tax proceeds by fabricating transactions and acquired 32 fraudulent invoices. The court found that Mr.~Zeng was the de facto operator of the company, bore direct responsibility, and exhibited a relatively low degree of subjective maliciousness. Although Mr.~Zeng did not deny the allegations during the trial, his defense counsel argued mitigating circumstances, including that he was a first-time offender and voluntarily pleaded guilty. Ultimately, pursuant to Article~205 of the Criminal Law of the People’s Republic of China, the court sentenced Mr.~Zeng to 11 years of fixed-term imprisonment.})\\
    \textbf{Example-Answer:} \{"犯罪行为": "利用瑞航塑胶公司名义，通过虚构交易方式获取税款，并获得虚假发票。", "所犯罪名": "虚开发票罪", "处罚结果": "判处有期徒刑11年。"\} (\textit{\{"Criminal Act": "Using the name of Ruihang Plastics Co., Ltd., the defendant obtained tax refunds through fictitious transactions and procured fraudulent invoices.", "Charged Offense": "Crime of issuing false invoices.", "Penalty Outcome": "Sentenced to 11 years' imprisonment."\}})\\
    \textbf{Text:} 税务局检查发现，A石油化工有限公司在2017年至2019年期间存在多起偷税行为。具体违法事实包括将购进的非应征消费税产品变为应征消费税产品进行销售，涉及煤油和普柴的数量分别为2017年590.7吨和2018年1927.6吨，以及2019年将有机化工原料变为汽油的6700.842吨。根据相关法律规定，该行为构成偷税行为，导致公司未缴纳的消费税总额达到1769.1万元。对此，税务局依法对A公司分别处以总计8845.5万元的罚款。 (\textit{During a tax inspection, the tax authority found that Company A (Petrochemical) committed multiple acts of tax evasion between 2017 and 2019. The specific violations included converting purchased non--excise-taxable products into excise-taxable products for sale. The quantities involved were 590.7 tons of kerosene and 1,927.6 tons of diesel in 2017 and 2018, respectively; additionally, in 2019, 6,700.842 tons of organic chemical feedstock were converted into gasoline. Under the relevant statutory provisions, these actions constitute tax evasion and resulted in unpaid excise tax totaling RMB 17.691 million. Accordingly, the tax authority imposed a total fine of RMB 88.455 million on Company A in accordance with law.})
    \\ \hline
    
    \textbf{Answer:} \{"犯罪行为": "A石油化工有限公司将购进的非应征消费税产品变为应征消费税产品进行销售，涉及煤油、普柴及有机化工原料，未缴纳消费税总额达到1769.1万元。", "所犯罪名": "逃税罪", "处罚结果": "对A公司分别处以总计8845.5万元的罚款。"\} (\textit{\{"Criminal Act": "Company A (a petroleum and petrochemical enterprise) sold products initially purchased as non-excise-taxable goods after converting them into excise-taxable products, involving kerosene, ordinary diesel, and organic chemical raw materials, and evaded a total of RMB 17.691 million in excise tax.", "Charged Offense": "Tax Evasion", "Penalty Outcome": "Company A was fined a total of RMB 88.455 million."\}}.)
    \\
    \bottomrule
    \end{tabular}
    \caption{An instance format with paired prompts for the TaxInspect dataset in the KA task.}
\label{tab:TaxInspect}
\end{table*}

\begin{table*}[t]
   \renewcommand{\thetable}{C17}
    \centering
    \footnotesize
    \setlength\tabcolsep{2pt}
    \renewcommand{\arraystretch}{1.2}
    
    \begin{tabular}{m{0.98\textwidth}}
    \toprule
    
    \textbf{Prompt:} 请补全下方空项，完成后请按题目原格式返回填写结果。核心思路为文字回答，其余字段统一填写数字，若为百分比项，请勿添加百分号，仅填数字。\{"核心思路": , "原年收入": , "筹划后年收入": , "节增收入": , "原应纳增值税": , "筹划后应纳增值税": \} (\textit{Please complete the missing fields below. After completion, return the filled results in the original format. The ``Core Rationale'' should be answered in text, and all other fields must be filled with numbers only. For percentage fields, do not include the percent sign; provide only the numeric value.\{"Core Rationale": , "Original Annual Income": , "Post-Planning Annual Income": , "Incremental Income": , "Original VAT Payable": , "Post-Planning VAT Payable": \}})
    \\ \hline
    
    \multicolumn{1}{c}{\textbf{Zero-Shot}}
    \\ \hline
    
    \textbf{Query:} \textcolor{blue}{\{Prompt\}}\\
    \textbf{Text:} 某个体工商户销售水果、杂货，每月含税销售额为20600元左右，当地财政厅和国家税务局规定的增值税起征点为20000元。请计算该个体工商户全年应纳增值税额，并提出纳税筹划方案。（不考虑小微企业免征增值税优惠） (\textit{A sole proprietorship engages in the retail sale of fruits and general groceries, with an average monthly tax-inclusive sales revenue of approximately RMB 20{,}600. The VAT threshold (minimum taxable turnover) stipulated by the local Department of Finance and the State Taxation Administration is RMB 20{,}000. Please calculate the annual value-added tax (VAT) payable by this taxpayer and propose an appropriate tax planning strategy. (Assume that the VAT exemption for small and micro enterprises does not apply.)})
    \\ \hline
    
    \textbf{Answer:} \{"核心思路": "通过将每月含税销售额降低到略低于起征点，个体工商户可以避免增值税的缴纳，从而实现全年收入的增加。", "原年收入": 24, "筹划后年收入": 24.6, "节增收入": 0.6, "原应纳增值税": 0.72, "筹划后应纳增值税": 0\} (\textit{\{"Core Rationale": "By reducing the monthly VAT-inclusive sales revenue to just below the VAT exemption threshold, a self-employed household business can avoid paying VAT, thereby increasing annual income.", "Original Annual Income": 24, "Post-Planning Annual Income": 24.6, "Incremental Income": 0.6, "Original VAT Payable": 0.72, "Post-Planning VAT Payable": 0\}}.)
    \\ \hline

    \multicolumn{1}{c}{\textbf{One-Shot}}
    \\ \hline

    \textbf{Query:} \textcolor{blue}{\{Prompt\}}\\
    \textbf{Example-Text:} 某企业集团下属甲、乙两个企业。其中，甲企业适用25\%的企业所得税税率，乙企业属于需要国家扶持的高新技术企业，适用15\%的企业所得税税率。2019纳税年度，甲企业的应纳税所得额为8000万元，乙企业的应纳税所得额为9000万元。请计算甲乙两个企业以及该企业集团在2019纳税年度分别应当缴纳的企业所得税税款，并提出纳税筹划方案。 (\textit{A corporate group comprises two subsidiaries, Firm~A and Firm~B. Firm~A is subject to the standard corporate income tax (CIT) rate of 25\%, whereas Firm~B qualifies as a state-supported high-tech enterprise and is subject to a preferential CIT rate of 15\%. In the 2019 tax year, Firm~A reports taxable income of RMB~80~million, and Firm~B reports taxable income of RMB~90~million. Please compute the CIT liabilities payable by Firm~A and Firm~B, as well as the total CIT liability of the corporate group for the 2019 tax year, and propose an appropriate tax-planning strategy.})\\
    \textbf{Example-Answer:} \{"核心思路": "通过将高税率企业的部分应纳税所得转移至低税率企业，可以在集团内实现整体税负优化，降低企业集团应缴税款。", "甲企业税前所得（原始）": 8000, "乙企业税前所得（原始）": 9000, "甲企业所得税（原始）": 2000, "乙企业所得税（原始）": 1350, "集团总所得税（原始）": 3350, "甲企业税前所得（筹划后）": 7000, "乙企业税前所得（筹划后）": 10000, "甲企业所得税（筹划后）": 1750, "乙企业所得税（筹划后）": 1500, "集团总所得税（筹划后）": 3250, "税负减少": 100\} (\textit{\{"Core Rationale": "By reallocating a portion of taxable income from the high-tax-rate entity to the low-tax-rate entity, the group can achieve overall tax burden optimization and reduce the total corporate income tax payable.", "Taxable Income of Company A (Pre-tax, Original)": 8000, "Taxable Income of Company B (Pre-tax, Original)": 9000, "Corporate Income Tax of Company A (Original)": 2000, "Corporate Income Tax of Company B (Original)": 1350, "Total Group Corporate Income Tax (Original)": 3350, "Taxable Income of Company A (Pre-tax, After Planning)": 7000, "Taxable Income of Company B (Pre-tax, After Planning)": 10000, "Corporate Income Tax of Company A (After Planning)": 1750, "Corporate Income Tax of Company B (After Planning)": 1500, "Total Group Corporate Income Tax (After Planning)": 3250, "Tax Burden Reduction": 100\}})\\
    \textbf{Text:} 某个体工商户销售水果、杂货，每月含税销售额为20600元左右，当地财政厅和国家税务局规定的增值税起征点为20000元。请计算该个体工商户全年应纳增值税额，并提出纳税筹划方案。（不考虑小微企业免征增值税优惠） (\textit{A sole proprietorship engages in the retail sale of fruits and general groceries, with an average monthly tax-inclusive sales revenue of approximately RMB 20{,}600. The VAT threshold (minimum taxable turnover) stipulated by the local Department of Finance and the State Taxation Administration is RMB 20{,}000. Please calculate the annual value-added tax (VAT) payable by this taxpayer and propose an appropriate tax planning strategy. (Assume that the VAT exemption for small and micro enterprises does not apply.)})
    \\ \hline
    
    \textbf{Answer:} \{"核心思路": "通过将每月含税销售额降低到略低于起征点，个体工商户可以避免增值税的缴纳，从而实现全年收入的增加。", "原年收入": 24, "筹划后年收入": 24.6, "节增收入": 0.6, "原应纳增值税": 0.72, "筹划后应纳增值税": 0\} (\textit{\{"Core Rationale": "By reducing the monthly VAT-inclusive sales revenue to just below the VAT exemption threshold, a self-employed household business can avoid paying VAT, thereby increasing annual income.", "Original Annual Income": 24, "Post-Planning Annual Income": 24.6, "Incremental Income": 0.6, "Original VAT Payable": 0.72, "Post-Planning VAT Payable": 0\}}.)
    \\
    \bottomrule
    \end{tabular}
     \caption{An instance format with paired prompts for the TaxPlan dataset in the KA task.}
\label{tab:TaxPlan}
\end{table*}

\begin{table*}[htb!]
    \renewcommand{\thetable}{C19}
    \centering
    \footnotesize
    \setlength\tabcolsep{1pt}
    \renewcommand{\arraystretch}{1}
    
    \begin{tabular}{m{1cm}<{\centering}|m{3cm}|m{3cm}|m{3cm}|m{3cm}|cccc}
    \toprule[1pt]
    \multirow{2}{=}{\centering\textbf{Data}} & \multirow{2}{=}{\centering\textbf{Prompt}}   &  \multirow{2}{=}{\centering\textbf{Input}}   & \multirow{2}{=}{\centering\textbf{Target}}  &  \multirow{2}{=}{\centering\textbf{Output}}  &  \multicolumn{4}{c}{\textbf{Metric}}    \\ 
    &    &   &   &   &        AC & NA & IN & AVG \\
    \midrule
    \rotatebox{60}{\makecell{TaxRecite}}     & 根据问题需求，请提供法条的原文部分，不必对其含义做出解释，也不需要指明是第几条。 & 2007年12月6日通过的《中华人民共和国企业所得税法实施条例》中第一百一十一条的内容是什么？ & 企业所得税法第四十一条所称合理方法，包括：(一)可比非受控价格法；(二)再销售...(六)其他符合独立交易原则的方法。 & 企业所得税法第四十一条所称合理方法，包括：(一)可比非受控价格法；(二)再销售...(六)其他符合独立交易原则的方法。 &3&3&3&3    \\
    \midrule
    \rotatebox{60}{\makecell{TaxSum}}     & 根请根据以下文本编写一个简短的概要，字数不超过300字，重点在于提炼核心内容。 & 为配合个人所得税专项附加扣除工作，依据相关法律法规，三部门联合发布通知，明确个人所得税住房贷款利息专项附加扣除相关信息归集事项。信息归集范围为1989年1月1日（含）后发放的商业性个人住房贷款，不包括个人商用房贷款。... & 信息归集工作时间紧、任务重、要求高，各商业银行要高度重视，充分认识贯彻施行《个人所得税专项附加扣除暂行办法》的重要性和紧迫性，切实加强组织领导，务必在规定时间内做好各项工作。 & 本次信息归集工作期限紧、工作量大、标准要求严格。各商业银行应提高站位，充分把握《个人所得税专项附加扣除暂行办法》贯彻实施的重要意义与紧迫要求，强化统筹协调和责任落实，完善工作机制与保障措施，确保在规定时限内高质量完成各项工作任务。 &3&2&2&2.3    \\
    \midrule
    \rotatebox{60}{\makecell{TaxTopic}}     & 阅读标题以了解文章主题，并在提供的选项中选择最适合表达这一主题的一个，选项有：A.财商资讯\ \ B.财税表单\ \ C.地方法规\ \ D.法规解读\ \ E.纳税调整\ \ F.纳税评估\ \ G.税收优惠\ \ H.税收筹划\ \ I.税务问答\ \ J.中央法规。 & 浙江省国家税务局公告2014年第10号\ 浙江省国家税务局关于企业资产损失税前扣除申报有关问题的公告 & \centering{C}&\centering{C} &3&3&3&3    \\
    \midrule
    \rotatebox{60}{\makecell{TaxRead}}     & 通过阅读以下文本，然后回答以下问题。 & 为综合反映与科学评价资产评估机构整体实力和专业胜任能力，...问题：《资产评估机构综合评价办法》何时正式发布？ & \centering{2025年2月18日}& 2024年7月6日。《资产评估机构综合评价办法》的发布时间为2024年7月6日。 &1&1&3&1.7    \\
    \midrule
    \rotatebox{60}{\makecell{TaxCalc}}     & 针对题目，请直接给出xx处的答案，并将答案带入到问题中复述一遍，例如："业务（1）应缴纳消费税为xx万元"，保证输出字符在30个字符以内。 & 位于某市区国家重点扶持的高新技术企业为增值税一般纳税人（制造业企业），2023年销售产品取得不含税收入6500万元... 问题为：新产品研发费应调整的应纳税所得额为xx万元？ & \centering{-72.75}&\centering{-72.75} &3&3&3&3    \\
    \midrule
    \rotatebox{60}{\makecell{TaxSCQ}}     & 关于下面提出的问题，请仔细考虑四个可能的答案，并选出你认为正确的那个，仅需指出A、B、C或D，不需要解释。 & 关于退还关税，下列说法正确的是()。   A.纳税人自缴纳税款之日起3年内...D.海关应当自受理退税申请之日起60日内查实并通知纳税义务人办理退还手续 & \centering{C}&\centering{C} &3&3&3&3    \\
    \bottomrule[1pt]
    \end{tabular}
    \caption{Examples of human evaluation with metrics like accuracy(AC), naturalness(NA), informativeness(IN) and average(AVG) for human written prompts in various tax datasets. We retain the prompts with AVG above 2.}
    \label{tab:prompt1}
\end{table*}

\begin{table*}[htb!]
    \renewcommand{\thetable}{C20}
    \centering
    \footnotesize
    \setlength\tabcolsep{1pt}
    \renewcommand{\arraystretch}{1}
    \begin{tabular}{m{1cm}<{\centering}|m{3cm}|m{3cm}|m{3cm}|m{3cm}|cccc}
    \toprule[1pt]
    \multirow{2}{=}{\centering\textbf{Data}} & \multirow{2}{=}{\centering\textbf{Prompt}}   &  \multirow{2}{=}{\centering\textbf{Input}}   & \multirow{2}{=}{\centering\textbf{Target}}  &  \multirow{2}{=}{\centering\textbf{Output}}  &  \multicolumn{4}{c}{\textbf{Metric}}    \\ 
    &    &   &   &   &        AC & NA & IN & AVG \\
    \midrule
    \rotatebox{60}{\makecell{TaxMCQ}}     & 针对以下问题，请分析五个选项，并选择最适合的两个或更多，然后按字母顺序提供A、B、C、D或E作为你的答案，输出格式为字母组合，字母之间不要有任何其他字符，且必须是顺序的。 & 关于不动产租赁服务的增值税处理，正确的有()。   A.个体工商户异地出租不动产，在不动产所在地预缴的增值税款，可在当期增值税税款中抵减   B.以经营租赁方式将土地出租给他人使用... E.纳税人向其他个人出租不动产，可以开具增值税专用发票 & \centering{ABC}&\centering{A} &1&3&1&1.7    \\
    \midrule
    \rotatebox{60}{\makecell{TaxQA}}     & 对于下面提出的税务问题，请准备简要的答案，注意每项答复应保持在300字之内。 & 转让债权如何确定其应予减除的财产原值和合理费用？ & 采用“加权平均法”确定其应予减除的财产原值和合理费用。即以纳税人购进的同一种类债券买入价和买进过程中缴纳的税费总和，除以纳税人购进的该种类债券数量之和，乘以纳税人卖出的该种类债券数量，再加上卖出的该种类债券过程中缴纳的税费。 & 转让债权时，一般可采用加权平均法来确定应予减除的财产原值和合理费用：将纳税人购入同一类别债券的累计买入金额汇总后，除以累计购入数量，计算出单位成本；再用该单位成本乘以本次卖出的数量，作为可减除的财产原值，相关合理费用按实际发生情况一并计入扣除。 &2&3&2&2.3    \\
    \midrule
    \rotatebox{60}{\makecell{TaxBoard}}     & 对于由群众提出的税务问题，请提供确切的解答，并指出相关的法律依据，确保每个回答都在300字以内。 & 标题：文化事业建设费今年还有减半优惠吗？ 内容：文化事业建设费今年还有减半优惠吗？ & 根据《财政部关于延续实施文化事业建设费优惠政策的通知》(财税〔2025〕7号)规定，自2025年1月1日至2027年12月31日，对归属中央收入的文化事业建设费，将按应缴费额的50\%减征；对归属地方收入...。&有。依据《财政部关于延续实施文化事业建设费优惠政策的通知》（财税〔2025〕7号）：（1）归属中央收入...。 &2&2&3&2.3    \\
    \midrule
    \rotatebox{60}{\makecell{TaxCrime}}     & 阅读下面的犯罪事实，请判断其主要违反了以下哪一法律条款：A. 刑法第二百零一条 B. 刑法第二百零三条 C. 刑法第二百零四条 D. 刑法第二百零五条 E. 刑法第二百零六条 F. 刑法第二百零七条 G. 刑法第二百零八条 H. 刑法第二百零九条 I. 刑法第二百一十条 J. 刑法第二百一十一条 K. 刑法第二百一十二条 L. 刑法第二百二十七条 M. 刑法第二百六十四条。只需输出对应字母，不需要解释原因。 & 被告人叶俊在2017年4月11日10时许，在北京市丰台区七里庄“速八酒店”门前向樊某出售北京市增值税普通发票，现场被北京市公安局丰台分局东铁匠营派出所民警当场查获，起获增值税普通发票19张，票面金额共计人民币1875921.23元，且经鉴定均为真发票。 & \centering{H}&\centering{H} &3&3&3&3    \\
    \bottomrule[1pt]
    \end{tabular}
    \caption{Examples of human evaluation with metrics like accuracy(AC), naturalness(NA), informativeness(IN) and average(AVG) for human written prompts in various tax datasets. We retain the prompts with AVG above 2.}
    \label{tab:prompt2}
\end{table*}

\begin{table*}[htb!]
    \renewcommand{\thetable}{C21}
    \centering
    \footnotesize
    \setlength\tabcolsep{1pt}
    \renewcommand{\arraystretch}{1}
    
    \begin{tabular}{m{1cm}<{\centering}|m{3cm}|m{3cm}|m{3cm}|m{3cm}|cccc}
    \toprule[1pt]
    \multirow{2}{=}{\centering\textbf{Data}} & \multirow{2}{=}{\centering\textbf{Prompt}}   &  \multirow{2}{=}{\centering\textbf{Input}}   & \multirow{2}{=}{\centering\textbf{Target}}  &  \multirow{2}{=}{\centering\textbf{Output}}  &  \multicolumn{4}{c}{\textbf{Metric}}    \\ 
    &    &   &   &   &        AC & NA & IN & AVG \\
    \midrule
    \rotatebox{60}{\makecell{TaxOpinion}}     & 请将下述税务舆情概括为一段约300字的摘要，要求突出主要事件、观点或问题核心。 & 记者从市市场监管委获悉，为期一年的加油机作弊专项治理行动已接近尾声。...。下一步，市市场监管委将继续加大执法力度，将专项治理与常态化执法检查相结合，推动建立健全加油机计量监管的长效机制，进一步规范成品油零售市场秩序。 & 自去年8月起，市市场监管委联合多部门开展为期一年的加油机作弊专项治理行动，已基本完成。...接下来，市市场监管委将加强执法力度，推动建立长期有效的加油机计量监管机制，进一步规范市场秩序。 & 市市场监管委通报，自去年8月起，会同公安、商务、税务等部门在全市开展为期一年的加油机作弊专项治理行动，目前已接近尾声并完成对加油站的全面检查。...下一步将把专项治理与常态化检查结合，健全长效监管机制。 &3&1&3&2.3    \\
    \midrule
    \rotatebox{60}{\makecell{TaxRisk}}     & 阅读下列财税风险控制相关内容，提炼出“面临的风险”及“风险对应的解决方案”，只需直接输出如下格式内容：\{"面临的风险": , "风险对应的解决方案": \} & 建筑企业存在较大的税收风险，主要涉及错误适用税目税率、简易计税进项税额未转出、合同金额与结算金额不一致。针对这些风险，建议企业要加强对税务风险的防控。...企业在实际操作中需谨慎遵守政策规定，避免风险的发生。 &  \{"面临的风险": "错误适用税目税率、简易计税进项税额未转出、合同金额与结算金额不一致。", "对应的解决措施": "加强税务风险防控、按照不同税率核算销售额、注意简易计税项目进项税额不可抵扣、及时进行印花税补缴。"\} & \{"面临的风险":"税目税率适用错误、简易计税进项未转出、合同金额与结算金额不一致。","风险对应的解决方案":"加强税务风控；按不同税率分别核算；简易计税进项不得抵扣并转出；按实际补缴印花税。"\} &2&3&3&2.7    \\
    \midrule
    \rotatebox{60}{\makecell{TaxInspect}}     & 请提取以下税务稽查文章中的关键信息，内容包括：犯罪行为、所犯罪名与处罚结果。所犯罪名仅可在以下罪名中选择其一：持有伪造的发票罪，非法出售增值税专用发票罪，非法购买增值税专用发票罪，购买伪造的增值税专用发票罪，骗取出口退税罪，逃税罪，虚开发票罪。直接按照以下格式作答，无需添加解释说明：\{"犯罪行为": , "所犯罪名": , "处罚结果": \} & 税务局检查发现，A石油化工有限公司在2017年至2019年期间存在多起偷税行为。...根据相关法律规定，该行为构成偷税行为，导致公司未缴纳的消费税总额达到1769.1万元。对此，税务局依法对A公司分别处以总计8845.5万元的罚款。 & \{"犯罪行为": "A石油化工有限公司将购进的非应征消费税产品变为应征消费税产品进行销售，涉及煤油、普柴及有机化工原料，未缴纳消费税总额达到1769.1万元。", "所犯罪名": "逃税罪", "处罚结果": "对A公司分别处以总计8845.5万元的罚款。"\}&\{"犯罪行为":"A石油化工有限公司将购进的非应征消费税产品转换为应征消费税产品后销售，涉及煤油、普柴及有机化工原料，少缴消费税1769.1万元。","所犯罪名":"逃税罪","处罚结果":"被税务机关处罚款8845.5万元。"\} &3&3&2&2.7    \\
    \midrule
    \rotatebox{60}{\makecell{TaxPlan}}     & 请补全下方空项，完成后请按题目原格式返回填写结果。核心思路为文字回答，其余字段统一填写数字，若为百分比项，请勿添加百分号，仅填数字。\{"核心思路": , "原年收入": , "筹划后年收入": , "节增收入": , "原应纳增值税": , "筹划后应纳增值税": \} & 某个体工商户销售水果、杂货，每月含税销售额为20600元左右，当地财政厅和国家税务局规定的增值税起征点为20000元。请计算该个体工商户全年应纳增值税额，并提出纳税筹划方案。（不考虑小微企业免征增值税优惠） & \{"核心思路": "通过将每月含税销售额降低到略低于起征点，个体工商户可以避免增值税的缴纳，从而实现全年收入的增加。", "原年收入": 24, "筹划后年收入": 24.6, "节增收入": 0.6, "原应纳增值税": 0.72, "筹划后应纳增值税": 0\} &\{"核心思路": "个体工商户如果把每月含税销售额控制在起征点以下，就可以免缴增值税，从而提高全年到手收入。", "原年收入": 24, "筹划后年收入": 27, "节增收入": 3, "原应纳增值税": 0.72, "筹划后应纳增值税": 0\} &1&3&3&2.3    \\
    \bottomrule[1pt]
    \end{tabular}
    \caption{Examples of human evaluation with metrics like accuracy(AC), naturalness(NA), informativeness(IN) and average(AVG) for human written prompts in various tax datasets. We retain the prompts with AVG above 2.}
    \label{tab:prompt3}
\end{table*}

\begin{table*}[htb!]
   \renewcommand{\thetable}{C23}
    \centering
    \footnotesize
    \setlength\tabcolsep{0.5pt}
    \renewcommand{\arraystretch}{1}
    
    \begin{tabular}{m{0.7cm}<{\centering}| m{1cm}<{\centering}| m{6.2cm} |m{8cm} }
    \toprule
    \textbf{Task}  & \textbf{Dataset} & \multicolumn{1}{c}{\textbf{Prompt}} & \multicolumn{1}{c}{\textbf{English Translation}}  \\ 
    \hline
    \multirow{2}{*}{\textbf{KM}} & \rotatebox{40}{\textbf{\tiny TaxRecite}} &  根据问题需求，请提供法条的原文部分，不必对其含义做出解释，也不需要指明是第几条。& Please provide the original text of the legal provision based on the question requirements, without explaining its meaning or indicating which article it is from. \\ 
    \hline
    \multirow{10}{*}{\textbf{KU}} & \rotatebox{40}{\textbf{\tiny TaxSum}}     & 请根据以下文本编写一个简短的概要，字数不超过300字，重点在于提炼核心内容。& Please provide a brief summary based on the following text, keeping it under 300 words and focusing on extracting the core content. \\ 
    \cline{2-4}
    & \rotatebox{40}{\textbf{\tiny TaxTopic}} & 阅读标题以了解文章主题，并在提供的选项中选择最适合表达的一个选项，选项有  A.财商资讯   B.财税表单   C.地方法规   D.法规解读   E.纳税调整   F.纳税评估   G.税收优惠   H.税收筹划   I.税务问答  J.中央法规 & Read the title to grasp the article's theme and select the option that best reflects it. The options are: A. Financial Information B. Tax Forms C. Local Regulations D. Regulatory Interpretation E. Tax Adjustments F. Tax Assessment G. Tax Incentives H. Tax Planning I. Tax Q\&A J. Central Regulations.\\
    \cline{2-4}
    & \rotatebox{40}{\textbf{\tiny TaxRead}} &   在阅读完下面的文本后，请根据其内容回答问题，确保回复是从文中摘录且言简意赅，不要重复问题，答案中不要出现句号。& After reading the text, please provide concise excerpts for the questions without repeating them or using periods in your answers. \\
    \hline
     \multirow{55}{*}{\textbf{KA}} & \rotatebox{40}{\textbf{\tiny TaxCalc}} &   针对题目，请直接给出xx处的答案，并将答案带入到问题中复述一遍，例如："业务（1）应缴纳消费税为xx万元"，保证输出字符在30个字符以内。& For the question, please provide the answer from the designated section, and incorporate it into the question. For example: "The consumption tax payable by the business (1) is xx million yuan." Ensure the output is within 30 characters. \\ 
     \cline{2-4}
    & \rotatebox{40}{\textbf{\tiny TaxSCQ}} & 关于下面提出的问题，请仔细考虑四个可能的答案，并选出你认为正确的那个，仅需指出“A”、“B”、“C”或“D”，不需要解释。& Regarding the question posed below, please carefully consider the four possible answers and select the one you believe is correct. Simply indicate "A", "B", "C", or "D" without any explanation.\\  
    \cline{2-4}
    & \rotatebox{40}{\textbf{\tiny TaxMCQ}} &  针对以下问题，请分析五个选项，并选择最适合的两个或更多，然后按字母顺序提供“A”、“B”、“C”、“D”或“E”作为你的答案，输出格式为字母组合，字母之间不要有任何其他字符，且必须按顺序的排列的。& For the following question, please analyze the five options and select the two or more that are most suitable. Provide your answer in the format of a letter combination, with letters in order and without any other characters between them: "A", "B", "C", "D" or "E." \\  
    \cline{2-4}
    & \rotatebox{40}{\textbf{\tiny TaxQA}}  & 对于下面提出的税务问题，请准备简要的答案，注意每项答复应保持在300字之内。&For the tax-related questions posed below, please prepare concise answers, ensuring that each response remains within 300 words.\\ 
    \cline{2-4}
    & \rotatebox{40}{\textbf{\tiny TaxBoard}}  & 对于由群众提出的税务问题，请提供确切的解答，并指出相关的法律依据，确保每个回答都在300字以内。& For the tax questions raised by the public, please provide precise answers and indicate the relevant legal basis, ensuring that each response is within 300 words. \\
    \cline{2-4}
    & \rotatebox{40}{\textbf{\tiny TaxCrime}} & 请判断其最主要违反哪个条款：A.刑法第二百零一条 B.刑法第二百零三条 C.刑法第二百零四条 D.刑法第二百零五条 E.刑法第二百零六条 F.刑法第二百零七条 G.刑法第二百零八条 H.刑法第二百零九条 I.刑法第二百一十条 J.刑法第二百一十一条 K.刑法第二百一十二条 L.刑法第二百二十七条 M.刑法第二百六十四条。请只返回选项字母。 & Please determine which article is primarily violated: A. No. 201 of the Criminal Law B. No. 203 of the Criminal Law C. No. 204 of the Criminal Law D. No. 205 of the Criminal Law E. No. 206 of the Criminal Law F. No. 207 of the Criminal Law G. No. 208 of the Criminal Law H. No. 209 of the Criminal Law I. No. 210 of the Criminal Law J. No. 211 of the Criminal Law K. No. 212 of the Criminal Law L. No. 227 of the Criminal Law M. No. 264 of the Criminal Law. Please return only the letter of the option.\\
    \cline{2-4}
    & \rotatebox{40}{\textbf{\tiny TaxOpinion}}     & 根据所提供的税务舆情内容，请提炼出主要信息，形成一段300字左右的摘要，确保表述准确、重点突出。& Based on the provided tax-related public opinion content, please extract the main information and create a summary of approximately 300 words, ensuring that the expression is accurate and the key points are highlighted. \\
    \cline{2-4}
    & \rotatebox{40}{\textbf{\tiny TaxRisk}} & 请根据下文内容提取文章所描述的税务风险及其应对措施，按以下结构输出即可：\texttt{\textbackslash n} 面临的风险：xxx \texttt{\textbackslash n} 风险对应的解决方案：xxx & Please extract the tax risks and their corresponding countermeasures described in the following content, and output them in the specified structure: \texttt{\textbackslash n} Facing risks: xxx \texttt{\textbackslash n}  Corresponding solutions: xxx\\
    \cline{2-4}
    & \rotatebox{40}{\textbf{\tiny TaxInspect}}  & 阅读下文税务稽查材料，请识别并输出文章中涉及的三项内容：犯罪行为、所犯罪名及处罚结果。所犯罪名应从以下列表中选取：持有伪造的发票罪、非法出售增值税专用发票罪、非法购买增值税专用发票罪、购买伪造的增值税专用发票罪、骗取出口退税罪、逃税罪、虚开发票罪。只需使用下述格式作答，无需其他说明：\texttt{\textbackslash n} 犯罪行为：xxx \texttt{\textbackslash n} 所犯罪名：xxx \texttt{\textbackslash n} 处罚结果：xxx & Please identify the three items: criminal behavior, crime committed, and penalty result. Select the crime from the following list: possession of forged invoices, illegal sale of special VAT invoices, illegal purchase of special VAT invoices, purchase of forged special VAT invoices, fraudulently obtaining export tax refunds, tax evasion, and issuing false invoices. Use the format below without additional explanation: \texttt{\textbackslash n} Criminal behavior: xxx \texttt{\textbackslash n} Crime committed: xxx \texttt{\textbackslash n} Penalty result: xxx: .\\
    \cline{2-4}
    & \rotatebox{40}{\textbf{\tiny TaxPlan}}  & 请填写以下空项内容，并按格式输出。除“核心思路”需写完整句子外，其它所有空项均应只写数字，不含单位，百分数项也不要带“\%”。 \{"核心思路": ,"原始应调增所得额": ,"原始企业所得税增加额": , "原始代扣代缴个税": , "筹划后企业所得税减少": ,"筹划后个税减少":  \} & Please fill in the fields and output as specified. "Core Idea" should be a complete sentence; all other fields should contain only numbers without units, and percentage fields should not include "\%". {"Core Idea": "xxx", "Original Adjusted Income Amount": xxx, "Original Corporate Income Tax Increase": xxx, "Original Withheld Individual Income Tax": xxx, "Post-Planning Corporate Income Tax Reduction": xxx, "Post-Planning Individual Income Tax Reduction": xxx}.\\
    \bottomrule
    \end{tabular}
    \caption{Examples of human-annotated prompts and their corresponding English translations in TaxPraBen.}
    \label{tab:prompt}
\end{table*}

\begin{table*}[htb!]
   \renewcommand{\thetable}{D}
    \centering
    \footnotesize
    \setlength\tabcolsep{1pt}
    \renewcommand{\arraystretch}{1}
   
    \begin{tabular}{m{1.2cm} | m{2.5cm} | m{4.5cm} | m{4.5cm}| m{3cm}}
    \toprule[1pt]
    \multicolumn{1}{c|}{\textbf{Domain}} &
    \multicolumn{1}{c|}{\textbf{Case Description}} &
    \multicolumn{1}{c|}{\textbf{Raw Text}} &
    \multicolumn{1}{c|}{\textbf{Structured JSON}} &
    \multicolumn{1}{c}{\textbf{Evaluation Metrics}} \\
    \midrule
    \rotatebox[origin=c]{60}{\textbf{Legal}}  &  刑事案件裁判要旨与量刑结果生成：给定案情与法条要点，让模型输出“裁判理由 + 刑期/罚金等可核对数值”。(\textit{Case Ruling Summary and Sentencing Generation: Given the circumstances of the case and legal provisions, the model is asked to output "ruling reasons + verifiable numerical values such as sentence length/fines"})  &  被告人张某入户盗窃财物价值约人民币50,000元，到案后如实供述并退赔部分损失，依法可从轻处罚。综合犯罪情节与认罪认罚情况，建议判处有期徒刑一年六个月，缓刑二年，并处罚金人民币10,000元。(\textit{The defendant, Zhang Mou, stole property worth approximately 50,000 yuan by entering a house, and truthfully confessed to the crime and partially compensated for the losses after being apprehended. He can be given a lighter punishment according to the law. Considering the criminal circumstances and his guilty plea, it is recommended to sentence him to one year and six months in prison, with a two-year probation period, and a fine of 10,000 yuan.}) & 
    \{"\textcolor{blue}{核心思路}":"基于入户盗窃的法定构成与认罪认罚、退赔等从轻情节，依法确定从轻量刑并适用缓刑。",
    \par
    "\textcolor{blue}{刑期\_月}":18,
    \par    
    "\textcolor{blue}{缓刑\_月}":24,
    \par    
    "\textcolor{blue}{罚金\_元}":10000\} 
    \par \bigskip
    \{\textit{"\textcolor{blue}{Core Idea}": "Based on the legal composition of housebreaking and mitigating circumstances such as a guilty plea and compensation, a lighter sentence is determined and probation is applied."},
    \par 
    \textit{"\textcolor{blue}{Sentence\_Months}": 18},
    \par 
    \textit{"\textcolor{blue}{Probation\_Months}": 24},
    \par 
    \textit{"\textcolor{blue}{Fine\_Yuan}": 10000}\}
    &  语义维度:
    \par
    BERTScore(核心思路)
    \par \bigskip
    数值维度:
    \par
    EM Accuracy(刑期\_月, 缓刑\_月, 罚金\_元) 
    \par \bigskip
    \textit{Semantic Dimension: BERTScore (Core\_Reasoning)}
    \par \bigskip
    \textit{Numerical Dimension: EM Accuracy (Sentence\_Months, Probation\_Months, Fine\_Yuan)}
    \\
    \midrule
    \rotatebox[origin=c]{60}{\textbf{Finance}}     & 审计风险说明 + 差异率核对：给定对账摘要与抽样结果，让模型输出“风险结论与原因 + 交易金额/差异率等关键数字”。(\textit{Audit Risk explanation + discrepancy rate verification: given reconciliation summaries and sampling results, the model outputs "risk conclusion with reasons + key figures such as transaction amount/discrepancy rate".}) & 本期收入确认存在较高错报风险：部分合同在未完成履约义务前提前确认收入，且回款与发货节点不一致。抽样核对显示异常交易金额约1,250,000元，占抽样总金额12,500,000元的10.0\%。建议对相关收入进行重分类或调减，并将该类交易纳入重点复核。(\textit{There is a high risk of misstatement in revenue recognition: Some contracts recognized revenue prematurely before fulfilling obligations, and payment/receipt nodes are inconsistent with delivery. Sampling verification shows abnormal transaction amount of approximately RMB 1,250,000, accounting for 10.0\% of the total sampled amount (RMB 12,500,000). It is recommended to reclassify or reduce relevant revenue and include such transactions in key review.}) & \{"\textcolor{blue}{风险结论}":"收入确认与履约/回款节点不一致导致潜在提前确认，形成重大错报风险，需调整并重点复核。",
    \par    
    "\textcolor{blue}{异常交易金额}":1250000,
    \par
    "\textcolor{blue}{抽样总金额}":12500000,
    \par    
    "\textcolor{blue}{差异率}":0.10\} 
    \par \bigskip
    \{\textit{"\textcolor{blue}{Risk\_Conclusion}": "Inconsistency between revenue recognition and performance/payment nodes leads to potential premature recognition, creating a material misstatement risk requiring adjustment and key review"}., 
    \par 
    \textit{"\textcolor{blue}{Abnormal\_Transaction\_Amount}": 1250000}, 
    \par 
    \textit{"\textcolor{blue}{Total\_Sampled\_Amount}": 12500000}, 
    \par 
    \textit{"Discrepancy\_Rate": 0.10} \}
    &   语义维度:
    \par
    BERTScore(风险结论)
    \par \bigskip
    数值维度:
    \par
    EM Accuracy(异常交易金额, 抽样总金额, 差异率) 
    \par \bigskip
    \textit{Semantic Dimension: BERTScore (Risk\_Conclusion)}
    \par \bigskip
    \textit{Numerical Dimension: EM Accuracy (Abnormal\_Transaction\_Amount, Total\_Sampled\_Amount, Discrepancy\_Rate)}
    \\ 
    \midrule
    \rotatebox[origin=c]{60}{\textbf{Medicine}}     & 出院小结自动生成：给定病程与化验单，模型输出“诊断与处置叙述 + 生命体征/化验值等可核对数值”。(\textit{Discharge summary auto-generation: given medical records and test reports, the model outputs "diagnosis and treatment narrative + verifiable values such as vital signs/laboratory results".}) & 患者以发热、咳嗽咳痰入院，影像提示右下肺感染，诊断为肺炎；合并2型糖尿病血糖控制欠佳。入院体温 $38.5\,^{\circ}\mathrm{C}$，WBC $12.3\times 10^{9}/\mathrm{L}$，空腹血糖 11.2 mmol/L。予以头孢类抗感染、雾化祛痰及胰岛素强化控制血糖后症状改善，体温恢复正常，建议规律随访复查血常规与血糖。(\textit{The patient was admitted with fever and cough; imaging indicated right lower lung infection, diagnosed as pneumonia; complicated with poorly controlled type 2 diabetes. Admission Temperature: $38.5\,^{\circ}\mathrm{C}$，WBC $12.3\times 10^{9}/\mathrm{L}$, Fasting Blood Glucose: 11.2mmol/L. After cephalosporin anti-infection, aerosol expectoration, and insulin-intensive glucose control, symptoms improved with normalized temperature. Regular follow-up for blood routine and glucose recheck is recommended.})
    & \{"\textcolor{blue}{诊断叙述}":"依据症状与影像诊断肺炎，结合白细胞升高支持感染；同时强化控糖并随访复查关键指标。",
    \par
    "\textcolor{blue}{体温($^{\circ}\mathrm{C}$)}":38.5,
    \par
    "\textcolor{blue}{WBC(10\textsuperscript{9}/L)}":12.3,
    \par
    "\textcolor{blue}{空腹血糖(mmol/L)}":11.2\}
    \par \bigskip
    \{\textit{"\textcolor{blue}{Diagnosis\_Narrative}": "Pneumonia diagnosed based on symptoms and imaging, supported by elevated white blood cells; concurrent intensive glucose control with follow-up for key indicators}"., 
    \par
    \textit{"\textcolor{blue}{Temperature($^{\circ}\mathrm{C}$)}": 38.5}, 
    \par
    \textit{"\textcolor{blue}{WBC($10^{9}$/L)}": 12.3}, 
    \par
    \textit{"\textcolor{blue}{Fasting\_Glucose(mmol/L)}": 11.2} \} 
    &  语义维度:
    \par
    BERTScore(诊断叙述)
    \par \bigskip
    数值维度:
    \par
    EM Accuracy(体温\_$^{\circ}\mathrm{C}$, WBC\_10\textsuperscript{9}/L, 空腹血糖\_mmol/L)  
    \par \bigskip
    \textit{Semantic Dimension: BERTScore (Diagnosis\_Narrative)}
    \par \bigskip
    \textit{Numerical Dimension: EM Accuracy (Temperature\_$^{\circ}\mathrm{C}$, WBC\_$10^{9}$/L, Fasting\_Glucose(mmol/L))}
    \\
    \bottomrule[1pt]
    \end{tabular}
     \caption{Examples of our structured evaluation approach applied in legal, financial, and medical domains.}
    \label{tab:fileds}
\end{table*}

\begin{figure*}[htb!]
   \renewcommand{\thefigure}{E1}
    \centering
    \begin{subfigure}
    \centering
    \fbox{\includegraphics[width=\textwidth]{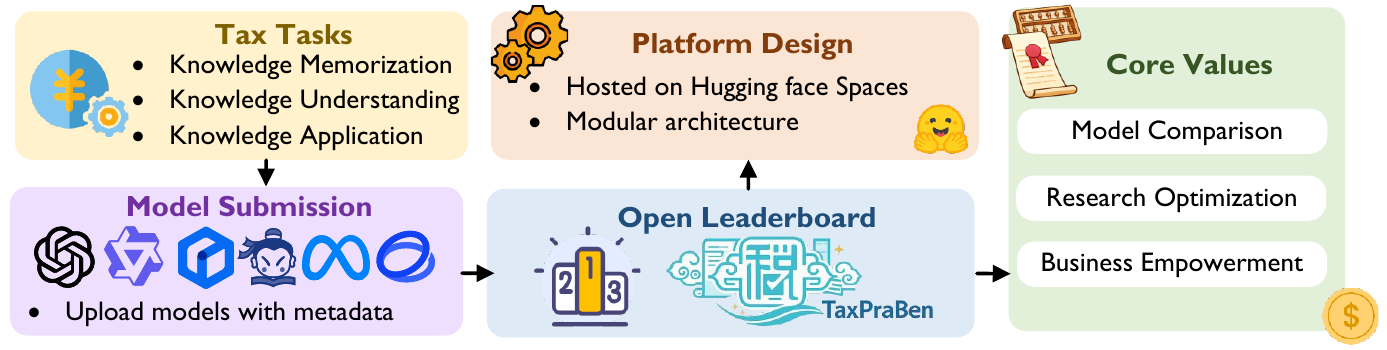}}
    \end{subfigure}

    \vspace{5em}
    
    \begin{subfigure}
    \centering
    \fbox{\includegraphics[width=\textwidth]{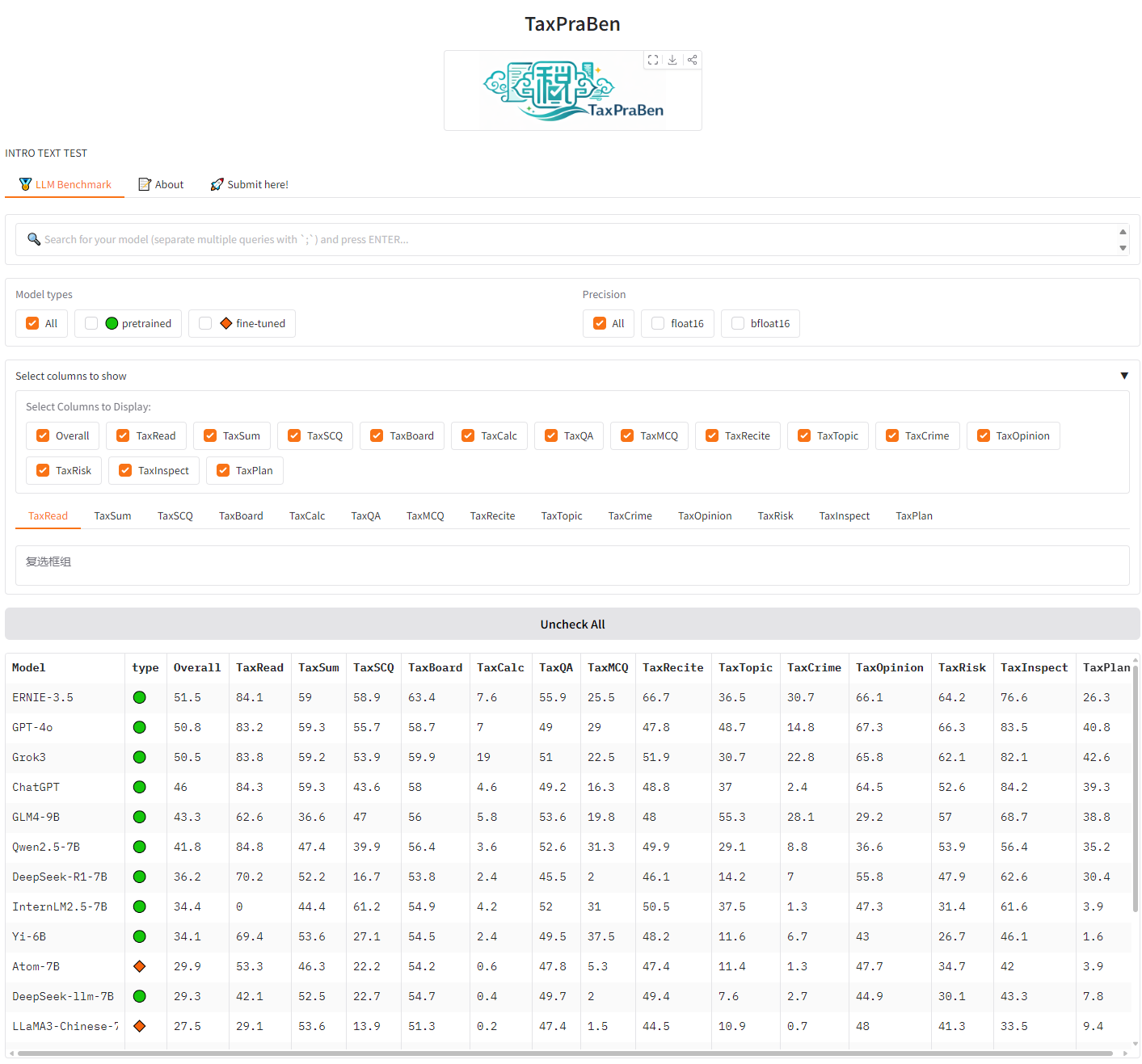}}
    \end{subfigure}
    \caption{Overview of the TaxPraBen open leaderboard and its representative interface screenshot.}
    \label{fig:leaderboard}
\end{figure*}



\begin{table*}[htb!]
    \renewcommand{\thetable}{F}
    \centering
    \footnotesize
    \setlength\tabcolsep{1pt}
    \renewcommand{\arraystretch}{1}
    
    \begin{tabular}{ m{1cm} | m{7.8cm} | m{7cm} }
    \toprule[1pt]
    \multicolumn{1}{c|}{\textbf{Dataset}} &
    \multicolumn{1}{c|}{\textbf{LLM Raw Response}} &
    \multicolumn{1}{c}{\textbf{Aligned Structured Output}} \\
    \midrule
    \rotatebox{60}{\textbf{TaxRisk}}   & \{"风险": "企业存在“不应享而享”的税收优惠问题，可能被追缴税款并取消高新技术企业资格。","措施": "企业需重视税务风险，确保合规经营，精准把握资格认定的关键要素，提升知识产权质量，加强研发费用核算，确保申报材料的真实性、准确性和完整性，严格执行后续管理措施。"\} 
    \par \bigskip
    \textit{\{"Risk": "The enterprise may have issues of improperly claiming tax incentives to which it is not entitled, and may be required to repay taxes and have its high and new technology enterprise qualification revoked.", "Measures": "The enterprise should attach importance to tax-related risks and ensure compliant operations; accurately grasp the key elements of qualification accreditation; improve the quality of intellectual property; strengthen the accounting of R\&D expenses; ensure the authenticity, accuracy, and completeness of the application materials; and strictly implement subsequent management measures."\}} & \{"\textcolor{blue}{面临的风险}": "企业存在“不应享而享”的税收优惠问题，可能被追缴税款并取消高新技术企业资格。", "\textcolor{blue}{对应的解决措施}": "企业需重视税务风险，确保合规经营，精准把握资格认定的关键要素，提升知识产权质量，加强研发费用核算，确保申报材料的真实性、准确性和完整性，严格执行后续管理措施。"\}    
    \par \bigskip
    \textit{\{"\textcolor{blue}{Risks Encountered}": "The enterprise may have issues of improperly claiming tax incentives to which it is not entitled, and may be required to repay taxes and have its high and new technology enterprise qualification revoked.", "\textcolor{blue}{Corresponding Mitigation Measures}": "The enterprise should attach importance to tax-related risks and ensure compliant operations; accurately grasp the key elements of qualification accreditation; improve the quality of intellectual property; strengthen the accounting of R\&D expenses; ensure the authenticity, accuracy, and completeness of the application materials; and strictly implement subsequent management measures."\}}\\
    \midrule
    \rotatebox{60}{\textbf{TaxInspect}}    & 从案情看，公司让财务人员黄某用其私人账户代收货款，导致账外收入无法完整入账，并在申报时通过虚假申报的方式少缴税款，属于典型的隐匿收入逃避纳税义务行为。按照刑法相关规定，可能构成逃税罪。处罚一般包括补缴税款、滞纳金及罚款，情节严重还可能追究刑责。答案：犯罪行为：通过财务人员黄某的私人账户收取货款并虚假申报，隐匿销售收入逃避税款。所犯罪名：逃税罪, 处罚结果：补缴税款、加收滞纳金，并处罚款共计1411.65万元。
    \par \bigskip
    \textit{From the case, the company instructed its finance employee, Mr.Huang, to collect sales proceeds through his personal account. As a result, off-book income could not be fully recorded, and the company underpaid taxes by making false declarations when filing. This constitutes a typical act of concealing income to evade tax obligations. Under the relevant provisions of the Criminal Law, it may constitute the crime of tax evasion. Penalties generally include paying back the underpaid taxes, late payment surcharges, and fines; in serious cases, criminal liability may also be pursued. Answer: Criminal Act: Collecting sales payments through the personal account of the finance employee, Mr. Huang, and making false tax filings to conceal sales revenue and evade taxes. Charged Offense: Tax evasion. Penalty Outcome: Payment of back taxes, imposition of late payment surcharges, and fines totaling RMB 14.1165 million.}
    & \{"\textcolor{blue}{犯罪行为}": "通过财务人员黄某的私人账户收取货款并虚假申报，隐匿销售收入逃避税款。","\textcolor{blue}{所犯罪名}": "逃税罪", "\textcolor{blue}{处罚结果}": "补缴税款、加收滞纳金，并处罚款共计1411.65万元。"\}  
    \par \bigskip
    \textit{\{"\textcolor{blue}{Criminal Act}": "Collecting sales payments through the personal account of the finance employee, Mr. Huang, and making false tax filings to conceal sales revenue and evade taxes.","\textcolor{blue}{Charged Offense}": "Tax evasion", "\textcolor{blue}{Penalty Outcome}": "Payment of back taxes, imposition of late payment surcharges, and fines totaling RMB 14.1165 million."\}}
    \\
    \midrule
    \rotatebox{60}{\textbf{TaxPlan}}  & 其中包含的计算过程：0.548-0.524 = 0.024(万元/人/年), 若10000人则合计240万元。 \{"核心思路": "通过为全体员工统一购买符合税法规定的商业健康保险，员工缴费可税前扣除，从而减少个人所得税负担。","人均年应纳个人所得税\_购买健康保险": 0.524, "人均年应纳个人所得税\_未购买健康保险": 0.548,
    "甲公司全体员工年节税": 240, "备注": "按题设人数推算"\} 
    \par \bigskip
    \textit{The calculation involved is as follows: 0.548-0.524=0.024 (ten-thousand yuan per person per year). For 10,000 people, the total is 2.4 million RMB. \{"Core Rationale": "By purchasing commercial health insurance for all employees in compliance with tax law, employees' premiums can be deducted pre-tax, thereby reducing their individual income tax burden.","Per-capita annual individual income tax payable\_with health insurance": 0.524, "Per-capita annual individual income tax payable\_without health insurance": 0.548, "Annual tax savings for all employees of Company A": 240, "Note": "Estimated based on the headcount given in the problem."\}}
    & \{"\textcolor{blue}{核心思路}": "通过为全体员工统一购买符合税法规定的商业健康保险，员工缴费可税前扣除，从而减少个人所得税负担。", "\textcolor{blue}{人均年应纳个人所得税\_购买健康保险}": 0.524, "\textcolor{blue}{人均年应纳个人所得税\_未购买健康保险}": 0.548, "\textcolor{blue}{人均节税}": 0.024, "\textcolor{blue}{甲公司全体员工年节税}": 240\} 
    \par \bigskip
    \textit{\{"\textcolor{blue}{Core Rationale}": "By purchasing commercial health insurance for all employees in compliance with tax law, employees' premiums can be deducted pre-tax, thereby reducing their individual income tax burden.", "\textcolor{blue}{Per-capita annual individual income tax payable\_with health insurance}": 0.524, "\textcolor{blue}{Per-capita annual individual income tax payable\_without health insurance}": 0.548, "\textcolor{blue}{Per-capita tax savings}": 0.024, "\textcolor{blue}{Annual tax savings for all employees of Company A}": 240\}}
    \\
    \bottomrule[1pt]
    \end{tabular}
    \caption{LLM raw responses with redundant descriptions are converted into aligned structured outputs by ChatGPT.}
    \label{tab:convert}
\end{table*}

\begin{table*}[htb!]
    \renewcommand{\thetable}{G}
    \centering
    \footnotesize
    \setlength\tabcolsep{0.2pt}
    \renewcommand{\arraystretch}{1.2}
    
    \begin{adjustbox}{max width=\textwidth}
    \begin{tabular}{ccc|ccccccccccccccccccc}
    \toprule
    \rotatebox[origin=c]{60}{\textbf{Tax Task}} & 
    \rotatebox[origin=c]{60}{\textbf{Tax Dataset}} & \rotatebox[origin=c]{60}{\textbf{Metrics}} & \rotatebox[origin=c]{90}{\textbf{ChatGPT}} &
    \rotatebox[origin=c]{90}{\textbf{GPT-4o}} &
    \rotatebox[origin=c]{90}{\textbf{Mistral-V0.3}} & \rotatebox[origin=c]{90}{\textbf{Gemma}} & \rotatebox[origin=c]{90}{\textbf{LLaMA3}} & \rotatebox[origin=c]{90}{\textbf{Bayling2}} & \rotatebox[origin=c]{90}{\textbf{Grok3}} & \rotatebox[origin=c]{90}{\textbf{$\text{DeepSeek}_\text{llm}$}} & \rotatebox[origin=c]{90}{\textbf{Baichuan2}} & \rotatebox[origin=c]{90}{\textbf{Atom}} & \rotatebox[origin=c]{90}{\textbf{Qwen2.5}} & \rotatebox[origin=c]{90}{\textbf{ChnLLaMA3}} & \rotatebox[origin=c]{90}{\textbf{ERNIE-3.5}} & \rotatebox[origin=c]{90}{\textbf{ChatGLM3}} & \rotatebox[origin=c]{90}{\textbf{Yi}} & \rotatebox[origin=c]{90}{\textbf{GLM4}} & \rotatebox[origin=c]{90}{\textbf{$\text{DeepSeek}_\text{R1}$}} & \rotatebox[origin=c]{90}{\textbf{InternLM2.5}} & \rotatebox[origin=c]{90}{\textbf{YaYi2}} \\
    \midrule
    \multicolumn{22}{c}{\textbf{Zero-Shot}} \\
    \midrule
    \multirow{2}{*}{KM} & \multirow{2}{*}{\rotatebox[origin=c]{30}{TaxRecite}} & BERTScore & 0.493 & 0.475 & 0.390 & 0.224 & 0.390 & 0.465 & 0.537 & 0.504 & 0.470 & 0.479 & 0.502 & 0.428 & \textbf{0.721} & 0.493 & 0.489 & 0.481 & 0.454 & 0.518 & 0.491 \\
    &  & BARTScore & -5.166 & -5.195 & -5.903 & -7.357 & -5.657 & -5.331 & -4.991 & -5.162 & -5.306 & -5.304 & -5.044 & -5.367 & \textbf{-3.869} & -5.189 & -5.245 & -5.221 & -5.327 & -5.073 & -5.220 \\
    \midrule
    \multirow{6}{*}{KU} & \multirow{2}{*}{\rotatebox[origin=c]{30}{TaxSum}} & BERTScore & \textbf{0.624} & 0.621 & 0.043 & 0.234 & 0.360 & 0.524 & 0.620 & 0.541 & 0.413 & 0.464 & 0.471 & 0.552 & 0.618 & 0.336 & 0.549 & 0.331 & 0.536 & 0.432 & 0.331 \\
    &  & BARTScore & -4.378 & \textbf{-4.362} & -6.826 & -7.288 & -5.934 & -5.003 & -4.364 & -4.907 & -5.523 & -5.380 & -5.234 & -4.794 & -4.389 & -5.965 & -4.775 & -5.994 & -4.932 & -5.444 & -6.058 \\
    \cline{2-22}
     & \multirow{3}{*}{\rotatebox[origin=c]{30}{TaxTopic}} & Accuracy & 0.442  & 0.556 & 0.152 & 0.000 & 0.222 & 0.054 & 0.346 & 0.128 & 0.146 & 0.211 & 0.340 & 0.193 & 0.424 & 0.116 & 0.210 & \textbf{0.630} & 0.226 & 0.434 & 0.004 \\
    &  & F1 & 0.402 & 0.559 & 0.092 & 0.000 & 0.094 & 0.084 & 0.368 & 0.071 & 0.083 & 0.093 & 0.356 & 0.095 & 0.416 & 0.147 & 0.097 & \textbf{0.618} & 0.143 & 0.437 & 0.007 \\
    &  & Macro F1 & 0.267 & 0.346 & 0.039 & 0.000 & 0.039 & 0.035 & 0.207 & 0.030 & 0.034 & 0.038 & 0.177 & 0.039 & 0.255 & 0.054 & 0.041 & \textbf{0.411} & 0.057 & 0.254 & 0.005 \\
    \cline{2-22}
    & \rotatebox[origin=c]{30}{TaxRead} & Accuracy & 0.843 & 0.832 & 0.558 & 0.001 & 0.207 & 0.658 & 0.838 & 0.421 & 0.210 & 0.533 & \textbf{0.848} & 0.291 & 0.841 & 0.806 & 0.694 & 0.626 & 0.702 & 0.000 & 0.554 \\
    \midrule
    \multirow{17}{*}{KA} & \rotatebox[origin=c]{30}{TaxCalc} & Accuracy & 0.046 & 0.070 & 0.000 & 0.000 & 0.002 & 0.004 & \textbf{0.190} & 0.004 & 0.000 & 0.006 & 0.036 & 0.002 & 0.076 & 0.018 & 0.024 & 0.058 & 0.024 & 0.042 & 0.002 \\
    \cline{2-22}
     & \multirow{3}{*}{\rotatebox[origin=c]{30}{TaxSCQ}} & Accuracy & 0.436 & 0.556 & 0.079 & 0.129 & 0.231 & 0.147 & 0.541 & 0.259 & 0.323 & 0.264 & 0.407 & 0.226 & 0.589 & 0.283 & 0.307 & 0.460 & 0.230 & \textbf{0.607} & 0.346 \\
    &  & F1 & 0.436 & 0.557 & 0.067 & 0.166 & 0.128 & 0.145 & 0.538 & 0.209 & 0.319 & 0.199 & 0.394 & 0.092 & 0.589 & 0.279 & 0.251 & 0.475 & 0.133 & \textbf{0.614} & 0.345 \\
    &  & Macro F1 & 0.437 & 0.557 & 0.072 & 0.167 & 0.134 & 0.148 & 0.538 & 0.213 & 0.320 & 0.202 & 0.395 & 0.098 & 0.589 & 0.281 & 0.255 & 0.474 & 0.138 & \textbf{0.614} & 0.344 \\
    \cline{2-22}
    & \rotatebox[origin=c]{30}{TaxMCQ} & Accuracy & 0.163 & 0.290 & 0.003 & 0.000 & 0.000 & 0.013 & 0.225 & 0.020 & 0.060 & 0.053 & 0.313 & 0.015 & 0.255 & 0.030 & \textbf{0.375} & 0.198 & 0.020 & 0.310 & 0.005 \\
    \cline{2-22}
    & \multirow{2}{*}{\rotatebox[origin=c]{30}{TaxQA}} & BERTScore & 0.477 & 0.470 & 0.373 & 0.243 & 0.407 & 0.477 & 0.487 & 0.480 & 0.458 & 0.460 & 0.508 & 0.457 & 0.531 & 0.491 & 0.479 & \textbf{0.538} & 0.434 & 0.513 & 0.508 \\
    &  & BARTScore & -4.922 & -4.895 & -5.651 & -7.287 & -5.446 & -5.091 & -4.675 & -4.860 & -4.979 & -5.051 & -4.555 & -5.098 & \textbf{-4.126} & -4.926 & -4.893 & -4.649 & -5.243 & -4.721 & -4.999 \\
    \cline{2-22}
    & \multirow{2}{*}{\rotatebox[origin=c]{30}{TaxBoard}} & BERTScore & 0.636 & 0.640 & 0.210 & 0.234 & 0.506 & 0.572 & 0.653 & 0.583 & 0.551 & 0.576 & 0.604 & 0.540 & \textbf{0.688} & 0.611 & 0.578 & 0.598 & 0.574 & 0.584 & 0.532 \\
    &  & BARTScore & -4.771 & -4.648 & -6.452 & -7.289 & -5.330 & -4.991 & -4.541 & -4.882 & -5.007 & -4.918 & -4.766 & -5.135 & \textbf{-4.200} & -4.862 & -4.888 & -4.788 & -4.981 & -4.854 & -5.212 \\
    \cline{2-22}
    & \multirow{3}{*}{\rotatebox[origin=c]{30}{TaxCrime}} & Accuracy & 0.025 & 0.135 & 0.000 & 0.035 & 0.010 & 0.010 & 0.290 & 0.030 & 0.025 & 0.030 & 0.100 & 0.015 & \textbf{0.385} & 0.000 & 0.080 & 0.315 & 0.080 & 0.030 & 0.125 \\
    &  & F1 & 0.012 & 0.135 & 0.000 & 0.001 & 0.001 & 0.001 & 0.168 & 0.001 & 0.002 & 0.001 & 0.123 & 0.001 & \textbf{0.349} & 0.000 & 0.079 & 0.347 & 0.096 & 0.002 & 0.060 \\
    &  & Macro F1 & 0.034 & 0.175 & 0.000 & 0.005 & 0.003 & 0.004 & \textbf{0.225} & 0.050 & 0.005 & 0.007 & 0.041 & 0.005 & 0.188 & 0.000 & 0.042 & 0.181 & 0.035 & 0.007 & 0.032 \\
    \cline{2-22}
    & \multirow{2}{*}{\rotatebox[origin=c]{30}{TaxOpinion}} & BERTScore & 0.764 & \textbf{0.807} & 0.104 & 0.179 & 0.237 & 0.311 & 0.786 & 0.499 & 0.203 & 0.544 & 0.384 & 0.532 & 0.784 & 0.078 & 0.468 & 0.285 & 0.650 & 0.527 & 0.374 \\
    &  & BARTScore & -4.729 & \textbf{-4.614} & -7.407 & -7.531 & -7.238 & -6.911 & -4.705 & -6.007 & -7.353 & -5.900 & -6.531 & -5.712 & -4.632 & -8.019 & -6.073 & -7.008 & -5.351 & -5.813 & -6.485 \\
    \cline{2-22}
    & \rotatebox[origin=c]{30}{TaxRisk} & RiskScore & 0.526 & \textbf{0.663} & 0.063 & 0.096 & 0.145 & 0.157 & 0.621 & 0.301 & 0.097 & 0.347 & 0.539 & 0.413 & 0.642 & 0.055 & 0.267 & 0.570 & 0.479 & 0.314 & 0.186 \\
    \cline{2-22}
    & \rotatebox[origin=c]{30}{TaxInspect} & InspectScore & \textbf{0.842} & 0.835 & 0.061 & 0.230 & 0.295 & 0.288 & 0.821 & 0.433 & 0.085 & 0.420 & 0.564 & 0.335 & 0.766 & 0.186 & 0.461 & 0.687 & 0.626 & 0.616 & 0.318 \\
    \cline{2-22}
    & \rotatebox[origin=c]{30}{TaxPlan} & PlanScore & 0.393 & 0.408 & 0.075 & 0.049 & 0.048 & 0.078 & \textbf{0.426} & 0.078 & 0.066 & 0.039 & 0.352 & 0.094 & 0.263 & 0.294 & 0.016 & 0.388 & 0.304 & 0.039 & 0.088 \\
    \midrule
    \multicolumn{22}{c}{\textbf{One-Shot}} \\
    \midrule
    \multirow{2}{*}{KM} & \multirow{2}{*}{\rotatebox[origin=c]{30}{TaxRecite}} & BERTScore & 0.510 & 0.511 & 0.398 & 0.218 & 0.374 & 0.482 & 0.547 & 0.470 & 0.484 & 0.467 & 0.495 & 0.401 & \textbf{0.666} & 0.342 & 0.474 & 0.152 & 0.454 & 0.074 & 0.318 \\
    &  & BARTScore & -5.161 & -5.133 & -5.716 & -7.399 & -5.593 & -5.308 & -4.940 & -5.208 & -5.225 & -5.325 & -5.075 & -5.495 & \textbf{-4.281} & -5.986 & -5.214 & -7.014 & -5.326 & -7.448 & -6.191 \\
    \midrule
    \multirow{6}{*}{KU} & \multirow{2}{*}{\rotatebox[origin=c]{30}{TaxSum}} & BERTScore & 0.624 & \textbf{0.629} & 0.339 & 0.549 & 0.426 & 0.471 & 0.623 & 0.576 & 0.562 & 0.573 & 0.264 & 0.587 & 0.617 & 0.200 & 0.589 & 0.254 & 0.534 & 0.206 & 0.066 \\
    &  & BARTScore & -4.385 & \textbf{-4.300} & -5.579 & -4.800 & -5.519 & -5.308 & -4.348 & -4.630 & -4.638 & -4.670 & -6.406 & -4.566 & -4.418 & -6.739 & -4.512 & -6.412 & -4.936 & -6.707 & -7.483 \\
    \cline{2-22}
    & \multirow{3}{*}{\rotatebox[origin=c]{30}{TaxTopic}} & Accuracy & 0.472 & \textbf{0.594} & 0.168 & 0.004 & 0.124 & 0.008 & 0.394 & 0.208 & 0.149 & 0.185 & 0.280 & 0.111 & 0.551 & 0.182 & 0.176 & 0.370 & 0.322 & 0.017 & 0.001 \\
    &  & F1 & 0.456 & \textbf{0.603} & 0.125 & 0.008 & 0.107 & 0.108 & 0.421 & 0.113 & 0.120 & 0.097 & 0.323 & 0.097 & 0.545 & 0.237 & 0.100 & 0.464 & 0.323 & 0.029 & 0.002 \\
    &  & Macro F1 & 0.273 & \textbf{0.379} & 0.052 & 0.003 & 0.039 & 0.052 & 0.228 & 0.047 & 0.060 & 0.040 & 0.199 & 0.037 & 0.305 & 0.085 & 0.044 & 0.283 & 0.117 & 0.012 & 0.002 \\
    \cline{2-22}
    & \rotatebox[origin=c]{30}{TaxRead} & Accuracy & 0.840 & 0.833 & 0.833 & 0.639 & 0.018 & 0.525 & 0.831 & 0.785 & 0.815 & 0.656 & 0.780 & 0.732 & \textbf{0.844} & 0.592 & 0.781 & 0.093 & 0.785 & 0.758 & 0.611 \\
    \midrule
    \multirow{17}{*}{KA} & \rotatebox[origin=c]{30}{TaxCalc} & Accuracy & 0.002 & 0.002 & 0.000 & 0.000 & 0.000 & 0.000 & 0.002 & 0.002 & 0.000 & 0.000 & 0.002 & 0.002 & \textbf{0.006} & 0.000 & 0.000 & 0.000 & 0.000 & 0.000 & 0.000 \\
    \cline{2-22}
     & \multirow{3}{*}{\rotatebox[origin=c]{30}{TaxSCQ}} & Accuracy & 0.444 & 0.551 & 0.241 & 0.234 & 0.224 & 0.271 & 0.569 & 0.241 & 0.279 & 0.221 & 0.490 & 0.230 & \textbf{0.590} & 0.201 & 0.303 & 0.333 & 0.246 & 0.301 & 0.206 \\
    &  & F1 & 0.441 & 0.551 & 0.141 & 0.129 & 0.087 & 0.279 & 0.566 & 0.111 & 0.238 & 0.093 & 0.492 & 0.088 & \textbf{0.591} & 0.236 & 0.251 & 0.451 & 0.204 & 0.313 & 0.249 \\
    &  & Macro F1 & 0.442 & 0.551 & 0.146 & 0.134 & 0.094 & 0.277 & 0.566 & 0.116 & 0.240 & 0.099 & 0.492 & 0.095 & 0.\textbf{590} & 0.236 & 0.254 & 0.452 & 0.207 & 0.315 & 0.249 \\
    \cline{2-22}
    & \rotatebox[origin=c]{30}{TaxMCQ} & Accuracy & 0.058 & 0.035 & 0.018 & 0.005 & 0.018 & 0.030 & 0.035 & 0.055 & 0.050 & 0.043 & \textbf{0.060} & 0.020 & 0.035 & 0.010 & 0.035 & 0.025 & 0.010 & 0.030 & 0.000 \\
    \cline{2-22}
    & \multirow{2}{*}{\rotatebox[origin=c]{30}{TaxQA}} & BERTScore & 0.500 & 0.490 & 0.428 & 0.232 & 0.393 & 0.513 & 0.513 & 0.467 & 0.469 & 0.441 & \textbf{0.542} & 0.426 & 0.526 & 0.508 & 0.465 & 0.492 & 0.450 & 0.538 & 0.454 \\
    &  & BARTScore & -4.778 & -4.700 & -5.412 & -7.484 & -5.501 & -5.059 & -4.498 & -4.934 & -4.992 & -5.271 & -4.400 & -5.283 & \textbf{-4.227} & -4.948 & -4.924 & -4.820 & -5.103 & -4.810 & -5.563 \\
    \cline{2-22}
    & \multirow{2}{*}{\rotatebox[origin=c]{30}{TaxBoard}} & BERTScore & 0.640 & 0.642 & 0.482 & 0.232 & 0.496 & 0.588 & 0.660 & 0.576 & 0.566 & 0.543 & 0.596 & 0.547 & \textbf{0.676} & 0.606 & 0.578 & 0.470 & 0.583 & 0.517 & 0.498 \\
    &  & BARTScore & -4.682 & -4.594 & -5.393 & -7.343 & -5.334 & -4.902 & -4.460 & -4.892 & -4.966 & -5.099 & -4.779 & -5.061 & \textbf{-4.282} & -4.803 & -4.840 & -5.474 & -4.927 & -5.246 & -5.418 \\
    \cline{2-22}
    & \multirow{3}{*}{\rotatebox[origin=c]{30}{TaxCrime}} & Accuracy & 0.150 & 0.320 & 0.080 & 0.025 & 0.070 & 0.060 & 0.425 & 0.130 & 0.090 & 0.035 & 0.250 & 0.030 & \textbf{0.535} & 0.085 & 0.195 & 0.180 & 0.140 & 0.035 & 0.105 \\
    &  & F1 & 0.180 & 0.373 & 0.097 & 0.002 & 0.080 & 0.089 & 0.408 & 0.153 & 0.101 & 0.002 & 0.275 & 0.021 & \textbf{0.505} & 0.089 & 0.146 & 0.221 & 0.156 & 0.011 & 0.082 \\
    &  & Macro F1 & 0.099 & 0.281 & 0.031 & 0.006 & 0.052 & 0.040 & \textbf{0.335} & 0.088 & 0.032 & 0.009 & 0.120 & 0.010 & 0.275 & 0.032 & 0.177 & 0.096 & 0.081 & 0.020 & 0.042 \\
    \cline{2-22}
    & \multirow{2}{*}{\rotatebox[origin=c]{30}{TaxOpinion}} & BERTScore & 0.769 & \textbf{0.808} & 0.363 & 0.299 & 0.221 & 0.183 & 0.792 & 0.690 & 0.678 & 0.643 & 0.506 & 0.674 & 0.793 & 0.063 & 0.513 & 0.088 & 0.670 & 0.101 & 0.403 \\
    &  & BARTScore & -4.710 & -4.663 & -6.365 & -6.956 & -7.328 & -7.522 & -4.669 & -5.136 & -5.224 & -5.327 & -6.030 & -5.098 & \textbf{-4.616} & -8.098 & -5.863 & -7.971 & -5.278 & -7.888 & -6.462 \\
    \cline{2-22}
    & \rotatebox[origin=c]{30}{TaxRisk} & RiskScore & 0.593 & \textbf{0.667} & 0.184 & 0.280 & 0.215 & 0.358 & 0.621 & 0.477 & 0.424 & 0.460 & 0.543 & 0.516 & 0.658 & 0.041 & 0.412 & 0.255 & 0.529 & 0.081 & 0.226 \\
    \cline{2-22}
    & \rotatebox[origin=c]{30}{TaxInspect} & InspectScore & \textbf{0.845} & 0.843 & 0.379 & 0.530 & 0.345 & 0.477 & 0.793 & 0.487 & 0.398 & 0.451 & 0.577 & 0.420 & 0.796 & 0.140 & 0.438 & 0.372 & 0.637 & 0.068 & 0.136 \\
    \cline{2-22}
    & \rotatebox[origin=c]{30}{TaxPlan} & PlanScore & 0.418 & \textbf{0.443} & 0.068 & 0.097 & 0.045 & 0.234 & 0.471 & 0.052 & 0.065 & 0.059 & 0.409 & 0.141 & 0.408 & 0.176 & 0.106 & 0.310 & 0.298 & 0.031 & 0.049 \\
    \bottomrule
    \end{tabular}
    \end{adjustbox}
    \begin{flushleft}
     \scriptsize
     \textit{\textbf{Note}: To ensure fair assessment across tasks and data types, we compute the overall average of all evaluation metrics from 5 runs as the results.}
     \end{flushleft}
    \caption{The detailed zero-shot and one-shot results for 19 popular LLMs evaluated on the TaxPraBen benchmark.}
    \label{tab:details}
\end{table*}

\end{CJK}

\end{document}